\documentclass[10pt]{article} 
\usepackage[accepted]{tmlr}

\usepackage{amsthm,amsmath,amsfonts,amssymb,mathrsfs,comment,bm,bbm}
\usepackage{mathtools}
\usepackage{hyperref}
\usepackage[sort,capitalise]{cleveref}
\usepackage{url}
\usepackage[ruled,vlined]{algorithm2e}
\SetKwInput{KwInput}{Input}                
\SetKwInput{KwOutput}{Output}

\newtheorem{lem}{Lemma}
\newtheorem{Thm}{Theorem}
\newtheorem{corollary}{Corollary}
\newtheorem{prop}{Proposition}
\theoremstyle{remark}
\newtheorem{ass}{Assumption}
 \newtheorem{remark}{Remark}

\newtheorem{fact}{Fact}
\newtheorem{definition}{Definition}

\newcommand\smallO{
  \mathchoice
    {{\scriptstyle\mathcal{O}}}
    {{\scriptstyle\mathcal{O}}}
    {{\scriptscriptstyle\mathcal{O}}}
    {\scalebox{.5}{$\scriptscriptstyle\mathcal{O}$}}
  }

\newcommand\norm[1]{\left\lVert#1\right\rVert}

\newcommand{\diag}{{\mbox{diag}}}

\numberwithin{equation}{section}

\usepackage{etoolbox}
\makeatletter
\newcommand*{\rom}[1]{\expandafter\@slowromancap\romannumeral #1@}
\makeatother

\def\eqref#1{equation~\ref{#1}}

\def\1{\bm{1}}

\DeclareMathAlphabet{\mathsfit}{\encodingdefault}{\sfdefault}{m}{sl}
\SetMathAlphabet{\mathsfit}{bold}{\encodingdefault}{\sfdefault}{bx}{n}

\newcommand{\Corr}{\mathrm{Corr}}

\DeclareMathOperator*{\argmax}{arg\,max}
\DeclareMathOperator*{\argmin}{arg\,min}

\title{Classification of high-dimensional data with spiked covariance matrix structure}

\author{\name Yin-Jen Chen \email acgoogwork@gmail.com \\
      \addr Department of Statistics\\
      North Carolina State University
      \AND
      \name Minh Tang \email mtang8@ncsu.edu \\
      \addr Department of Statistics\\
      North Carolina State University
      }

\def\month{January}  
\def\year{2026} 
\def\openreview{\url{https://openreview.net/forum?id=6bQDtTbaQs}} 

\begin{document}

\maketitle

\begin{abstract}
We study the classification problem for
high-dimensional data with $n$ observations on $p$ features where the
$p \times p$ covariance matrix $\Sigma$ exhibits a spiked eigenvalue structure and the
vector $\zeta$, given by the difference between the {\em whitened} mean
vectors, is sparse. We analyze an adaptive
classifier (adaptive with respect to the sparsity $s$) that first
performs dimension reduction on the feature vectors prior to classification in
the dimensionally reduced space, i.e., the classifier whitens
the data, then screens the features by keeping only those corresponding
to the $s$ largest coordinates of $\zeta$ and finally applies Fisher
linear discriminant on the selected features. Leveraging recent
results on entrywise matrix perturbation bounds for covariance
matrices, we show that the resulting classifier is Bayes optimal
whenever $n \rightarrow \infty$ and $s \sqrt{n^{-1} \ln p}  \rightarrow
0$. Notably, our theory also guarantees Bayes optimality for the corresponding quadratic discriminant analysis (QDA). Experimental results on real and synthetic data further indicate that the proposed approach is competitive with state-of-the-art methods while operating on a substantially lower-dimensional representation.

\end{abstract}

\section{Introduction}
Classification is one of the most important and widely studied inference tasks in
statistics and machine learning.
Among standard classifiers, the Fisher linear discriminant
analysis (LDA) rule is especially popular for its ease of
implementation and interpretation.
More specifically, suppose that we are given a 
$p$-variate random vector $\mathbf{Z}$ drawn from a mixture of 
two multivariate normal distributions $\pi_1
\mathcal{N}_{p}(\mu_{1},\Sigma)+ (1 - \pi_1)
\mathcal{N}_{p}(\mu_{2},\Sigma)$ and our goal is to classify
$\mathbf{Z}$ into one of the two classes. 
The Fisher LDA rule is then given by 
\begin{equation}\label{Fisher}
    \Upsilon_{F}(\mathbf{Z})=\begin{cases}
1 & \text{if
  $\bigl(\Sigma^{-1}(\mu_{2}-\mu_{1})\bigr)^{\top}\bigl(\mathbf{Z}-\tfrac{\mu_{1}+\mu_{2}}{2}\bigr)\leq
  \ln \tfrac{\pi_1}{1 - \pi_1}$,}
\\
2 & \text{if
  $\bigl(\Sigma^{-1}(\mu_{2}-\mu_{1})\bigr)^{\top}\bigl(\mathbf{Z}-\tfrac{\mu_{1}+\mu_{2}}{2}\bigr)
  > \ln \tfrac{\pi_1}{1 - \pi_1}$,}
\end{cases}
\end{equation}
provided that $\Sigma$ is positive definite. The Fisher
rule $\Upsilon_{F}$ is the Bayes decision rule, i.e., it
achieves the smallest mis-classification error with respect to $0$-$1$ loss for
classifying $\mathbf{Z}$. $\Upsilon_{F}$ is, however, not  directly  
applicable in practice as it involves the unknown parameters
$\Sigma^{-1}$, $\mu_{1}$ and $\mu_{2}$; we thus usually compute the
sample covariance matrix $\hat{\Sigma}$
and the sample means $\bar{X}_1$ and $\bar{X}_2$ from a given training data set with $n$
observations and then plugged these quantities into \cref{Fisher}; the resulting classifier is termed as the plug-in LDA rule $\hat{\Upsilon}_{F}$.

The classification accuracy of $\hat{\Upsilon}_{F}$ is well-understood in the
low-dimensional regime where $p\ll n$. In particular
$\hat{\Upsilon}_F$ is asymptotically optimal, i.e., it achieves the Bayes error rate as $n
\rightarrow \infty$ for fixed $p$.  This behavior, however, might no longer
holds in the high-dimensional settings $n \asymp p$ or even $n
\ll p$ where the sample covariance matrix $\hat{\Sigma}$ is
singular. Indeed, \cite{Bickel2014} proved that the error rate
for $\hat{\Upsilon}_{F}$ where we replace $\Sigma^{-1}$ with the
Moore–Penrose pseudo-inverse of $\hat{\Sigma}$ could be as bad as
random guessing and that, furthermore, the naive Bayes (NB) rule which
ignores the correlation structure in $\Sigma$ typically outperforms
$\hat{\Upsilon}_F$.

Continuing this line of inquiry, other independence rules (IR) have been proposed in
\cite{Tibshirani2002} and \cite{Fan2008} but these classifiers have
two potentially major drawbacks, namely (1) the accumulation of error in estimating
$\mu_{1}$ and $\mu_{2}$ and (2) mis-specification of the covariance matrix
$\Sigma$. In particular \cite{Fan2008} demonstrated that the noise accumulation in
estimating $\mu_{1}$ and $\mu_{2}$ alone is sufficient to degrade the
performance of IR classifiers, \cite{JunShao2011} extended this result
to show that the noise accumulation can also lead to the plug-in LDA classifier
being, asymptotically, no better than random guessing even when
$\Sigma$ is known, and finally 
\citet{Fan2012} showed that ignoring the correlations
structure in $\Sigma$ prevents IR classifiers from achieving Bayes optimality.
These results indicate the need for imposing sparsity conditions
on the features and the important role of feature selection
for mitigating the estimation errors associated with
growing dimensionality. 

To address the above limitations of IR classifiers,
\cite{JunShao2011} proposed
thresholding of {\em both} $\Sigma$ and $\mu_{2}-\mu_{1}$; this
is similar to the motivation for regularized covariance estimators in
\cite{thresholding_covariance}. In contrast, 
\citet{TonyCai2011,Fan2012,ZouHui2012,WittenTibshirani2011} and
\citet{TonyCai2019} imposed sparsity
conditions on the discriminant direction
$\beta=\Sigma^{-1}(\mu_{2}-\mu_{1})$ and use penalized estimation approaches
to recover $\beta$. \citet{TonyCai2011} also noted two 
potential advantages to this approach, namely that the assumption of sparsity on $\beta$
is less restrictive than assuming sparsity for both $\Sigma$ (or $\Sigma^{-1}$) and
$\mu_{2}-\mu_{1}$, and secondly $\Upsilon_{F}$ only depends on $\Sigma^{-1}$ and $\mu_{2}-\mu_{1}$ through their
product $\Sigma^{-1}(\mu_2 - \mu_1)$ and thus consistent estimation of $\beta$ is
sufficient. The sparsity assumption on $\beta$ 
leads to procedures and results that resemble those for 
high-dimensional linear regression even though the classification
problem is generally not formulated in terms of a linear model. For example,
\citet{TonyCai2011} and \cite{TonyCai2019} considered a linear
programming approach similar to the Dantzig selector
\citep{Dantzig} while \citet{ZouHui2012} studied a sparse discriminant
analysis rule that used the Lasso \citep{Lasso}.
Nevertheless it had been observed that, empirically, these approaches can
lead to classification rules which select a larger number of features than
necessary, and 
one possible explanation is
that the correlation structure in $\Sigma$ also induced correlations
among the entries for any estimate $\hat{\beta}$ of $\beta$. 

In this paper we consider a different approach where we first perform dimension
reduction on the feature vectors (using PCA) prior to classification
(using LDA) in the dimensionally reduced space. As PCA is an important and ubiquitous pre-processing step in
high-dimensional data analysis, there is a sizable number of work
devoted to this approach. We refer to a generic classifier from this combination as $\mathrm{lda} \circ \mathrm{pca}$. 
For example Section~9.1 of
\cite{PCA_Jolliffe} provides a detailed review of combining LDA with
different variants of PCA in the low-dimensional setting while
\cite{reduced_rank} proposed the use of the reduced rank LDA together
with class-conditional PCA in the high-dimensional settings. In terms
of applications,
$\mathrm{lda} \circ \mathrm{pca}$ is also used for faces and images recognition \citep{Face_LDA,eigenfaces,Orthogonal_written}, and
recovering genetic patterns \citep{PCA_LDA_gene} 

While $\mathrm{lda} \circ \mathrm{pca}$ classifiers arise quite
naturally, their theoretical properties in the high-dimensional
setting remains an open problem. In
particular their analysis requires possibly different techniques and assumptions
compared to those based on direct estimation of the discriminant direction $\beta$. More
specifically consistency results for $\beta$ are usually based on
ideas from high-dimensional regression including assumptions
on bounded and concentrated eigenvalues of $\Sigma$, see e.g.,
\citet{Bickel2014,Fan2008,JunShao2011,TonyCai2019}. However 
\citet{bounded_eigenvalue} noted that bounded eigenvalues are incompatible with
the presence of strong signals (eigenvalues) in the data and might be
problematic in fields such as genomics, economics and finance. These
assumptions are nevertheless imposed mainly due to the limitation of
quantifying the estimation error for $\Sigma$ in terms of the spectral
norm difference for $\hat{\Sigma} - \Sigma$.  
In contrast the idealized setting for
dimension reduction via PCA is when $\Sigma$ contains a small subset
of signal eigenvalues that accounts for most of the variability in
$\Sigma$. This idealized setting for PCA is also distinct from the
idealized setting for graphical models wherein
$\Sigma^{-1}$ is typically assumed to be sparse.
%

In summary, our contributions in this paper are as follows. We analyze the theoretical properties of a prototypical $\mathrm{lda} \circ \mathrm{pca}$
classifier under a spiked-covariance structure assumption -- 
a widely-adopted covariance model for high-dimensional data -- 
where $\Sigma$ contains a few large eigenvalues that are well-separated from the
remaining (small) eigenvalues. In particular we show in Section~\ref{Theory} that
$\mathrm{lda} \circ \mathrm{pca}$ is asymptotically Bayes-optimal as $n \rightarrow \infty$ and $n^{-1} \ln p
\rightarrow 0$. This is, to the best of our knowledge, the first
Bayes optimal consistency result for classification after performing dimension reduction via PCA.
In Section~\ref{Numerical} we demonstrate empirically, for both simulated and real data, that $\mathrm{lda} \circ \mathrm{pca}$
selects fewer features while
also having error rates that are competitive with existing
classifiers based on estimating the discriminant direction
$\beta$. The theoretical and numerical results provide a clear
example of the synergy linking dimension reduction with classification. 
Finally, in Section~\ref{Extension}
we extend $\mathrm{lda} \circ \mathrm{pca}$ to classify data with
(1) $K \geq 3$ classes or (2) two classes but with unequal covariance matrices or (3)
feature vectors that are elliptical distributed but not necessarily multivariate normal.

\section{Methodology} \label{Methodology}
\subsection{Notation and settings}
For a vector $x \in \mathbb{R}^{p}$, the
conventional $\ell_{0}$ quasi-norm and the $\ell_{1}$, $\ell_{2}$ and
$\ell_{\infty}$ norms are denoted by $\norm{x}_{0}$, $\norm{x}_{1}$,
$\norm{x}_{2}$ and $\norm{x}_{\infty}$, respectively.  For $p \in
\mathbb{N}$, we denote the set $\{1,...,p\}$ by $[p]$. Given $x \in
\mathbb{R}^{p}$ and a non-empty set $\mathcal{A} \subset [p]$, we write $x_{\mathcal{A}} = (x_{j},j\in\mathcal{A})$ to denote the {\em column} vector obtained
by keeping only the elements in $x$ whose indices belong to
$\mathcal{A}$. The operation '$\circ$', when applied to matrices, represents the Hadamard (entrywise)
product. For $i \in[p]$, $\mathbf{e}_{i}^{(p)}$ is the $i$th standard
basis vector of $\mathbb{R}^{p}$ and $\bm{1}_{p}$ is a vector in $\mathbb{R}^{p}$ whose
elements are all $1$; we also write $e_i$ and $\bm{1}$ when the choice
of $p$ is clear from context and $\mathcal{I}_{p}$ stands for the identity matrix in $\mathbb{R}^{p \times p}$. For $x,y\in \mathbb{R}^{p}$, the standard
Euclidean inner product between $x$ and $y$ is denoted as $\langle
x,y\rangle\vcentcolon=y^{\top}x$. For a matrix $M \in \mathbb{R}^{p\times q}$,
the Frobenius norm and spectral norm of $M$ are written as
$\norm{M}_{F}$ and $\norm{M}_{2}$, respectively. We will omit the subscript in $\norm{\cdot}_{2}$ when it is clear from context that the spectral norm or $\ell_{2}$ norm is intended.

The two-to-infinity norm of $M$ is
defined as
\begin{align}
    \norm{M}_{2 \rightarrow \infty}\vcentcolon= \sup_{\norm{x}_{2}=1}
  \norm{Mx}_{\infty} \equiv \max_{i\in[p]} \|M_i\|_2
\end{align}
where $M_i$ represent the $i$th row of $M$. 
We note that $\norm{M}_{2 \rightarrow \infty}\leq
\norm{M} \leq\norm{M}_{F}$.  Let $\mathbf{tr}(M)$ and $|M|$ denote the trace and determinant
of a square matrix $M$. The effective rank of a square matrix $M$ is
defined as $\mathbf{r}(M)\vcentcolon=\mathbf{tr}(M)/\norm{M}$; the
effective rank of a matrix is a useful surrogate measure for its
complexity, see e.g., \citet{HDProbability}.
Let $\mathcal{O}(\cdot)$, $\smallO(\cdot)$, $\Theta(\cdot)$ and $\Omega(\cdot)$ represent the standard big-O, little-o, big-Theta and big-Omega relationships. Finally, $\mathbbm{1}\{\mathcal{A}\}$ and $|\mathcal{A}|$ stand for the indicator
function of a set $\mathcal{A}$ and its cardinality, respectively.

In the subsequent discussion we shall generally assume, unless
specified otherwise, that we have access to training set
$\{X_{11},\dots,X_{1n_{1}}\}$ and $\{X_{21},\dots,X_{2n_{2}}\}$ whose
elements are independently and identically distributed random vector
from the $p$-variate distributions $\mathcal{N}_{p}(\mu_{1},\Sigma)$
(class 1) and $\mathcal{N}_{p}(\mu_{2},\Sigma)$ (class 2).
The sample means for each class and the {\em pooled} sample covariance matrix are denoted as 
\begin{align}
  \label{eq:sample_mean}
    &\Bar{X}_{1}=\frac{1}{n_{1}}\sum_{j=1}^{n_{1}}X_{1j}, \qquad \Bar{X}_{2}=\frac{1}{n_{2}}\sum_{j=1}^{n_{2}}X_{2j}, \\
  \label{eq:hat_Sigma}
    &
      \Hat{\Sigma}= \frac{1}{n}\Bigl(\sum_{j=1}^{n_{1}}(X_{1j}-\bar{X}_{1})(X_{1j}-\bar{X}_{1})^{\top} + \sum_{j=1}^{n_{2}}(X_{2j}-\bar{X}_{2})(X_{2j}-\bar{X}_{2})^{\top}\Bigr)
\end{align}
On numerous occasions we also need a variant of the
sample covariance matrix where we replace the sample means with
the true means. We denote this matrix as
\begin{equation}
  \label{eq:Sigma_0}
\Hat{\Sigma}_{0}=\frac{1}{n}\Bigl(\sum_{j=1}^{n_{1}}(X_{1j}-\mu_{1})(X_{1j}-\mu_{1})^{\top} + \sum_{j=1}^{n_{2}}(X_{2j}-\mu_{2})(X_{2j}-\mu_{2})^{\top}\Bigr).
\end{equation}

\subsection{Linear discriminant analysis and whitening matrix}
\label{sec:lda_whitening}

Let $\Sigma$ be a $p \times p$ positive definite matrix.
The whitening matrix $\mathcal{W}$ is a linear
transformation satisfying $\mathcal{W}\mathcal{W}^{\top}=\Sigma^{-1}$, and
is generally used to decorrelate random variables and scale their
variances to $1$. The whitening transformation is unique only up to orthogonal
transformation as $\mathcal{W} T T^{\top} \mathcal{W}^{\top} = \mathcal{W} \mathcal{W}$ for any $p
\times p$ orthogonal matrix $T$; see \cite{whitening} for a comparison between several common choices of whitening
transformation. 
Hence, for this paper, we take $\mathcal{W} = \Sigma^{-1/2}$
as the {\em unique} positive semidefinite square root of $\Sigma^{-1}$. 
Note that $\Sigma^{-1/2}$ is the whitening transformation
which minimizes the expected mean square error between the original data and
the whitened data \citep{squareroot}.
Given $\mathcal{W}$,
we define the {\em whitened} direction as
$\zeta=\mathcal{W}(\mu_{2}-\mu_{1})$ and let
$\mathcal{S}_{\zeta}=\{j:\zeta_{j}\neq0\}$. We referred to the
elements of $\mathcal{S}_{\zeta}$ as the whitened coordinates or
variables. The Fisher linear discriminant rule is then equivalent to
\begin{equation}\label{LDA whitened}
    \Upsilon_{F}(\mathbf{Z})=
\begin{cases}
1 & \text{if $\zeta^{\top} \bigl(\mathcal{W}\bigl(\mathbf{Z}-\tfrac{\mu_1
    + \mu_2}{2}\bigr)\bigr) = \zeta_{\mathcal{S}_{\zeta}}^{\top} \bigl(\mathcal{W}\bigl(\mathbf{Z}-\tfrac{\mu_1
    + \mu_2}{2}\bigr)\bigr)_{\mathcal{S}_{\zeta}}  \leq \ln \tfrac{\pi_1}{1 - \pi_1}$}, \\
2 & \text{otherwise}, 
\end{cases}
\end{equation}
where $\mathbf{Z}\sim \pi_1 \mathcal{N}_{p}(\mu_{1},\Sigma)+ (1 -
\pi_1) \mathcal{N}_{p}(\mu_{2},\Sigma)$. Recall that, for a vector
$\xi$ and a set of indices $\mathcal{A}$, the vector $\xi_{\mathcal{A}}$ is obtained from $\xi$ by keeping only the elements indexed by $\mathcal{A}$. Since $\beta=\Sigma^{-1}(\mu_{2}-\mu_{1})$ is the Bayes direction,
a significant portion of previous research is devoted to recovering $\beta$ and the
discriminative set $\mathcal{S}_{\beta}=\{j:\beta_{j}\neq0\}$. We refer
to the elements of $\mathcal{S}_{\beta}$  as the discriminative coordinates or variables.
We emphasize that in general the discriminative set and whitened set are
not the same, except for when $\Sigma$ has certain special structures,
e.g., $\Sigma$ being diagonal.

If $p \ll n$ then the empirical whitening matrix
$\hat{\mathcal{W}}=\Hat{\Sigma}^{-1/2}$ is well-defined and
the empirical whitened variables
$\hat{\zeta}=\Hat{\mathcal{W}}(\bar{X}_{2}-\bar{X}_{1})$ are
approximately decorrelated in the plug-in LDA rule, i.e., by the law
of large numbers, $\Hat{\mathcal{W}}\Sigma\Hat{\mathcal{W}}
\rightarrow \mathcal{I}_{p}$ and hence $\mathrm{Var}[\hat{\zeta}]
\approx c \mathcal{I}_{p}$ for some constant $c>0$. In contrast, the coordinates of the estimated discriminant
direction $\hat{\beta}$ are not decorrelated since
$\mathrm{Var}[\hat{\beta}] \approx c \Sigma^{-1}$. It is thus easier to quantify the impact of any arbitrary
{\em whitened} feature toward the classification accuracy than to quantify
the impact of an arbitrary {\em raw} feature. 

\subsection{Combining LDA and PCA}
\label{sec:spiked}


\cref{LDA whitened} can be viewed under the framework of first
performing dimension reduction and then doing classification in the
lower-dimensional spaces with the main focus being that thresholding
of the discriminant direction corresponds to dropping a subset of the
{\em raw} features while thresholding of the whitened direction
corresponds to dropping a subset of the {\em whitened} features. We
now discuss the estimation of the whitening matrix $\mathcal{W}$.

If $n \asymp p$ then the sample covariance matrix $\Hat{\Sigma}$ is often-times
ill-conditioned, and is furthermore singular when $n < p$. The estimation of either
the precision matrix $\Sigma^{-1}$ or the whitening matrix
$\mathcal{W}$ is thus challenging in these regimes, especially since 
$\Sigma^{-1}$ and $\mathcal{W}$ both contains $O(p^2)$ entries. An universal
approach to address this difficulty is to reduce the number of
parameters needed for estimating $\Sigma$, for example by assuming that
$\Sigma$ has a parametric form with $o(p^2)$ parameters,
or by introducing regularization terms to induce certain structures in the 
estimate $\hat{\Sigma}$. 
%
%

The use of regularized estimators for covariance matrices are popular
in high-dimensional classification, see e.g.,
\citet{thresholding_covariance,JunShao2011}, and theoretical analysis for these estimators are usually
based on assumptions about the sparsity of either $\Sigma$ or
$\Sigma^{-1}$ and then bounding the estimation error in terms of the
spectral norm differences $\|\hat{\Sigma}^{-1} - \Sigma^{-1}\|$ or
$\|\hat{\Sigma} - \Sigma\|$. There are, however, two potential drawbacks to
this approach. Firstly, quantifying the estimation error in terms of the spectral
norm can be quite loose and the resulting bounds might not capture the
difference in geometry between the eigenspaces of $\Sigma$ and that of
the perturbed matrix $\hat{\Sigma}$ (e.g., \cite{Spiked}
observed that the first few largest
eigenvalues of the sample covariance matrix $\hat{\Sigma}$ are always
larger than those for $\Sigma$ in the high-dimensional setting). Secondly, while it is not always appropriate to assume sparsity of 
$\Sigma$ or $\Sigma^{-1}$, \cite{TonyCai2011} noted that some form of sparsity is needed to guarantee consistent estimation of $\Sigma$ or $\Sigma^{-1}$
under spectral norm error. 

In this paper we assume a different structure for $\Sigma$, namely that $\Sigma$ is a spiked-covariance matrix
with a few leading eigenvalues (the spike) that are
well-separated from the remaining
eigenvalues (the bulk). 
More specifically, we assume that $\Sigma$ satisfies the
following condition. 
\begin{ass}[Spiked covariance matrix]
  \label{spiked_covariance} Let
  $\mathbf{u}_{1},\dots,\mathbf{u}_{d}$ be orthonormal vectors in
  $\mathbb{R}^{p}$ and assume that the covariance matrix $\Sigma$ for the
  $p$-variate distributions $\mathcal{N}_p(\mu_1, \Sigma)$ and
  $\mathcal{N}_p(\mu_2, \Sigma)$ is of the form
\begin{align}
    \Sigma&=\:\sum_{k=1}^{d}\lambda_{k}\mathbf{u}_{k}\mathbf{u}_{k}^{\top}+\sigma^{2}\mathcal{I}_{p}
            = \mathcal{U}\Lambda\mathcal{U}^{\top}+\sigma^{2}\mathcal{I}_{p}.
\end{align}
Here $\Lambda=\mathrm{diag}(\lambda_{k})$ is a $d \times d$ diagonal matrix,
$\mathcal{U}=(\,\mathbf{u}_{k}\,),\:k\in[d]$ is a $p \times d$
matrix with orthonormal columns,
and $\mathcal{I}_{p}$ is the identity matrix. We assume implicitly that
$\lambda_{1}\geq\cdots\geq\lambda_{d} > 0$, $\sigma>0$ and $d\ll p$.
\end{ass}
Covariance matrices with spiked structures have been studied extensively in
the high-dimensional statistics literature, see e.g., \citet{Spiked}
and Chapter 11 of \citet{bai_sample_cov}, and there is a significant
number of results for consistent estimation of $\mathcal{U}$ under the
assumption that the support of $\mathcal{U}$ is sparse, e.g., either
that the number of non-zero rows of $\mathcal{U}$ is small compared to
$p$ or that the $\ell_{q}$ quasi-norm, for some $q \in [0,1]$, of the
columns of $\mathcal{U}$ are bounded. The case when $q = 0$ and $q >
0$ correspond to ``hard'' and ``soft'' sparsity constraints,
respectively; 
see for example \cite{sparse_AOS,sparse_AOS1,sparse,Tony_sparse} and
the references therein. In contrasts to the above cited results, in this
paper we do not impose sparsity conditions on $\mathcal{U}$ but
instead assume that $\mathcal{U}$ have bounded coherence, namely that
the maximum $\ell_2$ norm of the rows of $\mathcal{U}$ are of order
$O(p^{-1/2})$; see \cref{Bounded_Coherence} for a precise
statement. The resulting matrix $\Sigma$ will no longer be sparse. 
The main rationale for assuming bounded coherence is that
the spiked eigenvalues $\Lambda$ can grow linearly with
$p$ while still guaranteeing bounded variance in $\Sigma$.
There is thus a large gap between the spiked eigenvalues and
the bulk eigenvalues, and this justifies the use of PCA as a
pre-processing step.  

If $\Sigma$ satisfies Condition~\ref{spiked_covariance} then the
whitening matrix is given by
\begin{align}\label{eq:population_whitening}
    \mathcal{W}=\mathcal{U}\mathcal{D}\mathcal{U}^{\top}+\frac{1}{\sigma}(\mathcal{I}_{p}-\mathcal{U}\mathcal{U}^{\top})
\end{align}
where $\mathcal{D}=\mathrm{diag}(\eta_{k})$ is a $d \times d$ diagonal matrix
with diagonal entries $\eta_{k}=(\lambda_{k}+\sigma^{2})^{-1/2}$.
The above form for $\mathcal{W}$ suggests the following
estimation procedure. 
\begin{enumerate}
  \item Extract the $d$ largest eigenvalues and corresponding
    eigenvectors of the pooled sample covariance matrix
    $\hat{\Sigma}$. Let $\hat{\Lambda}$ denote the diagonal matrix of
    these $d$ largest eigenvalues and $\hat{\mathcal{U}}$ denote the
    $p \times d$ orthogonal matrix whose columns are the
    corresponding eigenvectors. 
  \item Estimate the non-spiked eigenvalues by
    \begin{equation}
      \label{eq:hat_sigma}
      \Hat{\sigma}^{2}= \frac{\mathrm{tr}(\Hat{\Sigma})-\mathrm{tr}(\Hat{\Lambda})}{p-d}.
      \end{equation}
\item Let
  $\hat{\mathcal{D}}=(\hat{\Lambda}+\hat{\sigma}^{2}\mathcal{I}_{d})^{-1/2}$
  and estimate $\mathcal{W}$ by
  $\hat{\mathcal{W}}=\Hat{\mathcal{U}}\hat{\mathcal{D}}\Hat{\mathcal{U}}^{\top}+\hat{\sigma}^{-1}(\mathcal{I}_{p}-\Hat{\mathcal{U}}\Hat{\mathcal{U}}^{\top})$.
\end{enumerate}

Although the sample covariance matrix $\hat{\Sigma}$ is generally a poor estimate of
$\Sigma$ when $n \ll p$, the eigenvectors $\hat{\mathcal{U}}$
corresponding to the $d$ largest eigenvalues of $\hat{\Sigma}$ are 
nevertheless accurate estimates of $\mathcal{U}$. In particular,
\cite{l_infinity} and \cite{Cape2019} provide uniform error
bound for $\min_{T} \|\hat{\mathcal{U}} T - \mathcal{U}\|_{2
  \to \infty}$ where the minimum is taken over all $d \times d$
orthogonal transformations $T$; see also Theorem~\ref{sharp
  two-to-infinity} of the current paper. We can thus transform
$\hat{\mathcal{U}}$ by an orthogonal transformation $T$ so that the
resulting (transformed) rows of
$\hat{\mathcal{U}}$  are uniformly close to the corresponding rows of
$\mathcal{U}$. 

Given the above estimate for $\mathcal{W}$, we then have the prototypical $\mathrm{lda} \circ \mathrm{pca}$ classifier in
Algorithm~\ref{alg:PCLDA}. 
Algorithm~\ref{alg:PCLDA} contains two tuning
parameters, namely $d$, the number of principal components in the
PCA step and $s$, the number of coordinates of estimated whitened direction $\hat{\zeta}$ that
we preserve. The choices for $d$ and $s$ correspond to the number of
spiked eigenvalues in $\Sigma$ and the sparsity level of the whitened
direction $\zeta$. 
We show in Section~\ref{Theory} that, under certain mild conditions,
one can consistently estimate $s$ using \cref{hard
thresholding}. Consistent estimation of $d$ can be obtained using results in
\cite{num_bai2002,num1_alessi2010,num2_JASA} among others. Algorithm~\ref{alg:PCLDA} thus yields a classifier
that is adaptive with respect to both $d$ and $s$.

\begin{algorithm}[H] 
\DontPrintSemicolon 
\SetNoFillComment 

\KwInput{$\Bar{X}_{1}$, $\Bar{X}_{2}$, $\hat{\Sigma}$ and the test sample $\mathbf{Z}$}
\KwOutput{$\hat{\Upsilon}_{\mathrm{lda} \circ \mathrm{pca}}(\mathbf{Z})$}


\SetKwProg{Proc}{Algorithm}{}{} 
\Proc{}{ 

  \tcp*[h]{Step 1:} Perform PCA on the feature vectors for the training data (standard PCA approach)\; 

  \tcp*[h]{Step 2:} Extract the $d$ largest principal components and obtain $\hat{\mathcal{W}}=\Hat{\mathcal{U}}\hat{\mathcal{D}}\Hat{\mathcal{U}}^{\top}+\hat{\sigma}^{-1}(\mathcal{I}_{p}-\Hat{\mathcal{U}}\Hat{\mathcal{U}}^{\top})$.\; 

  \tcp*[h]{Step 3:} Take $\Tilde{X}_{i}=\hat{\mathcal{W}}\Bar{X}_{i},\:i=1,2$ and $\Tilde{X}_{a}=0.5(\Tilde{X}_{1}+\Tilde{X}_{2})$. Let $\hat{\zeta} = \tilde{X}_2 - \tilde{X}_1$ and form the indices set $\hat{\mathcal{S}}$ by selecting the $s$ largest coordinates of $\Hat{\zeta}$ in modulus.\; 

  \tcp*[h]{Step 4:} Given the test sample $\mathbf{Z}$, take $\Tilde{\mathbf{Z}}=\hat{\mathcal{W}}\mathbf{Z}$ and plug the sub-vector of $\hat{\zeta}$, $\Tilde{\mathbf{Z}}$ and $\Tilde{X}_{a}$ corresponding to the indices in $\hat{\mathcal{S}}$ into the Fisher discriminant rule, i.e.,\; 
  \begin{align}\label{SCLDA rule}
    \hat{\Upsilon}_{\mathrm{lda} \circ \mathrm{pca}}(\mathbf{Z})=
    \begin{cases}
    1 & \text{ if $\Hat{\zeta}_{\Hat{S}}^{\top}\big[\Tilde{\mathbf{Z}}-\Tilde{X}_{a}\big]_{\hat{\mathcal{S}}}\leq \ln \frac{n_1}{n_2}$},\\
    2 & \text{otherwise} \\
    \end{cases}
  \end{align}\; 

} 
\caption{$\mathrm{lda} \circ \mathrm{pca}$ decision rule}
\label{alg:PCLDA}
\end{algorithm}

\subsection{Related works}
\subsubsection{Whitening matrix and spiked covariance}
We first discuss the relationship between Algorithm~\ref{alg:PCLDA} and two other
relevant classifiers, namely the features annealed independence rule (FAIR) of
\citet{Fan2008} and the LDA with CAT scores (CAT-LDA) of \cite{CAT_classify}. FAIR is
the independence rule applied to the features prescreened by the
two-sample $t$ test while CAT-LDA scores decorrelate the $t$ statistics
by first whitening the data using the {\em sample correlation} matrix
$\mathrm{diag}(\hat{\Sigma})^{-1/2} \, \hat{\Sigma} \,
\mathrm{diag}(\hat{\Sigma})^{-1/2}$; CAT-LDA is motivated by the
empirical observation that accounting for correlations is essential 
in the analysis of proteomic and metabolic data.

If there are no correlation or if the correlations are negligible, i.e., when
$\Sigma$ is diagonal or approximately diagonal, then 
Algorithm~\ref{alg:PCLDA} is essentially equivalent to both FAIR and CAT-LDA.
However, when the correlations are not negligible, the performance of
IR classifiers such as FAIR degrades significantly due to mis-specification of the covariance structure
\citep{Fan2012}. Both Algorithm~\ref{alg:PCLDA} and
CAT-LDA apply a whitening transformation before doing LDA but the choice of
whitening matrices are different between the two
procedures. In particular CAT-LDA uses the asymmetric whitening transformation
$\Corr{(\Sigma)}^{-1/2}\diag(\Sigma)^{-1/2}$ where $\Corr{(\Sigma)}$
is the matrix of correlations. The choice of whitening transformation is not
important if $\Sigma$ is known or if we are in the low-dimensional setting where $p \ll n$ for
which consistent estimation of $\Sigma$ is straightforward. It is only
when $p \asymp n$ or $p \gg n$ when the different regularization strategies used in
CAT-LDA and Algorithm~\ref{alg:PCLDA} lead to possibly different behaviors for the
resulting classifiers; see for example the numerical comparisons given in
Section~\ref{Numerical} of the current paper. Finally we view the
theoretical results in Section~\ref{Theory} as not only providing 
justification for Algorithm~\ref{alg:PCLDA} but also serve as examples of theoretical analysis
that can be extended to other classifiers in which dimension
reduction is done prior to performing LDA. Indeed, to the best of
our knowledge, there are no theoretical guarantees for the
error rate for the CAT-LDA classifier in the high-dimensional setting.

Finally we note that our paper is 
not the first work to study high-dimensional
classification under spiked covariance matrices of the form in
Condition~\ref{spiked_covariance}. In particular \cite{spike_LDA} also analyzed
the performance of LDA under spiked covariance structures; \emph{however}, their theoretical results are premised under different settings and framework from ours and most importantly they \emph{do not} show
that their classification rule achieves the Bayes error rate; see Remark~\ref{rem:spiked_compare}
for detailed comparisons of the assumptions and theoretical results in our paper against that of \cite{spike_LDA}.
With a different focus, \cite{rotation} examined the effects of 
spiked eigenvalues when trying to find an
orthogonal transformation $T$ of the discriminant direction $\beta$ so
that $T\beta$ is sparse. More specifically they show that if $\Sigma$ is known then
$T$ can be computed using the eigen-decomposition of the matrix $\Sigma_{\mathrm{rot}}$ where
\begin{align}\label{rot}
    \Sigma_{\mathrm{rot}}=\Sigma+\gamma\Delta\mu\Delta\mu^{\top},\qquad\text{for a given $\gamma>0$}
\end{align}
where $\Delta\mu = \mu_2 - \mu_1$ is the vector of mean differences
between the two classes.
Let $\mathcal{U}_{\mathrm{rot},m}$ be the $p \times m$ matrix whose columns
consist of the orthonormal eigenvectors of $\Sigma_{\mathrm{rot}}$
corresponding to the $m$ largest eigenvalues. When $m=p$ then
$\mathcal{U}_{\mathrm{rot},p}$ diagonalizes
$\Sigma_{\mathrm{rot}}$ and \cite{rotation} showed 
that $\mathcal{U}_{\mathrm{rot},p}$ further sparsifies $\beta$ in that 
$\|\mathcal{U}_{\mathrm{rot},p}^{\top}\,\beta\|_{0}\leq d+1$
whenever $\Sigma$ satisfies
the spiked covariance assumption in
\cref{spiked_covariance}, and hence
it might be beneficial to rotate the data before performing
classification.

When $\Sigma$ is unknown \cite{rotation} propose the following
procedure for estimating $\mathcal{U}_{\mathrm{rot}, m}$ for some
choice of $m \leq \min\{n,p\}$; here $\gamma>0$ is a user-specified parameter.

\begin{algorithm}[H] 
\DontPrintSemicolon 
\SetNoFillComment 

\SetKwProg{Proc}{Algorithm}{}{} 
\Proc{}{ 

  \tcp*[h]{Step 1:} Perform PCA on $\Hat{\Sigma}_{\mathrm{rot}}=\Hat{\Sigma}+\gamma(\bar{X}_{2}-\bar{X}_{1})(\bar{X}_{2}-\bar{X}_{1})^{\top}$ to extract the $m$ largest principal components. Let $\hat{\mathcal{U}}_{\mathrm{rot},m}$ be the resulting $p \times m$ matrix.\; 

  \tcp*[h]{Step 2:} Rotate the training set to $\{\hat{\mathcal{U}}_{\mathrm{rot},m}^{\top}X_{i1},\dots,\hat{\mathcal{U}}_{rot,m}^{\top}X_{in_{i}}\}$ where $\hat{\mathcal{U}}_{\mathrm{rot},m}^{\top}X_{ij}\in\mathbb{R}^{m}$ for $i\in\{1,2\}$ and $j\in \{1,2,\dots,n_i\}$.\;

  \tcp*[h]{Step 3:} Apply some discriminant direction based LDA method such as \cite{Fan2012,TonyCai2011,DSDA_multi} on the rotated data set $\{\hat{\mathcal{U}}_{\mathrm{rot},m}^{\top}X_{ij}\}$.\;

} 
\caption{Rotation as a preprocessing in LDA}
\label{alg:pca_lda_steps}
\end{algorithm}
We will use this pre-processing step in some of our numerical experiments in Section \ref{Numerical}.
Nevertheless we emphasize that the theoretical results of \cite{rotation} assume 
$\Sigma$ is known as they are mainly concerned with the analysis of different approaches for sparsifying
$\beta$, i.e., they explicitly chose not to
address the important issue of how the estimation error for $\hat{\mathcal{U}}_{\mathrm{rot},m}$ impacts the classification accuracy.

\subsubsection{Connection to high-dimensional sparse LDA}
Let $\mathbf{X}$ be the $(n_1 + n_2) \times p$ matrix whose rows are
the $\{X_{ij}\}$
and $\mathcal{Y} \in \mathbb{R}^{n}$ be the vector whose first $n_1$ elements are set to $-(n/n_1)$ and the remaining $n_2$ elements are set to $n/n_2$.
\cite{ZouHui2012} reframed LDA in high-dimension
as the solution to a Lasso-type problem (we have omitted the intercept term for simplicity of presentation)
\begin{align}\label{eq:lasso_LDA}
    \hat{\beta}&=\argmin_{\beta\in \mathbb{R}^{p}}\;\frac{1}{2}\|\mathcal{Y}-\mathbf{X}\beta\|^{2}+ \lambda \|\beta\|_{1}
\end{align}
for some $\lambda > 0$. In a similar spirit to \cref{eq:lasso_LDA}, we might, {\em conceptually}, also reformulate Algorithm~\ref{alg:PCLDA} 
as a (general) Lasso-type problem (\cite{general_lasso}) 
\begin{align}\label{eq:general_lasso}
    \hat{\beta}&=\argmin_{\beta\in \mathbb{R}^{p}}\;\frac{1}{2}\|\mathcal{Y}-\mathbf{X}\beta\|^{2}+ \lambda \|\Psi \beta\|_{1}
\end{align}
for some $\lambda > 0$, where $\Psi = \Sigma^{1/2}$ (once again omitting the intercept for simplicity).  
As $\Sigma$ is invertible, \cref{eq:general_lasso} is equivalent to first solving
\begin{align}\label{eq:general_lasso_PCLDA}
    \hat{\zeta}&=\argmin_{\zeta\in \mathbb{R}^{p}}\;\frac{1}{2}\|\mathcal{Y}-\Tilde{\mathbf{X}}\zeta\|^{2}+ \lambda \|\zeta\|_{1}
\end{align}
where $\Tilde{\mathbf{X}}=\mathbf{X} \Sigma^{-1/2}$, and then setting
$\hat{\beta} = \Sigma^{-1/2} \hat{\zeta}$. 
Comparing \cref{eq:general_lasso,eq:general_lasso_PCLDA} against \cref{eq:lasso_LDA},
one could argue that the main difference between $\mathrm{lda} \circ \mathrm{pca}$ in Algorithm~\ref{alg:PCLDA}
and the lassoed LDA classifier of \cite{ZouHui2012} is due to the
different choice of the predictor variables $\mathbf{X}$ vs $\Tilde{\mathbf{X}}$ in the Lasso regression.
This argument, however, overlooks the important fact that the transformation $\Sigma^{-1/2}$ needs to be
estimated in Algorithm~\ref{alg:PCLDA} while for the general Lassso in \cref{eq:general_lasso},
the matrix $\Psi$ is specified {\em a priori} and thus free
from any estimation error. Indeed, while there have been significant efforts devoted to understanding the solution path of
\cref{eq:general_lasso} in the high-dimensional setting -- see, for example, 
\cite{general_lasso_underdetermined, general_lasso} among others  --
its behavior when $\Psi$ is determined empirically from the data is yet to be theoretically investigated. 

Next we note that, instead of \cref{eq:lasso_LDA}, one can also estimate $\zeta$ via an $\ell_1$ optimization problem similar to that of the Dantzig selector
\citep{Dantzig,TonyCai2011,TonyCai2019}, namely
\begin{align}\label{LPD_PCLDA}
    &\hat{\zeta}=\argmin_{\zeta\in \mathbb{R}^{p}}\Big\{\|\zeta\|_{1} \,\, \text{subject to }\:\|\Hat{\Sigma}^{1/2}\zeta-(\bar{X}_{2}-\bar{X}_{1})\|_{\infty}\leq\lambda\Big\}.
\end{align}
The main difference between Algorithm \ref{alg:PCLDA}  and \cref{LPD_PCLDA} is that Algorithm~\ref{alg:PCLDA} first whitens the feature vectors using an empirical estimate of the covariance matrix followed by
feature selection on the whitened data (see \cref{eq:general_lasso_PCLDA}).  
In contrast, \cref{LPD_PCLDA} attempts to find a $\zeta$ for which its transformation $\hat{\Sigma}^{1/2} \zeta$ is most similar to $\bar{X}_2 - \bar{X}_1$. While the approaches underlying
Algorithm~\ref{alg:PCLDA} and \cref{LPD_PCLDA} are quite similar, we believe that Algorithm~\ref{alg:PCLDA} provides a more direct link between PCA and LDA for high-dimensional classification 
as it is particularly suitable for high-dimensional data generated from a (possibly low-dimenisonal) factor model
(where the covariance matrix will now have a spiked structure with leading eigenvalues that grow with the dimension $p$). %
See \cite{fan2021robust} for further discussion of factor models and their applications to statistics and machine learning. 

Although Algorithm~\ref{alg:PCLDA} and \cref{eq:lasso_LDA} share a conceptual link in optimization formulation, they differ significantly in algorithmic structure, leading to distinct computational profiles. We now compare the computational complexity of \(\mathrm{lda} \circ \mathrm{pca}\) and Lassoed LDA. For our method, the dominant cost comes from computing the top \(\bar{d}\) singular values of the centered data matrix \(\mathbf{X}\), with cost \(O(n p \bar{d})\), where \(\bar{d}\) upper-bounds the rank \(d\); see \cite{random_svd} and \cite{adaptive_fast_pca} for more details. In contrast, Lassoed LDA solves an \(\ell_1\)-penalized least squares problem with complexity \(O(n p T)\), where \(T\) is the number of iterations until convergence. The iteration count \(T\) can be substantial in practice, especially when regularization is weak or \(\mathbf{X}\) is ill-conditioned as often occurs in settings with low-rank or highly correlated features. Consequently, our method provides substantial computational advantages in high-dimensional latent variable models, while maintaining competitive classification accuracy and working in a reduced-dimensional space (see \cref{Numerical}).

Finally we conclude this section by comparing
Algorithm~\ref{alg:PCLDA} with the principal component classifiers
proposed in \cite{bing_lda_regularized, bing_pca_lda}. In particular
both \cite{bing_pca_lda} and \cite{bing_lda_regularized} assumed that
the discriminant direction $\beta$ (and the whitening direction
$\zeta$ as in our work) lies entirely within the column space spanned
by the $d$ leading principal components $\mathcal{U}$.  
However, as noted by
\cite{non_spike_matter}, the principal components associated with
small eigenvalues can be just as important as those associated with
large eigenvalues in real data analysis (see also the simulation
settings for Model~1 and Model~3 in \cref{Numerical}).  By working
with the whitened data, Algorithm~\ref{alg:PCLDA} also takes into
account the {\em non-leading} principal components and furthermore, by
performing feature selection in the whitened space, avoid the need to
specify {\em a priori} the low-dimensional subspaces containing
$\beta$ and/or $\zeta$.

\section{Theoretical properties} \label{Theory}
In this section, we derive the theoretical properties of the $\mathrm{lda} \circ \mathrm{pca}$ classifier in Algorithm~\ref{alg:PCLDA} for
the case where the feature vectors $X$ are sampled from
a mixture of two multivariate Gaussians. Extensions of these results to the
case where $X$ is a mixture of $K \geq 2$ elliptical, but not necessarily multivariate normal, 
distributions are discussed in
Section~\ref{sec:elliptical} and Section~\ref{sec:multi-class}.

If $X \sim \pi_1 \mathcal{N}_{p}(\mu_1, \Sigma) + (1 - \pi_1)
\mathcal{N}_{p}(\mu_2, \Sigma)$ then the Fisher's rule $\Upsilon_{F}$ has
error rate 
\begin{equation}
  \label{eq_RF}
    R_{F}= \pi_1 \Phi\Bigl(-\frac{1}{2} \|\zeta\| + \|\zeta\|^{-1} \ln
  \frac{1 - \pi_1}{\pi_1}\Bigr) + (1 - \pi_1) \Phi\Bigl(-\frac{1}{2} \|\zeta\| - \|\zeta\|^{-1} \ln
  \frac{1 - \pi_1}{\pi_1}\Bigr).
\end{equation}
Here $\Phi$ is the cumulative distribution function for $\mathcal{N}(0,1)$ 
and $\zeta = \Sigma^{-1/2}(\mu_2 - \mu_1)$;   
see e.g., \citet[Section~2.1]{ripley_nn}. 
We note that $R_F$ is the smallest mis-classification
error achievable by any classifier and, furthermore,
is monotone decreasing as $\|\zeta\|$ increases.  
If $\norm{\zeta} \rightarrow 0$ then
$R_{F}\rightarrow \min\{\pi_1, 1 - \pi_1\}$
and $\Upsilon_{F}$ is no better than assigning every data point to the
most prevalent class, while
if $\norm{\zeta} \rightarrow \infty$ then $R_{F}\rightarrow
0$ and $\Upsilon_{F}$ achieves perfect accuracy. 
The cases where $R_{F} \rightarrow 0$ or $R_{F} \rightarrow
\min\{\pi_1, 1 - \pi_1\}$ are, theoretically, uninteresting and hence
in this paper we only focus on the case where $0 < \|\zeta\| <
\infty$.
We therefore make the following assumption. 
\begin{ass}\label{min+card} Let $S_{\zeta}$ denote the set of indices $i$ for which $\zeta_i
  \not = 0$. Also let $\mathcal{C}_{0}>0$, $M > 0$ and
  $\mathcal{C}_{\zeta}>0$ be constants not depending on $p$ such that $s_0 := |\mathcal{S}_{\zeta}| \leq M$
  and
  \begin{align*}
\min_{j\in\mathcal{S}_{\zeta}} |\zeta_{j}| \geq \mathcal{C}_{0}, \qquad
                                \max\bigl\{\|\Sigma^{-1/2}
                                \mu_{1}\|,\|\Sigma^{-1/2} \mu_{2}\| \bigr\} \leq \mathcal{C}_{\zeta}.
\end{align*}
In addition, we assume that the number of spikes $d$ in the spiked covariance model (see \cref{spiked_covariance}) is fixed and does not grow with $p$ or $n$.
\end{ass} 

\begin{remark}\label{rem:sparsity_discussion}
\cref{min+card} implies
$0<\mathcal{C}_{0} \sqrt{s_0} \leq \|\zeta\| \leq 2\mathcal{C}_{\zeta}<\infty$. 
We emphasize that sparsity is imposed on the {\emph{whitened}} direction so that only a few {\emph{transformed}} features contribute to classification outcome; similar assumptions can be found in \cite{nonsparse_regression}. To further explain this condition, we note that sparsity of the
{\em discriminant} direction $\beta = \Sigma^{-1}(\mu_1 - \mu_2)$ also implies that only a small subset of the raw covariates affects the response (as the classification boundary for a feature vector $\bm{x}$ is given by $\bm{1}\{\bm{x}^{\top} \beta > a\}$ for some $a \in \mathbb{R}$).
However, as noted by \cite{non-sparse-gene} and \cite{many-features-classification}, sparsity of $\beta$ is incompatible with many real data application such as
genome-wide gene expression profiling where all genes are believed to play a role in disease markers, or analysis of micro-array data to identify leukemia or colon/prostate cancer.
In contrast sparsity of the whitened direction $\zeta = \Sigma^{-1/2} (\mu_1 - \mu_2)$ still allows for $\beta = \Sigma^{-1/2} \zeta$ to be non-sparse, thereby circumventing the issue
described above. Lastly, throughout our theoretical results and proofs, we assume that the sparsity level $s_{0}$ is fixed and independent of the sample size $n$, though it may be arbitrary. The case where $s_{0}$ grows with $n$ can be handled \emph{mutatis mutandis}. However, it requires more careful control of technical arguments and is therefore left to the interested reader.
\end{remark}

Our theoretical results are large-sample results in which the sample
sizes $n_1$ and $n_2$ for the training data increase as the dimension $p$ increases, and thus our next assumption
specifies the asymptotic relationships between these quantities
and the eigenvalues of $\Sigma$. 
\begin{ass} \label{n_order_Divergent} Let $\sigma > 0$ be
  fixed and suppose that
\begin{align*} 
    \frac{n_{1}}{n_{2}} = \Theta(1), \quad \ln{p}=o(n).
\end{align*}
Furthermore, the spiked eigenvalues
$\lambda_1, \dots, \lambda_d$ of $\Sigma$ satisfy
\begin{equation*}
    \lambda_{k}=\Theta(p),\qquad \text{for all $k\in[d]$}.
\end{equation*}
\end{ass}
If $\Sigma$ satisfies \cref{n_order_Divergent} then $\mathbf{tr}(\Sigma)=\Theta(p)$ and
$\bm{r}(\Sigma) = \mathrm{tr}(\Sigma)/\|\Sigma\| =
\mathcal{O}(1)$. Recall that $\bm{r}(\Sigma)$ is the {\em effective} rank of
$\Sigma$. \cref{n_order_Divergent} also implies that the spiked eigenvalues of $\Sigma$ are
unbounded as $p$ increases and this assumption distinguishes
our theoretical results from existing results in the literature
wherein it is generally assumed that the eigenvalues of $\Sigma$ are
bounded; see \cite{Bickel2014,Fan2008,JunShao2011,TonyCai2019} for a
few examples of results under the bounded eigenvalues assumption. 
While the bounded eigenvalues assumption is prevalent,
it can also be problematic for high-dimensional data as it ignores the strong signals present
in many real data applications; see \cite{SpikedFan} and \cite{bounded_eigenvalue} for further
discussions of this issue. Indeed, a standard heuristic for PCA is to keep the $d$ largest principal components that explains $90\%$ or $95\%$ of the variability in the data, and hence if $p$ is large and $\Sigma$ has bounded eigenvalues then one has to choose $d = \Theta(p)$ which
is inconsistent with the use of PCA as a dimension reduction procedure. 

If $\Sigma$ has a spiked eigenvalue structure as in
\cref{spiked_covariance} then $\mathcal{W}$ is given by
\begin{equation}
  \label{eq:whitening_form}
\mathcal{W} = \Sigma^{-1/2} = \mathcal{U} (\Lambda + \sigma^2 \mathcal{I})^{-1/2} \mathcal{U}^{\top} + \sigma^{-1}
(\mathcal{I} - \mathcal{U} \mathcal{U})^{\top}
\end{equation}
and a natural estimate for $\mathcal{W}$ is
\begin{equation}
  \label{eq:whitening_estimate}
\hat{\mathcal{W}} = \hat{\mathcal{U}} (\hat{\Lambda} +
\hat{\sigma}^2 \mathcal{I})^{-1/2}
\hat{\mathcal{U}}^{\top} + \hat{\sigma}^{-1} (\mathcal{I} -
\hat{\mathcal{U}} \hat{\mathcal{U}}^{\top})
\end{equation}
where $\hat{\Lambda}$ is the diagonal matrix containing the $d$
largest eigenvalues of the {\em pooled} covariance matrix, 
$\hat{\mathcal{U}}$ is the $p \times d$ orthonormal matrix whose columns are the
corresponding eigenvectors, and $\hat{\sigma}^2$ is as defined in \cref{eq:hat_sigma}. We now make the following assumption on
$\mathcal{U}$. 

\begin{ass}[Bounded Coherence] There is a constant $\mathcal{C}_{\mathcal{U}}\geq1$ independent of $n$ and $p$ such that
 \label{Bounded_Coherence}
\begin{equation*}
    \norm{\mathcal{U}}_{2\rightarrow\infty}\leq \frac{\mathcal{C}_{\mathcal{U}} \sqrt{d}}{\sqrt{p}}.
\end{equation*}
\end{ass}
\begin{remark}
The bounded coherence assumption appears frequently
in statistical inference for matrix-valued data. More specifically, as every column in  
$\mathcal{U}$ has $\ell_2$ norm equal to $1$, the rows
$\mathcal{U}$ have, on average, $\ell_2$ norm of order $\sqrt{d/p}$. \cref{Bounded_Coherence} then guarantees that the {\em maximum} $\ell_2$ norm of the rows of $\mathcal{U}$ is also of order
$\sqrt{d/p}$. For more discussions about the bounded coherence
assumption in the context of matrix completion, covariance matrix
estimation, and random matrix theory, see \cite{bounded_coherence,l_infinity,rudelson_vershynin,bloemdal_et_al} among others.  
Finally we note that if $\mathcal{U}$ satisfies the bounded coherence
assumption with constant $\mathcal{C}_{\mathcal{U}}$ then 
\begin{gather*}\Sigma_{ii}\leq\, \frac{\mathcal{C}_{\mathcal{U}}^2 \lambda_{1} d}{p}+\sigma^{2}, \qquad
  \text{for all $i\in [p]$}, 
\end{gather*}
which together with \cref{n_order_Divergent} implies $\max_{i \in [p]} \Sigma_{ii}=\mathcal{O}(1).$ In
summary, \cref{Bounded_Coherence} allows for the spiked eigenvalues
$\lambda_1, \dots, \lambda_d$ of the covariance matrix $\Sigma$ to grow linearly with $p$ while
also guaranteeing that the entries of $\Sigma$
remains bounded, i.e., each variable in $X_{ij}$ has a finite variance.  
\end{remark}

We next state a result on the estimation accuracy of
$\hat{\mathcal{U}}$. This result is a slight extension of an earlier
result in \cite{Cape2019}. More specifically, \cite{Cape2019}
assume $\mathbb{E}[\mathbf{X}] = 0$ and hence the sample covariance matrix is simply $\tfrac{1}{n}
\mathbf{X}^{\top} \mathbf{X}$. In this paper we used the {\em
  pooled} sampled covariance matrix which requires first centering the
feature vectors by the sample means of each class. 
\begin{Thm} \label{sharp two-to-infinity} Let $\mathbf{X}$ be a $n
  \times p$ matrix where the rows $X_1, \dots, X_n$ are i.i.d samples
  from $\pi_1 \mathcal{N}(\mu_1, \Sigma) + (1 - \pi_1)
  \mathcal{N}(\mu_2, \Sigma)$ and $\Sigma$ satisfies \cref{spiked_covariance}, \cref{n_order_Divergent}, and \cref{Bounded_Coherence}. Let $\hat{\mathcal{U}}$ be the matrix of eigenvectors corresponding to the $d$ largest eigenvalues of the pooled sample
  covariance matrix $\hat{\Sigma}$. Then there exists a $d \times d $ orthogonal matrix $\Xi_{\mathcal{U}}$ and a constant $C>0$
such that with probability at least $1-\mathcal{O}(p^{-2})$,
\begin{align}
  \label{eq:bound_U_Uhat}
\|\hat{\mathcal{U}}-\mathcal{U}\Xi_{\mathcal{U}}\|_{2 \rightarrow
  \infty}\leq\mathcal{C}\sqrt{\frac{d^3 \ln{p}}{np}}.
\end{align}
\end{Thm}
Note that, for simplicity, we assume in \cref{sharp two-to-infinity} as well in the subsequent part of this paper that $d$, the number of spiked eigenvalues, is known.
If $d$ is unknown then it can be
{\em consistently} estimated using the ratio between consecutive eigenvalues, similar to the procedures in \cite{ahn_horenstein}. More specifically, from \cref{n_order_Divergent}
and \cref{lem:basic_bounds} we have with high probabliity that $\hat{\lambda}_{k} = \Theta(p)$ for $k \leq d$ and $\hat{\lambda}_{k} = O(\sigma^2) $ for $k \geq d+1$. Here $\hat{\lambda}_k$ are the eigenvalues of $\hat{\Sigma}$.
There thus exists a significant gap between $\hat{\lambda}_{d-1}/\hat{\lambda}_d = O(1)$ and $\hat{\lambda}_d/\hat{\lambda}_{d+1} = \Omega(p)$, and hence, letting $\hat{d}$ be
  the smallest index $k$ for which $\hat{\lambda}_{k}/\hat{\lambda}_{k+1} = \Omega(\ln p)$, we have $\hat{d} = d$ asymptotically almost surely. 

We now analyze the classification accuracy of $\mathrm{lda} \circ \mathrm{pca}$. 
Recall that the main idea behind $\mathrm{lda} \circ \mathrm{pca}$ is that we first construct an 
estimate $\hat{\zeta}$ for the whitened direction $\zeta$ and then
project our {\em whitened} data onto the $d$ largest coordinates, in magnitude, of $\hat{\zeta}$ (see Algorithm~\ref{alg:PCLDA}).
By \cref{eq_RF}, the Bayes error rate $R_F$ is a monotone
decreasing function of $\|\zeta\|$ and hence, to achieve $R_F$ it is only necessary to recover the indices in $S_{\zeta} = \{i \colon \zeta_i \not = 0\}$.
In summary the error rate for $\mathrm{lda} \circ \mathrm{pca}$ converges
to $R_F$ provided we can (1) bound the
error $\hat{\zeta}
- \zeta$ and (2) show that thresholding
$\hat{\zeta}$ perfectly recovers $S_{\zeta}$. 
\begin{Thm}\label{infty norm converge}
Under Assumptions\ref{spiked_covariance}-\ref{Bounded_Coherence}, there exists a constant $C>0$ such that with probability at least $1-\mathcal{O}(p^{-2})$, 
\begin{align}
  \label{eq:zetahat_zeta}
\|\hat{\zeta}-\zeta\|_{\infty}\leq C \sqrt{\frac{\ln{p}}{n}}.
\end{align}
\end{Thm}

\cref{min+card} implies $\zeta_i \not = 0$ if and only if $\zeta_i > \mathcal{C}_{0}$ for some constant $\mathcal{C}_{0}>0$. By Theorem~\ref{infty norm converge},
if $\zeta_i = 0$ then $\hat{\zeta}_i = O((n^{-1} \ln p)^{1/2})$ with high
probability. Let $\Tilde{\zeta}$ be a hard thresholding of $\hat{\zeta}$, namely
\begin{align}\label{hard thresholding}
    \Tilde{\zeta}_{j}= \hat{\zeta}_{j}\mathbbm{1}(|\hat{\zeta}_{j}|>t_{n}),\qquad j \in [p]
\end{align}
where $t_{n}=(n^{-1} \ln{p})^{\alpha}$ for some
$0<\alpha<\frac{1}{2}$. 
Given $\Tilde{\zeta}$, define the active set
$\Tilde{S}=\{j:\Tilde{\zeta}_{j}\neq0\}$. The mis-classification rate
for $\mathrm{lda} \circ \mathrm{pca}$ {\em conditional} on the training data $\{X_{11}, \dots,
X_{1n_1}\}$ and $\{X_{11}, \dots,
X_{1n_1}\}$, is then
\begin{align}
	\hat{R}_{\mathrm{lda} \circ \mathrm{pca}}=\mathbb{P}(\text{label}(\mathbf{Z})\neq\hat{\Upsilon}_{\mathrm{lda} \circ \mathrm{pca}}(\mathbf{Z}) \mid \{X_{11}, \dots, X_{1n_1}\}, \{X_{21}, \dots, X_{2n_2}\})
\end{align}
The following result shows that $\Tilde{S}$ recovers $S_{\zeta}$
exactly and $\mathrm{lda} \circ \mathrm{pca}$
is asymptotically Bayes-optimal.
\begin{Thm}\label{selection consistency and error consistency} Suppose
  that $\mathbf{Z}\sim \pi_1 \mathcal{N}_{p}(\mu_{1},\Sigma)+ (1 -
  \pi_1) \mathcal{N}_{p}(\mu_{2},\Sigma)$ where $\pi_1 \in
  (0,1)$. Suppose \cref{spiked_covariance} through
  \cref{Bounded_Coherence} are satisfied. We then have
\begin{equation} 
  \label{eq:exact_recovery}
    \mathbb{P}(\Tilde{S}\neq\mathcal{S}_{\zeta}) = \mathcal{O}(p^{-2}).
\end{equation}
Furthermore, suppose that $\ln{p}=\smallO(n)$. We then have
\begin{equation} 
  \label{eq:convergence_Rhat}
    |\hat{R}_{\mathrm{lda} \circ \mathrm{pca}}-R_{F}|\longrightarrow 0
\end{equation}
almost surely as $n, p \rightarrow \infty$.
\end{Thm}
Our theoretical framework and results can be extended to elliptical distributions with minimal modifications; see the discussion in Section~\ref{sec:elliptical}. Key remarks on our main results are presented below.
\begin{remark} Theorem~\ref{infty norm converge} and Theorem~\ref{selection
consistency and error consistency} appear, at first blush, quite similar to
the results in \cite{ZouHui2012} for lassoed LDA using the
discriminant direction $\beta = \Sigma^{-1}(\mu_1 - \mu_2)$. However
the underlying assumptions behind these results are very different. In
particular lassoed LDA relies on irrepresentable condition
analogous to those in \cite{irrepresentable} to achieve consistent variable
selection, and thus the discriminative variables cannot
	be highly correlated with the remaining (irrelevant) variables. In contrast
\cref{selection consistency and error consistency} do not assume an irrepresentable condition and 
furthermore \cref{spiked_covariance} only
assume boundedness of the non-spiked eigenvalues but allows for diverging spiked eigenvalues.
\end{remark}

\begin{remark}\label{rem:spiked_compare}
We now compare our results and settings with those in \cite{spike_LDA}. Firstly, \cite{spike_LDA} assumed: 
(1) proportional growth rate $p/n \rightarrow c$ for some \emph{finite} constant $c > 0$, (2) \emph{distinct} spiked eigenvalues 
$\lambda_{1}>\lambda_2 > \dots>\lambda_{d}$ and lastly (3) $\|\Sigma\|=\mathcal{O}(1)$. In contrast, our work is derived under a 
more challenging set-up wherein (1) the dimension $p$ and the number of training samples $n$ satisfy 
$n^{-1} \ln{p} \rightarrow 0$ (\cref{n_order_Divergent}), (2) some or even all of the eigenvalues $\lambda_{1}\geq\dots\geq\lambda_{d} > 0$ can be equal
(\cref{spiked_covariance}) and (3) the leading eigenvalues $\{\lambda_k\}_{k=1}^{d}$ can diverge with $n$ (\cref{n_order_Divergent}). 
We emphasize that $n^{-1} \ln p \rightarrow 0$ allows for $p = n^{\gamma}$ for any $\gamma > 1$, so that $p/n \rightarrow \infty$.  

The assumptions used in \cite{spike_LDA} are because their theoretical analysis rely heavily on existing results in random matrix theory as presented in 
\citet{condition_spiked,shrink_spiked}, which generally require $p/n < \infty$ as well as bounded eigenvalues. In contrast, if $p/n\rightarrow \infty$, 
then \cite{condition_spiked} shows that the best estimate of the true covariance matrix is typically a diagonal matrix (see also \cite{Bickel2014}), which 
might make the results in \cite{spike_LDA} sub-optimal for the high-dimensional regime $p/n \rightarrow \infty$ considered in our paper. 

Finally, \cite{spike_LDA} \emph{did not} guarantee that their classification rule is Bayes-optimal. Rather, 
they only show that the its mis-classification rate converge to some expression given in Theorem~3 of \cite{spike_LDA}, and 
is strictly larger than Bayes error when $p/n$ increases. 
\cite{spike_QDA} subsequently extended the results in \cite{spike_LDA} to the case of heterogeneous covariance matrices but the resulting classifier 
is once again not Bayes-optimal. In contrast, Theorem~\ref{selection consistency and error consistency} and Theorem~\ref{thm:infty_norm_QDA}
showed that the error rate for $\mathrm{lda} \circ \mathrm{pca}$ and $\mathrm{qda} \circ \mathrm{pca}$ (Quadratic Discriminant Analysis, QDA)
converge to the Bayes error rate in the case of equal and unequal (class-conditional) covariance matrices, respectively. These 
results are, to the best of our knowledge, are among the first to show Bayes consistency when combining 
PCA with LDA/QDA for high-dimensional classification under divergent spikes.
A key reason for this improvement lies in two elements: (1) the assumption of sparsity in the signal vector $\zeta$, which mitigates estimation error high-dimensional means and covariances (motivated by \cite{Tibshirani2002, Fan2008}) and (2) the use of the entrywise matrix perturbation bounds, providing precise control over the estimated eigenspace $\hat{\mathcal{U}}$ and the resulting whitened direction $\hat{\zeta}$.
\end{remark}

\section{Numerical results} \label{Numerical}
We now present simulation results and real data analysis
for $\mathrm{lda} \circ \mathrm{pca}$. Comparisons will be made against the nearest shrunken centroids method (NSC) of
\citet{Tibshirani2002}, sparse linear discriminant analysis (SLDA) of
\citet{JunShao2011}, direct sparse discriminant analysis (DSDA)
of \citet{ZouHui2012}, adaptive linear discriminant analysis
(AdaLDA) of \citet{TonyCai2019}, and LDA rule with
CAT scores (CAT-LDA) of \citet{CAT_classify}; note that
CAT-LDA uses a different sphering transformation compared to $\mathrm{lda} \circ \mathrm{pca}$. There are three other
commonly used classifiers that are not included in our comparisons,
namely the naive
Bayes rule, the linear programming discriminant (LPD)
\citep{TonyCai2011} and the regularized optimal affine
discriminant (ROAD)\citep{Fan2012}. We omit these classifiers because
(1) the naive Bayes rule is a special case of the NSC rule
without soft thresholding of the mean vectors (2) \citet{ZouHui2013}
showed that the ROAD and DSDA classifiers are equivalent and (3) \citet{TonyCai2019} showed that the AdaLDA rule is a
refinement of the LPD rule, i.e., compared to the LPD rule, the AdaLDA rule allows for
``heteroscedastic constraints'' and requires no tuning parameters.

Implementations of the DSDA and CAT-LDA rules are based on
the \href{https://cran.r-project.org/web/packages/TULIP/index.html}{TULIP}
and \href{https://cran.r-project.org/web/packages/sda/}{sda} library
in $\mathtt{R}$ while the implementation of the AdaLDA rule is based on $\mathtt{Matlab}$ codes provided in
\href{https://github.com/linjunz/ADAM}{ADAM} github repository. We also consider the rotation pre-processing
step of \citet{rotation}, which yields the transformed data
$\{\mathcal{U}_{\mathrm{rot},n}^{\top}X_{11},\dots,\mathcal{U}_{\mathrm{rot},n}^{\top}X_{1n_{1}}\}$
and
$\{\mathcal{U}_{\mathrm{rot},n}^{\top}X_{21},\dots,\mathcal{U}_{\mathrm{rot},n}^{\top}X_{2n_{2}}\}$ based on $\mathtt{Matlab}$ codes from 
\href{http://bicmr.pku.edu.cn/~dongbin/Publications.html}{HDRotation} where we set $\gamma=0.25$ in \cref{rot}. 
These transformed data are then used as input to the DSDA and AdaLDA rules; we
denote the resulting classifiers as DSDA(rot) and AdaLDA(rot). 
\subsection{Simulated examples}\label{Simulation}

We consider three different simulation settings. For each setting, the number of features
is set to $p = 800$ and we generate $n_{1}=n_{2}=100$ data points
from each class for the training data and also generate $n_1 = n_2 = 100$ 
data points from each class for the testing data. The classification
accuracy of the classifiers in each simulation setting are computed based on $200$ Monte Carlo replications. 
%
The mean vectors of the two classes are $\mu_{1} =
\mathbf{0}_{800}$ and $\mu_{2} =
(\bm{1}_{10},\mathbf{0}_{790})$, i.e., the vector
$\mu_2$ contains $10$ non-zero entries with values all equal to $1$.
We consider the following models for the covariance matrix
$\Sigma$. These models were considered previously in
\citet{Fan2012,ZouHui2012,TonyCai2019,TonyCai2011}, among others. Throughout this sub-section, the number of selected features refers to the dimensionality of the representation on which each method operates. In particular, $\mathrm{lda} \circ \mathrm{pca}$ performs feature selection in the whitened feature space, whereas other methods select features directly from the raw feature space. See \cref{sec: sensitive} for a detailed study in which the sparsity levels of the mean difference $\mu_{2}-\mu_{1}$, the discriminant direction $\beta$, and the whitened direction $\zeta$ are matched.

%


\begin{enumerate}
\item \emph{Model 1} (\emph{equal correlation}): Here
  $\Sigma=(\sigma_{ij})_{p \times p} = \rho \bm{1} \bm{1}^{\top} + (1
  - \rho) \mathcal{I}_{p}$, i.e., $\sigma_{ij}=\rho$ for $i\neq j$ and
  $\sigma_{ij}=1$ for $i=j$. With this covariance structure $\Sigma$, the discriminant direction $\beta =
  \Sigma^{-1}(\mu_1 - \mu_0)$ and whitened direction $\zeta =
  \Sigma^{-1/2}(\mu_1 - \mu_0)$ are non-sparse (all entries of $\beta$ and $\zeta$ are non-zero). 
\item \emph{Model 2} (\emph{block diagonal with equal correlation}) Here
  $\Sigma=(\sigma_{ij})_{p \times p}$ is assumed to be a block diagonal
  matrix with two blocks of size $20 \times 20$ and $(p - 20) \times
  (p - 20)$. Both diagonal blocks are also of the form $\rho \bm{1} \bm{1}^{\top} + (1
  - \rho) \mathcal{I}$ where the correlation $\rho$ is the
  same for both blocks. The discriminant direction $\beta$ and
  whitened direction $\zeta$ are sparse in this model, with $\beta$ and $\zeta$ both having
  $20$ non-zero entries.  
\item \emph{Model 3} (\emph{random correlation}) Here
$\Sigma=\mathcal{L}\mathcal{L}^{\top}+c_{\mathcal{L}}\mathcal{I}_{p}$
where $\mathcal{L} \in \mathbb{R}^{p\times 10}$ with $\mathcal{L}_{ij}$
generated from $\mathcal{N}(0,1) $ and $c_{\mathcal{L}}=\min_{i\in
[p]}[\mathcal{L}\mathcal{L}^{\top}]_{ii}$. Note that we generate a new $\mathcal{L}$ for every Monte Carlo
replicate. For further comparison, we also consider $\mathcal{L}_{ij}$
generated from the uniform distribution on $[-1,1]$ and the Student's
t-distribution with $5$ degrees of freedom. The discriminant direction
$\beta$ and $\zeta$ are generally non-sparse in this model, i.e., all
of their entries are non-zero. 
\end{enumerate}
The tuning parameters for each classifier are chosen using five-fold cross validation (CV). 
In particular, SLDA \citep{JunShao2011} requires two tuning
parameters, one being the number of non-zero entries in
$\mu_1 - \mu_2$ and the other being the number of non-zero
entries in $\Sigma$. For
simplicity we shall assume that the sparsity of $\mu_1 - \mu_2$ is known when implementing
SLDA and thus the only tuning parameter required is the number of non-zero entries in 
$\Sigma$; following \citet{TonyCai2019}, this tuning parameter is selected from the set of values $\{\sqrt{n^{-1} \ln{p}},1.5\sqrt{n^{-1}
  \ln{p}},\dots, 5\sqrt{n^{-1} \ln{p}}\}$ using five-fold CV.
The $\mathrm{lda} \circ \mathrm{pca}$ classifier in Algorithm~\ref{alg:PCLDA} also requires
two tuning parameters, namely (1) the number of spikes $d$ in the
estimation of $\Sigma$ and (2) the sparsity level $s$ in
$\zeta$. 
We chose $d$ to account for at least $90\%$
of the total variability in the data, i.e., $d$ is the smallest integer of $k$ satisfying $(\sum_{i=1}^{k}
\hat{\lambda}_i)/\mathbf{tr}(\hat{\Sigma}) \geq 0.9$; here
$\hat{\lambda}_1 \geq \hat{\lambda}_2 \geq \dots$ are the eigenvalues
of the pooled sample covariance matrix $\hat{\Sigma}$. We acknowledge that the choice of 
$d$ can significantly impact performance; therefore, a sensitivity analysis is provided in \cref{sec: sensitive}. For each Monte Carlo replication, the training data are randomly split into five folds for CV, with new splits generated each time. The sparsity
level $s$ is chosen from $s \in \{1,2,\dots, 30\}$ to minimize the average CV misclassification error. In the case where multiple values of $s$ yield the same minimum error, we choose the smallest such $s$ to promote a more parsimonious model. 
Similarly, the number of features
in CAT-LDA is selected among the top $\{1,2,\dots,30\}$ features ranked
by CAT scores using five-fold CV. The choice of the upper limit $30$ is motivated by the common sparsity condition $s_{0} =\smallO((n/\ln{p})^{\tau}),\;\text{for some $\tau>0$}$, for instance, $\tau = 1/2$ in \cite{TonyCai2011}. We deliberately choose $n/\ln{p}\;(\approx 30)$ as a relaxed upper bound to broaden the grid and ensure adequate coverage.

For data generated according to Model 1, Table \ref{tab:err model1}
and Table \ref{tab:nonzero model1}
show that $\mathrm{lda} \circ \mathrm{pca}$ achieves the highest accuracy while using
only a small number of features compared to the other classifiers. 
Note that although CAT-LDA also apply a whitening transformation
before performing LDA, the accuracy of CAT-LDA is much worse compared
to that of $\mathrm{lda} \circ \mathrm{pca}$. The DSDA and DSDA(rot) classifiers have slightly better accuracies
compared to those for the AdaLDA and AdaLDA(rot) classifiers; recall
that the DSDA(rot) and AdaLDA(rot) classifiers first applied the
rotation pre-processing step of \citet{rotation} before running DSDA
and AdaLDA on the transformed data. The NSC classifier has the largest
mis-classification error; this is a consequence of the NSC rule ignoring 
the correlation structure in $\Sigma$. Finally, the oracle classifier
correspond to the LDA rule where $\pi_1, \mu_0, \mu_1$ and $\Sigma$
are known, and thus its error rate is the
Bayes error rate from \cref{eq_RF}. 

\begin{table}[t]
\caption{Mis-classification rate ($\%$) with standard deviations ($\%$) in parentheses for equal correlation setting (model 1), 
    based on $200$ independent Monte Carlo replicates.}
\label{tab:err model1}
\begin{center}
\begin{tabular}{l|ccccc}

$\rho$ & 0.50 & 0.60 & 0.70 & 0.80 & 0.90 
\\ \hline \\
  Oracle & 1.37 (0.87)& 0.61 (0.53)& 0.19 (0.30)& 0.19 (0.30)& 0.00 (0.00)\\ 
  $\mathrm{lda} \circ \mathrm{pca}$ & 1.74 (1.00)& 1.00 (0.82)& 0.55 (0.67)& 0.55 (0.67)& 0.22 (0.39)\\ 
 CAT-LDA & 7.18 (2.22) & 4.70 (1.77) & 2.60 (1.34) & 1.10 (0.88) & 0.38 (0.56) \\ 
  DSDA& 3.27 (1.47)& 1.96 (1.08)& 0.74 (0.62)& 0.76 (0.65)& 0.00 (0.04)\\ 
  DSDA(rot)& 6.31 (1.83)& 3.26 (1.27)& 1.16 (0.83)& 0.30 (0.41)& 0.01 (0.06)\\ 
  AdaLDA&4.15 (1.61)&2.62 (1.24)&1.22 (0.91)&0.31 (0.42)&0.00 (0.04)\\
  AdaLDA(rot)&7.04 (2.07)&3.70 (1.26)&1.19 (0.79)&0.13 (0.26)&0.00 (0.00)\\
  SLDA& 17.97 (3.53)& 14.46 (2.99)& 11.14 (2.43)& 11.14 (2.43)& 1.64 (0.92)\\
  NSC& 20.38 (8.53)& 22.60 (8.38)& 24.64 (8.36)& 24.64 (8.36)& 29.26 (8.04)\\
\end{tabular}
\end{center}
\end{table}

\begin{table}[t]
\caption{\label{tab:nonzero model1} Average number of nonzero coefficients with standard deviations in parentheses for equal correlation setting (model 1), 
based on $200$ independent Monte Carlo replicates.
}
\begin{center}
\begin{tabular}{l|ccccc}
  $\rho$ & 0.50 & 0.60 & 0.70 & 0.80 & 0.90
  \\ \hline \\
  $\mathrm{lda} \circ \mathrm{pca}$ & 12.04 (4.53)& 11.31 (4.10)& 9.52 (4.01)& 9.52 (4.01)& 3.68 (0.98)\\
  CAT-LDA & 24.85 (3.74) & 24.70 (3.58) & 24.60 (3.97) & 22.29 (3.93) & 15.45 (2.39) \\ 
  DSDA & 96.20 (31.06)& 106.42 (32.57)& 117.15 (30.52)& 117.05 (29.97)& 96.48 (8.66)\\
  DSDA(rot) & 33.78 (32.71)& 36.88 (38.13)& 57.23 (61.02)& 147.57 (66.97)& 176.29 (5.08)\\ 
  AdaLDA&46.54 (5.95)&45.84 (5.25)&46.72 (6.41)&47.86 (5.88)&48.65 (5.02)\\
  AdaLDA(rot)&5.68 (1.82)&5.57 (1.84)&5.86 (1.82)&7.83 (2.54)&19.85 (9.96)\\
  SLDA & 728.84 (226.63)& 788.16 (95.69)& 799.96 (0.21)& 799.96 (0.21)& 799.96 (0.21)\\

\end{tabular}
\end{center}
\end{table}

\begin{remark}
\label{rem:strong_faint_signal}
Recall that for Model $1$, all entries of the discriminant direction
$\beta$ and the whitened direction $\zeta$
are non-zero. 
These entries however can be classified
into those representing strong signals vs weak signals based on
their magnitudes; see Table \ref{tab:stron and weak model1} in the supplementary. The strong signals appear in the
first $10$ entries of $\beta$ (similarly $\zeta$) and the
remaining $p - 10$ entries of $\beta$ (similarly $\zeta$)
correspond to the weak signals. For example, if $\rho = 0.5$ then the
first $10$ entries of $\beta$ are all equal to $1.98$ and the remaining $p
= 10$ entries are all equal to $-0.02$. From Table~\ref{tab:nonzero model1} we
see that $\mathrm{lda} \circ \mathrm{pca}$ kept all of the coordinates corresponding to
the strong signals and only added a few coordinates corresponding to
the weak signals; this explains the small number of features used in
$\mathrm{lda} \circ \mathrm{pca}$. In contrast, the DSDA and SLDA classifiers include a 
large number of (noisy) features with weak signals.
\end{remark}

For Model 2, Table \ref{tab: err model2} shows that SLDA performs
slightly better than $\mathrm{lda} \circ \mathrm{pca}$ and DSDA. However, from
Table~\ref{tab: nonzero model2}, we see that SLDA
also selects a much larger number of features compared to both $\mathrm{lda} \circ \mathrm{pca}$ and DSDA. Recall that, for model $2$, both $\beta$ and $\zeta$ contains exactly $20$ non-zero
entries and hence SLDA is selecting a large number of
extraneous, non-informative features; a similar, albeit much
less severe, phenomenon is observed for DSDA. 
Table~\ref{tab: err model2} and Table~\ref{tab:
  nonzero model2} shows that $\mathrm{lda} \circ \mathrm{pca}$, CAT-LDA, and AdaLDA have
comparable accuracy with a similar number of selected features. The
NSC rule once again has the largest
mis-classification error due to it ignoring the correlation structure
in $\Sigma$. Finally we see that the rotation pre-processing step
described in \citet{rotation} lead to a substantial loss in accuracy
for the DSDA and AdaLDA classifiers; to understand why this happens, we 
need to extend the theoretical analysis in \citet{rotation} (which
assume that $\Sigma$ and $\mathcal{U}_{\mathrm{rot},p}$ are known) to the 
setting where $\Sigma$ and $\mathcal{U}_{\mathrm{rot},p}$ have to be
estimated. We leave this investigation for future work. 

Finally for Model~3, Table~\ref{tab: err model5} and Table~\ref{tab:
  random correlation} show that $\mathrm{lda} \circ \mathrm{pca}$ has both the highest
accuracy as well as the smallest number of selected features among all
the considered classifiers; DSDA has a slightly worse
accuracy and also selected a much larger number of features, when
compared to $\mathrm{lda} \circ \mathrm{pca}$. SLDA now has the worst accuracy and
also selects almost all $p = 800$ features, and this is a consequence
of $\Sigma$ being a dense matrix. 
The mis-classification rate for AdaLDA and AdaLDA(rot) are also quite large, and
we surmise that this is due to the numerical instability when
solving the linear programming problem in AdaLDA. Indeed, the
condition numbers for $\Sigma$ can be quite large; see
Table~\ref{tab:condition number} in the supplementary for summary statistics of
these condition numbers using the same $200$ Monte Carlo replicates
as that for generating Table~\ref{tab: err model5}. Following
the suggestion in \citet{TonyCai2019}, we replace
the sample covariance matrix $\hat{\Sigma}$ used in the optimization
problem for AdaLDA and AdaLDA(rot) (see Eq.~(8) and Eq.~(9) in \citet{TonyCai2019}) with 
$\Tilde{\Sigma}=\hat{\Sigma}+\sqrt{n^{-1}\ln{p}}\:\mathcal{I}_{p}$; 
the resulting classifiers are denoted as AdaLDA(reg) and AdaLDA(rot +
reg), respectively. Table~\ref{tab: err model5} however shows that
using $\Tilde{\Sigma}$ only leads to a minimal increase in
accuracy. Finally we observe that the rotation pre-processing step
once again leads to a
substantial loss in accuracy for the DSDA and AdaLDA classifiers.

\begin{remark}
\label{rem:non_sparse}
While our theoretical results primarily focus on fixed sparsity (see Remark~\ref{rem:sparsity_discussion} for more on growing sparsity), we deliberately include dense examples with $s_{0}=p$, such as Model~1 and Model~3, to demonstrate robustness beyond the sparse regime. In the dense models, full recovery of $\zeta$ or $\beta$ is not anticipated but $\mathrm{lda} \circ \mathrm{pca}$ still delivers strong classification performance due to effective decorrelation. In particular, for Model~1, our method’s performance is very close to the oracle across varying $\rho$.
\end{remark}

\begin{table}[t]
\caption{\label{tab: err model2}
    Mis-classification rate ($\%$) with standard deviations ($\%$) in parentheses for block diagonal setting (model 2), based on $200$ independent Monte Carlo replicates.
    }
\begin{center}
\begin{tabular}{l|ccccc}

  $\rho$ & 0.50 & 0.60 & 0.70 & 0.80 & 0.90
    \\ \hline \\
  Oracle & 5.30 (1.71)& 5.21 (1.52)& 5.34 (1.64)& 5.48 (1.62)& 5.40 (1.74)\\
   $\mathrm{lda} \circ \mathrm{pca}$ & 9.08 (2.71)& 8.61 (2.59)& 8.30 (2.55)& 8.26 (2.45)& 8.12 (2.58)\\
CAT-LDA & 10.08 (3.00) & 9.36 (2.57) & 9.31 (2.50) & 9.25 (2.59) & 9.21 (2.89) \\ 
  DSDA& 9.63 (2.61)& 9.62 (2.64)& 9.50 (2.66)& 9.58 (2.67)& 9.47 (2.42)\\
  DSDA(rot)& 20.64 (4.16)& 20.52 (3.99)& 20.75 (4.18)& 21.02 (4.44)& 20.90 (4.40)\\
  AdaLDA&12.70 (2.99)&13.06 (3.39)&13.88 (3.45)&13.75 (3.48)&14.68 (3.86)\\
  AdaLDA(rot)&23.91 (3.79)&23.81 (3.72)&24.33 (4.10)&23.89 (3.92)&24.12 (3.94)\\
  SLDA & 6.99 (2.89)& 6.66 (1.92)& 6.71 (2.17)& 6.62 (1.97)& 6.56 (1.90)\\
  NSC & 25.14 (3.11)& 26.47 (3.31)& 28.61 (3.35)& 29.59 (3.24)& 30.52 (3.82)\\

\end{tabular}
\end{center}
\end{table}

\begin{table}
 \caption{\label{tab: nonzero model2}
    Average number of nonzero coefficients with standard deviations in parentheses for block diagonal setting (model 2), 
    based on $200$ independent Monte Carlo replicates.
    }
\begin{center}
    \begin{tabular}{l|ccccc}
  
  $\rho$ & 0.50 & 0.60 & 0.70 & 0.80 & 0.90 
  \\ \hline \\
  $\mathrm{lda} \circ \mathrm{pca}$ & 20.19 (4.87)& 20.48 (5.06)& 20.48 (4.88)& 20.96 (4.96)& 20.58 (4.44)\\ 
  CAT-LDA & 20.02 (4.95) & 19.86 (4.31) & 20.05 (4.62) & 20.32 (4.21) & 20.33 (4.54) \\ 
  DSDA & 50.28 (26.61)& 50.74 (28.92)& 47.48 (25.77)& 43.66 (22.42)& 45.15 (23.78)\\
  DSDA(rot) & 51.02 (39.23)& 46.13 (29.09)& 51.93 (36.08)& 45.56 (31.15)& 49.62 (33.50)\\
  AdaLDA&19.45 (2.37)&18.53 (2.23)&17.93 (2.27)&17.40 (2.20)&16.85 (2.07)\\
  AdaLDA(rot)&14.40 (3.14)&14.81 (3.53)&15.16 (3.40)&15.69 (3.99)&14.97 (3.84)\\
  SLDA & 405.98 (390.84)& 320.17 (380.33)& 210.99 (336.12)& 242.19 (352.81)& 172.00 (309.62)\\ 
   \hline
\end{tabular}
\end{center}

\end{table}

\begin{table}[t]
\caption{\label{tab: err model5}Mis-classification rate ($\%$) with standard deviations ($\%$) in parentheses for random correlation setting (model 3), based on $200$ independent Monte Carlo replicates.
}
\begin{center}
\begin{tabular}{l| c c c} 
Method & $\mathcal{U}(-1,1)$ &$\mathcal{N}(0,1)$& $\mathbf{t}_{5}$ 
\\ \hline \\
$\mathrm{lda} \circ \mathrm{pca}$ &5.07 (2.40)& 12.39 (4.17)&13.72 (5.00) \\
CAT-LDA &11.93 (4.04)& 23.67 (6.13)&25.68 (6.82)\\
DSDA&7.32 (3.05)&16.92 (5.21)&17.90 (5.49)\\
DSDA(rot)&18.51 (6.37)&31.81 (7.11)&33.65 (7.28)\\
AdaLDA&17.94 (6.03)&44.55 (7.12)&47.89 (5.10)\\
AdaLDA(rot)&23.00 (7.49)&39.29 (6.11)&42.52 (6.28)\\
AdaLDA(reg)&29.48 (7.58)&47.49 (4.52)&48.96 (4.04)\\
AdaLDA(reg+rot)&21.23 (7.43)&38.93 (6.60)&42.52 (6.28)\\
SLDA&50.10 (4.40)&49.98 (4.25)&50.09 (3.70)\\
NSC&30.30 (6.28)&44.55 (5.16)&46.70 (4.49)\\

\end{tabular}
\end{center}
\end{table}

\begin{table}
\caption{\label{tab: random correlation}Model size with standard deviations in parentheses for random correlation setting (model 3), 
based on $200$ independent Monte Carlo replicates.
}
\begin{center}
\begin{tabular}{l| c c c} 
Method & $\mathcal{U}(-1,1)$ &$\mathcal{N}(0,1)$& $\mathbf{t}_{5}$ 
\\ \hline \\
$\mathrm{lda} \circ \mathrm{pca}$ &11.93 (4.10)& 11.48 (3.88)&11.37 (3.92)\\
CAT-LDA &21.19 (6.51)& 20.64 (6.88)&20.75 (6.90)\\
DSDA&66.00 (31.92)&65.63 (34.45)&64.91 (30.51)\\
DSDA(rot)&51.08 (34.82)&71.94 (41.64)&77.56 (49.29)\\
AdaLDA&15.99 (2.32)&10.46 (1.97) &8.51 (2.22)\\
AdaLDA(rot)&14.62 (4.61)&19.08 (3.93) &16.26 (3.43)\\
AdaLDA(reg)&11.38 (1.77)&8.50 (2.00)& 7.12 (1.96)\\
AdaLDA(reg+rot)&9.86 (4.71)&15.99 (3.63)& 16.26 (3.43)\\
SLDA&799.51 (1.31)&800.00 (0.07)&800.00 (0.00)\\
NSC&12.21 (8.47)&55.38 (131.65)& 102.89 (192.83)\\

\end{tabular}
\end{center}
\end{table}

\subsection{Real data analysis}
We now assess the performance of $\mathrm{lda} \circ \mathrm{pca}$ on two
gene expression data sets for leukaemia and lung cancer.  The
leukaemia dataset \citep{leukaemia1999} includes $p=7128$ gene
measurements on $72$ patients with either acute lymphoblastic leukemia
(ALL) or acute myeloid leukemia (AML) and we wish to classify patients
into either ALL or AML based on their gene expressions.  The training
set consists of the gene expressions for $n_{1}=27$ patients with
ALL and $n_{2}=11$ patients with
AML while the test set contains the gene expression data for $20$ patients
with ALL and $14$ patients with AML. The lung
cancer data was originally analyzed in \citet{lungcancer2002} and we use a
version of the data wherein the predictor
variables with low variances are removed; see \citet{lungcancer}. The
resulting data set contains tumor tissues with $p=1577$ features 
collected from patients with adenocarcinoma (AD) or malignant pleural
mesothelioma (MPM). According to \cite{lungcancer2002}, MPM is highly
lethal but distinguishing between MPM and AD is quite challenging in
both clinical and pathological settings. The training set consists of $n_1
= 120$ gene expressions for patients with AD and $n_2 = 25$ gene
expressions for patients with MPM, and the test set consists of gene
expressions for $30$ patients with AD and $6$ patients with MPM. 
Table \ref{tab:leukaemia} and Table \ref{tab: lung
cancer} presents the classification accuracy and number of selected features for various
classifiers when applied to the leukemia and lung cancer data,
respectively. The hyperparameters for these classifiers are selected using
leave-one-out CV. 
Table \ref{tab:leukaemia} and \ref{tab: lung cancer}
indicate that the performance of $\mathrm{lda} \circ \mathrm{pca}$ is competitive with existing state-of-the-art classifiers 
while also operating on a substantially lower-dimensional representation.

\begin{table} [t]
\caption{\label{tab:leukaemia}Mis-classification rate and model size of various methods
  for the leukaemia data}
\begin{center}
\begin{tabular}{l|c c c} 
Method & Training Error & Test Error & Model Size 
\\ \hline \\
$\mathrm{lda} \circ \mathrm{pca}$ & 0/38 &1/34& 12 \\
DSDA&0/38&1/34&36\\
AdaLDA (reg)&0/38&1/34&18\\
NSC&1/38&3/34&24\\

\end{tabular}
\end{center}
\end{table}
\begin{table}[t]
\caption{\label{tab: lung cancer} Mis-classification rate and model size of various methods for the lung cancer data}
\centering

\begin{center}
\begin{tabular}{l|c c c} 
Method & Training Error & Test Error & Model Size 
\\ \hline \\
$\mathrm{lda} \circ \mathrm{pca}$ & 0/145 &0/36& 20 \\
DSDA&1/145&0/36&25\\
AdaLDA (reg)&0/145&0/36&388\\
NSC&1/145&0/36&1206\\
\end{tabular}
\end{center}
\end{table}

\section{Extensions of $\mathrm{lda} \circ \mathrm{pca}$}
\label{Extension}

\subsection{Multi-class classification}
\label{sec:multi-class}
Suppose we are given training data from $K \geq 3$
classes where the feature vectors for each class are iid samples from a $p$-dimensional multivariate normal
distribution, i.e., 
\begin{align*}
    X_{i1},\dots,X_{i n_{i}}\sim\mathcal{N}_{p}(\mu_{i},\Sigma),\qquad i\in[K].
\end{align*}
Here $n_i$ denote the number of training data points from class $i \in
[K]$. 
The testing data point $\mathbf{Z}$ is drawn from a mixture of 
$K$ multivariate normal distributions namely, $\mathbf{Z} \sim \sum_{i=1}^{K}\pi_i
\mathcal{N}_{p}(\mu_{i},\Sigma)$ with $\pi_{i}\geq0$ and
$\sum_{i=1}^{K}\pi_i=1$. Note that, for ease of exposition, the $\{X_{ij}\}$ are
assumed to be multivariate normals but the subsequent results also hold when $\{X_{ij}\}$ are
elliptically distributed as in Section~\ref{sec:elliptical}.

Now define, for $2 \leq i \leq K$, the whitened direction
$\zeta^{(i)}$ and the
whitened indices $\mathcal{S}_i$ via
\begin{align*}
    \zeta^{(i)}=\mathcal{W}(\mu_{i}-\mu_{1}), \qquad
  \mathcal{S}_{i}=\{j:\zeta^{(i)}_{j} \neq0\}.
\end{align*} 
Define the global whitened set as $\mathcal{S}_{\zeta}=\mathcal{S}_{2} \cup \mathcal{S}_{3} \cup \dots \cup \mathcal{S}_{K}$. The indices
in $\mathcal{S}_{\zeta}$ are the important variables for feature selection. The extension of $\mathrm{lda} \circ \mathrm{pca}$ to $K \geq 3$ classes is then given in Algorithm~\ref{alg:multi_PCLDA} below, and theoretical results are presented in the supplementary. 

\begin{algorithm}[H] 
\DontPrintSemicolon 
\SetNoFillComment 

\KwInput{$\Bar{X}_{i},\: i\in[K]$, $\hat{\Sigma}$ and the test sample $\mathbf{Z}$}
\KwOutput{$\hat{\Upsilon}_{\mathrm{lda} \circ \mathrm{pca}}(\mathbf{Z})$}


\SetKwProg{Proc}{Algorithm}{}{} 
\Proc{}{ 

  \tcp*[h]{Step 1:} Perform PCA on the feature vectors for the training
  data (standard PCA approach). Extract the $d$ largest principal components to obtain $\Hat{\mathcal{W}}$.\;

  \tcp*[h]{Step 2:} For $2 \leq i \leq K$, let $\Tilde{X}_{i}=\hat{\mathcal{W}}\Bar{X}_{i}$ and 
  $\hat{\zeta}^{(i)} = \Tilde{X}_{i}-\Tilde{X}_{1}$. Also let
  $\Hat{\mathcal{S}}_{i}$ be the set of indices corresponding to the
  $s_i$ largest elements of $\Hat{\zeta}^{(i)}$ in absolute
  values. The value of $s_i$ is, in general, a user-specified or
  tuning parameter. Nevertheless, under certain conditions, we can
  also estimate $s_i$; see \cref{eq:threshold_Kclass} in the supplementary.\;
  
  \tcp*[h]{Step 3:} Given the test data point $\mathbf{Z}$, let
  $\Tilde{\mathbf{Z}}=\hat{\mathcal{W}}\mathbf{Z}$ and
  denote by $\Hat{\zeta}^{(i)}_{\Hat{\mathcal{S}}_{i}}$ the vector
  obtained from $\hat{\zeta}^{(i)}$ by keeping only those coordinates belonging to $\hat{\mathcal{S}}_{i}$.\;

  \tcp*[h]{Step 4:} Set $\Tilde{D}_{1}=0$ and calculate, for $2 \leq i \leq K$, the discriminant score for class $i$ relative to class $1$ as 
\begin{align}
    \Tilde{D}_{i}=\bigl[\Tilde{\mathbf{Z}}-\bigl(\tfrac{\Tilde{X}_{i}+\Tilde{X}_{1}}{2}\bigr)\bigr]_{\hat{\mathcal{S}}_{i}}^{\top}\:\Hat{\zeta}^{(i)}_{\Hat{\mathcal{S}}_{i}}+\ln
  \frac{n_i}{n_1}.
\end{align} .\;

\tcp*[h]{Step 5:} Assign the label of $\mathbf{Z}$ to the class that maximizes the
  discriminant score, i.e.,
\begin{align}\label{K-PCLDA rule}
    \hat{\Upsilon}_{\mathrm{lda} \circ \mathrm{pca}}(\mathbf{Z})=\argmax_{i\in[K]} \Tilde{D}_{i}.
\end{align} \;

} 
\caption{$K$-classes $\mathrm{lda} \circ \mathrm{pca}$ decision rule}
\label{alg:multi_PCLDA}
\end{algorithm}

\subsection{Heterogeneous covariance matrices}
\label{PCQDA}
Despite the simplicity and popularity of regularized or sparse LDA for high-dimensional data,  
the assumption of equal covariances is not always tenable in practice. More specifically, suppose we are given a
$p$-variate random vector $\mathbf{Z}$ drawn from a mixture $\pi_1
\mathcal{N}_{p}(\mu_{1},\Sigma_{1})+ (1 - \pi_1)
\mathcal{N}_{p}(\mu_{2},\Sigma_{2})$ with $\Sigma_1$ possibly distinct from $\Sigma_2$. The
Bayes classifier is then the QDA rule given by
\begin{equation}\label{FisherQDA}
    \Upsilon_{F}(\mathbf{Z})=\begin{cases}
1 & \text{if
  $\bigl(\mathbf{Z}-\mu_{1}\bigr)^{\top}\Sigma_{1}^{-1}\bigl(\mathbf{Z}-\mu_{1}\bigr)-\bigl(\mathbf{Z}-\mu_{2}\bigr)^{\top}\Sigma_{2}^{-1}\bigl(\mathbf{Z}-\mu_{2}\bigr)\leq \kappa$}
\\
2 & \text{otherwise}
\end{cases}
\end{equation}
where $\kappa=2\ln \tfrac{\pi_1}{1 - \pi_1}-\ln{\tfrac{|\Sigma_{1}|}{|\Sigma_{2}|}}$. If 
$\Sigma_{1}=\Sigma_{2}$, then \cref{FisherQDA} reduces to \cref{Fisher}.
We now discuss how the results in \cref{Theory} can be 
extended to quadratic discriminant analysis (QDA) with PCA. 
Firstly we assume that $\Sigma_{1}$ and $\Sigma_{2}$ both have 
spiked covariance structures as specified below.
\begin{ass}
  \label{QDA spiked_covariance} Let
  $\bm{u}_{i1}, \dots, \bm{u}_{id_{i}},$ for $i=1,2$ be orthonormal vectors in
  $\mathbb{R}^{p}$ and assume that the covariance matrix $\Sigma_{i}$ for the $p$-variate normal distributions $\mathcal{N}_p(\mu_i, \Sigma_{i})$ is of the form
\begin{align}
    \Sigma_{i}&=\:\sum_{k=1}^{d_{i}}\lambda_{ik}\mathbf{u}_{ik}\mathbf{u}_{ik}^{\top}+\sigma_{i}^{2}\mathcal{I}_{p}
            = \mathcal{U}_{i}\Lambda_{i}\mathcal{U}_{i}^{\top}+\sigma_{i}^{2}\mathcal{I}_{p},\quad i=1,2.
\end{align}
Here $\Lambda_{i}= \mathrm{diag}(\lambda_{ik})$ is a $d_{i} \times d_{i}$ diagonal matrix 
and $\mathcal{U}_{i}$ is a $p \times d_{i}$ matrix with orthonormal columns. We assume implicitly that
$\lambda_{i1}\geq\cdots\geq\lambda_{id} > 0$, $\sigma_{i}>0$ and $d_{i}\ll p,\:i=1,2$.
\end{ass}

Note that a recent line of research on QDA for high-dimensional classification is based on the assumption that 
$\Sigma_{2}^{-1}-\Sigma_{1}^{-1}$ is sparse; see e.g., \cite{QDA_Shao, QDA_lasso,QDA_TonyCai}. In contrast, \cref{QDA spiked_covariance} 
do not enforce any sparsity assumption and also allows for $\Sigma_{1}$ and $\Sigma_{2}$ to have different {\em spiked} eigenvalues and eigenvectors; the latter is 
is a generalization of 
the common principal components (CPC) assumption in \cite{CPC_Biometrika,CPC_regularized,CPC_first_Flurry} where
the leading principal components are the same for both $\Sigma_1$ and $\Sigma_2$.

Under \cref{QDA spiked_covariance}, the whitening transformation $\mathcal{W}_{i}=\Sigma_{i}^{-1/2}$ for $i = 1,2$ is of the form in 
\cref{eq:population_whitening} and thus a suitable estimate for $\mathcal{W}_i$ is given by \cref{eq:whitening_estimate}. 
Let $\zeta_{i} = \mathcal{W}_i \mu_i$ for $i=1,2,$ and denote the whitened index sets by
$\mathcal{A}_{i}=\{j: \zeta_{ij} \neq 0\}$ for $i=1,2$.
Let $\mathcal{A}_{0}=\mathcal{A}_{1}\cup\mathcal{A}_{2}$ and note that the elements in $\mathcal{A}_0$ are the {\em signal} coordinates for the QDA rule 
(after the PCA step). 
\cref{FisherQDA} can be written as
\begin{align}\label{eq: QDA_decision boundary}
    \Upsilon(\mathbf{Z})
    := \begin{cases} 1 & \text{if 
      $\|[\mathcal{W}_{1}\bigl(\mathbf{Z}-\mu_{1}\bigr)]_{\mathcal{A}_{0}}\|^2 - 
\bigl\|[\mathcal{W}_{2}\bigl(\mathbf{Z}-\mu_{2}\bigr)]_{\mathcal{A}_{0}}\|^2 \leq \kappa$ }
      \\ 2 & \text{otherwise} \end{cases}
\end{align}

A plugin decision rule is then obtained by replacing $\mu_i, \Sigma_i$ and $\mathcal{A}_0$ with their estimates $\bar{X}_i, \hat{\Sigma}_i$ and $\hat{\mathcal{A}}_0$. Note that the intercept $\kappa$ is non-trivial to estimate
in the high-dimensional setting as it involves the log-determinant $\ln{\tfrac{|\Sigma_{1}|}{|\Sigma_{2}|}}$; see 
\cite{log_det_TonyCai} for further details. For our paper we employ the data-driven approach of \cite{QDA_lasso}
which circumvents the need to estimate the determinants of $\Sigma_{1}$ and $\Sigma_{2}$. 
The full details of $\mathrm{qda} \circ \mathrm{pca}$ are specified in Algorithm~\ref{alg:PCQDA} below, and theoretical results for $\mathrm{qda} \circ \mathrm{pca}$ are
provided in the supplementary.

\begin{algorithm}[H] 
\DontPrintSemicolon 
\SetNoFillComment 

\KwInput{$\Bar{X}_{1}$, $\Bar{X}_{2}$, $\hat{\Sigma}_{1}$, $\hat{\Sigma}_{2}$ and a
  test data point $\mathbf{Z}$ \vskip10pt}
\KwOutput{$\hat{\Upsilon}_{\mathrm{qda} \circ \mathrm{pca}}(\mathbf{Z})$ \vskip10pt}


\SetKwProg{Proc}{Algorithm}{}{} 
\Proc{}{ 

  \tcp*[h]{Step 1:} Perform PCA on the feature vectors for the training data (standard PCA approach). Extract, for $i = 1,2,$ the
    $d_{i}$ largest principal components $\hat{\mathcal{U}}_i$ and compute $\Hat{\mathcal{W}}_{i}$ as in \cref{eq:whitening_estimate}. \;

  \tcp*[h]{Step 2:} For $i=1,2,$, set $\hat{\zeta}_{i}=\hat{\mathcal{W}}_{i}\Bar{X}_{i}$ and form the indices set $\hat{\mathcal{A}}_{i}$ by selecting the $s_{i}$ largest (in magnitude) coordinates of
  $\Hat{\zeta}_{i}$. \;
  
  \tcp*[h]{Step 3:}Let $\hat{\mathcal{A}}_{0}=\hat{\mathcal{A}}_{1}\cup\hat{\mathcal{A}}_{2}$ and define 
  for any $\bm{x} \in \mathbb{R}^{p}$ 
  \[ \mathbf{Q}(x \mid \{\bar{X}_1, \bar{X}_2, \hat{\mathcal{W}}_1, \hat{\mathcal{W}}_2, \hat{\mathcal{A}}_0\}) =
    \bigl\|[\hat{\mathcal{W}}_{1}(\bm{x}-\bar{X}_{1})]_{\hat{\mathcal{A}}_{0}}\bigr\|^2 -     \bigl\|[\hat{\mathcal{W}}_{2}(\bm{x}-\bar{X}_{2})]_{\hat{\mathcal{A}}_{0}}\bigr\|^2.
\].\;

  \tcp*[h]{Step 4:}Find $\hat{\kappa}$ to minimize the empirical $0$-$1$ loss of the decision rule
  induced by $\mathbf{Q}$, i.e., 
\begin{align}
  \hat{\kappa}=\argmin_{\eta \in \mathbb{R}} \frac{1}{n}\sum_{i=1}^{2}\sum_{j=1}^{n_{i}}\mathbbm{1}(\Upsilon(X_{ij} \mid
  \{\bar{X}_1, \bar{X}_2, \hat{\mathcal{W}}_{1}, \hat{\mathcal{W}}_2, \hat{\mathcal{A}}_{0}, \eta_{\mathrm{thresh}}\})\neq i)
\end{align}
where, for any $\eta_{\mathrm{thresh}} \in \mathbb{R}$, we define
\[ \Upsilon(X_{ij}\:|\,\{\bar{X}_{1}, \bar{X}_2, \hat{\mathcal{W}}_{1}, \hat{\mathcal{W}}_2,
  \hat{\mathcal{A}}_{0},\eta_{\mathrm{thresh}}\}) = \begin{cases} 1 & \text{if $\mathbf{Q}(X_{ij} \mid \{\bar{X}_1, \bar{X}_2, \hat{\mathcal{W}}_1, \hat{\mathcal{W}}_2, \hat{\mathcal{A}}_0\})  \leq \eta_{\mathrm{thresh}}$} \\
    2 & \text{otherwise} \end{cases} \]
 .\;
\tcp*[h]{Step 5:} Given a test data point $\mathbf{Z}$, return the decision rule 
\begin{equation}
    \hat{\Upsilon}_{\mathrm{qda} \circ \mathrm{pca  }}(\mathbf{Z})=\begin{cases}
1 & \text{if
  $\mathbf{Q}(\mathbf{Z}\:|\,\{\bar{X}_{1}, \bar{X}_2, \hat{\mathcal{W}}_{1}, \hat{\mathcal{W}}_2,\hat{\mathcal{A}}_{0}\})\leq \hat{\kappa}$}\\
2 & \text{otherwise}
\end{cases}
\end{equation}

} 
\caption{$\mathrm{qda} \circ \mathrm{pca}$ decision rule}
\label{alg:PCQDA}
\end{algorithm}
As indicated by \citet{QDA_lasso}, $\eta_{\mathrm{thresh}}$ is selected by minimizing the in-sample misclassification error based on $\Bar{X}{1}$, $\Bar{X}{2}$, $\hat{\Sigma}{1}$, and $\hat{\Sigma}{2}$. Under \cref{QDA spiked_covariance}, the grid search can be narrowed to a neighborhood around $-2\log\left(\frac{|\Hat{\mathcal{W}}{2}|}{|\Hat{\mathcal{W}}{1}|}\right) - 2\log\left(\frac{n_1}{n_2}\right)$. This is justified by the improved stability of the log-determinant ratio in the spiked setting, where bulk eigenvalue $\sigma_{i}^{2}$ is consistently estimated via the pooled sample covariance and the leading spikes remain well-separated (from the bulks).

\section{Discussion} \label{discussion}
In this paper we addressed the classification problem  for
high-dimensional data by analyzing the prototypical $\mathrm{lda} \circ \mathrm{pca}$ 
classifier that first transforms the feature vectors using a whitening transformation, then performs
feature selection on the whitened data, and finally applies LDA in  
the dimensionally reduced space. We show that, under a spiked
eigenvalue structure for $\Sigma$, the mis-classification error rate for $\mathrm{lda} \circ \mathrm{pca}$ is
asymptotically Bayes optimal whenever $n \rightarrow \infty$ and
$n^{-1} \ln p \rightarrow 0$. While the Bayes consistency
of $\mathrm{lda} \circ \mathrm{pca}$ is similar to that of classifiers based on estimating the
discriminant direction $\beta$, the underlying assumptions and
motivations for our results are substantially different. Indeed, the
focus on PCA and the whitening matrix leads to the natural assumption
that $\Sigma$ has spiked/diverging eigenvalues while earlier results
that focus on estimation of $\beta$ had generally assumed that
$\Sigma$ is sparse or that the eigenvalues of $\Sigma$ are bounded. 
Numerical experiments indicate that $\mathrm{lda} \circ \mathrm{pca}$ is competitive with existing state-of-the-art classifiers while operating on a substantially lower-dimensional representation. This behavior persists even when the underlying sparsity levels in the raw and whitened feature spaces are matched; see more details in \cref{sec: sensitive}.




We now mention two interesting issues for future research.  The first
is to extend the theoretical results in this paper for combining
LDA with PCA to other, possibly non-linear, dimension reduction techniques such as
(classical) multidimensional scaling, kernel PCA, and Laplacian
eigenmaps \citep{belkin03:_laplac},
followed by learning a classifier in the dimensionally
reduced space. The second issue concerns the spiked covariance structure 
in \cref{spiked_covariance}. In particular, while 
\cref{spiked_covariance} is widely used in the literature, see e.g.,
\cite{rotation,Tony_sparse,Spiked,sparse_AOS}, its assumption on the
non-spiked eigenvalues and eigenvectors might be somewhat restrictive. 
We can consider relaxing
\cref{spiked_covariance} by assuming that $\Sigma$ arises
from an approximate factor model \citep{SpikedFan} or that $\Sigma$
can be decomposed into a low-rank plus sparse matrix structure
\citep{lowrank}. We surmise that, due to the focus on the whitening matrix
$\Sigma^{-1/2}$, theoretical analysis of $\mathrm{lda} \circ \mathrm{pca}$ under these more
general covariance structure will also leads to interesting technical
developments; e.g., while perturbation results for
$\hat{\Sigma}^{-1} - \Sigma^{-1}$ given $\hat{\Sigma}
- \Sigma$ are well-studied,
much less is known about
perturbation bounds for $\hat{\Sigma}^{-1/2} - \Sigma^{-1/2}$
given $\hat{\Sigma} - \Sigma$.

\bibliography{main}
\bibliographystyle{tmlr}

\section{Supplementary Numerical Results}
\subsection{Supplementary Results to Model~1 to Model~3}
We provide here several tables that supplement the 
simulation results in Section~\ref{Numerical}.
Table~\ref{tab:stron and weak model1} reports the magnitudes for the strong and weak signals in the discriminant direction $\beta$ and whitened direction $\zeta$ for Model 1.
Table
\ref{tab:strong model1} and \ref{tab:faint model1} reports the average
number of strong signals and weak signals 
captured by $\mathrm{lda} \circ \mathrm{pca}$, DSDA,
AdaLDA and SLDA under Model 1. From Table~\ref{tab:faint model1}
we see that $\mathrm{lda} \circ \mathrm{pca}$ 
includes only a few
features corresponding to the weak signals and thus selects only a
small number of features compared to the other classifiers. We also
report in Table \ref{tab:condition number} the summary statistics for 
the condition numbers of the covariance matrix $\Sigma$ in Model 3;
these statistics indicate that the performance of AdaLDA for Model 3 
can be sub-optimal as AdaLDA estimates $\beta$ by solving a 
linear programming problem which is 
numerically unstable when $\hat{\Sigma}$ have large 
condition numbers. 

\begin{table}[h]

\caption{\label{tab:stron and weak model1} Strong and faint signals for equal correlation setting (model 1). See the discussion in Remark~\ref{rem:strong_faint_signal}}
\begin{center}
\begin{tabular}{l|ccccc}
  $\rho$ & 0.50 & 0.60 & 0.70 & 0.80 & 0.90 
\\ \hline \\
 strong signal $\beta$ & 1.98 & 2.47 & 3.29 & 4.94 & 9.88 \\ 
strong signal $\zeta$ & 1.40 & 1.56 & 1.80 & 2.21 & 3.12 \\ 
\hline
 faint signal $\beta$ & -0.02 & -0.03 & -0.04 & -0.06 & -0.12\\
 faint signal $\zeta$ & -0.02 & -0.02 & -0.02 & -0.03 & -0.04 \\ 
\end{tabular}    
\end{center}

\end{table}

\begin{table}[h] 
\caption{\label{tab:strong model1}Average number of strong discriminative and whitened variable with standard deviations in parentheses for equal correlation setting (model 1)}
\begin{center}
\begin{tabular}{l|ccccc}
  $\rho$ & 0.50 & 0.60 & 0.70 & 0.80 & 0.90 \\ 
  \hline \\
  $\mathrm{lda} \circ \mathrm{pca}$ & 9.59 (0.82)& 9.42 (0.85)& 8.53 (1.41)& 8.53 (1.41)& 3.68 (0.98)\\
  DSDA & 10.00 (0.00)& 9.99 (0.07)& 9.99 (0.07)& 9.99 (0.07)& 10.00 (0.00)\\
  AdaLDA& 9.95 (0.22)&9.93 (0.26)&9.95 (0.23)&10.00 (0.07)&10.00 (0.07)\\
  SLDA & 9.98 (0.22)& 9.98 (0.28)& 10.00 (0.00)& 10.00 (0.00)& 10.00 (0.00)\\
\end{tabular}    
\end{center}
\end{table}
\begin{table}[h]
\caption{\label{tab:faint model1}Average number of weak discriminative and whitened variable with standard deviations in parentheses for equal correlation setting (model 1)}
\begin{center}
\begin{tabular}{l|ccccc}
  $\rho$ & 0.50 & 0.60 & 0.70 & 0.80 & 0.90 \\ 
  \hline \\
  $\mathrm{lda} \circ \mathrm{pca}$ & 2.44 (4.22)& 1.90 (3.73)& 0.98 (3.35)& 0.98 (3.35)& 0.00 (0.00)\\
  DSDA & 86.20 (31.06)& 96.42 (32.57)& 107.16 (30.53)& 107.06 (29.96)& 86.48 (8.66)\\
  ADaLDA&36.59 (5.94)&35.91 (5.22)&36.78 (6.36)&37.86 (5.88)&38.65 (5.02)\\
  SLDA & 718.86 (226.57)& 778.18 (95.53)& 789.96 (0.21)& 789.96 (0.21)& 789.96  (0.21)\\ 
\end{tabular}
\end{center}
\end{table}
\begin{table}[h]
\caption{\label{tab:condition number}Summary statistics of condition numbers of generated $\Sigma$ from uniform, normal and student distributions}
\begin{center}
     \begin{tabular}{l| c c c} 
    Statistics & $\mathcal{U}(-1,1)$ &$\mathcal{N}(0,1)$& $\mathbf{t}_{5}$  \\\hline \\
        Mean&415.3&688.0&  1059.2\\
        SD &102.4&195.2&400.0\\ 
        Median &391.5&658.5&966.3\\ 
        IQR &108.0&205.9&314.1\\
        Max &849.9&1622.1&3708.4\\ 
        Min &266.3&396.9&536.0\\
    \end{tabular}
\end{center}
  \end{table}

\clearpage
\subsection{Sensitive Analysis}\label{sec: sensitive}
This subsection aims to explore the sensitivity of $\mathrm{lda} \circ \mathrm{pca}$ to the number of spikes $d$. As a baseline, we list the performance of Oracle and DSDA. Consider a covariance matrix \( \Sigma = (\sigma_{ij})_{p \times p} \) with parameters \( \rho_1, \rho_2 > 0 \) and \( \rho_1 + \rho_2 < 2 \):
\begin{equation}\label{eq:model4}
    \Sigma = \rho_1 p\, \mathbf{u}_1 \mathbf{u}_1^\top + \rho_2 p\, \mathbf{u}_2 \mathbf{u}_2^\top + (2 - \rho_1 - \rho_2)(\mathcal{I}_{p}-\mathbf{u}_1 \mathbf{u}_1^\top-\mathbf{u}_2 \mathbf{u}_2^\top)
\end{equation}
where
$
\mathbf{u}_1 = \frac{1}{\sqrt{p}} \bm{1}_p,$ and $
\mathbf{u}_2 = \frac{1}{\sqrt{2}}(\mathbf{e}_{1}-\mathbf{e}_{2}) 
$.

This covariance structure is a slight modification of Model~1 (the equal correlation model) to incorporate two spikes corresponding to the eigenvalues \(\rho_1 p\) and \(\rho_2 p\). The class means are set to be \(\mu_1 = \mathbf{0}_p\) and \(\mu_2 = \mathbf{e}_1+\mathbf{e}_2 - 2\mathbf{e}_3\) so that the mean difference $\mu_2-\mu_1$, discriminant direction $\beta$ and whitened direction $\zeta$ are all sparse with exactly three nonzero entries, located at the first three coordinates. This contrasts with the dense structures exhibited in Model~1 and Model~3.

For sensitivity analysis, our method variants are labeled as \(\mathrm{lda} \circ \mathrm{pca}(d)\) for \(d = 1, 2, 3\), with \(d=2\) being the correctly specified case. Training and test samples are generated using $\Sigma$ from \cref{eq:model4}, following Monte Carlo setups and CV procedure detailed in Section~\ref{Simulation}. Table~\ref{tab:err model4} shows that correct specifying or overestimating
$d$ results in classification errors comparable to DSDA whereas underestimating $d$ leads to a notable degradation in performance. Since overestimation does not harm accuracy, we recommend, in practice, selecting $d$ as the smallest integer such that the cumulative eigenvalue ratio $(\sum_{i=1}^{d}
\hat{\lambda}_i)/\mathbf{tr}(\hat{\Sigma}) \geq 0.9$, ensuring $90\%$ of the total variance is retained in the selected subspace. 

DSDA achieves similar accuracy but tends to over-select non-informative features is as shown in Table~\ref{tab: nonzero model4} with average $49.46$ selected features well above the true sparsity level of $3$.
In comparison, $\mathrm{lda} \circ \mathrm{pca}$ maintains comparable accuracy across all simulation settings. Our method also yields more parsimonious and offers greater computational efficiency especially under the spike covariance structure.

\begin{table}[h]
\caption{Mis-classification rate ($\%$) with standard deviations ($\%$) in parentheses for \cref{eq:model4}, 
    based on $200$ independent Monte Carlo replicates.}
\label{tab:err model4}
\begin{center}
\begin{tabular}{l|ccccc}

$(\rho_{1}, \rho_{2})$ & (1.5, 0.3) 
\\ \hline \\
  Oracle & 0.30 (0.36) \\
  $\mathrm{lda} \circ \mathrm{pca}(d=1)$ & 24.19 (23.09) \\
  $\mathrm{lda} \circ \mathrm{pca}(d=2)$ & 0.55 (0.60) \\
  $\mathrm{lda} \circ \mathrm{pca}(d=3)$ & 0.55 (0.60) \\
  DSDA& 0.50 (0.49)

\end{tabular}
\end{center}
\end{table}

\begin{table}
 \caption{\label{tab: nonzero model4}
    Average number of nonzero coefficients with standard deviations in parentheses for for \cref{eq:model4}, 
    based on $200$ independent Monte Carlo replicates.
    }
\begin{center}
    \begin{tabular}{l|ccccc}
  
  $(\rho_{1},\rho_{2})$ & (1.5, 0.3) 
  \\ \hline \\
  $\mathrm{lda} \circ \mathrm{pca}(d=1)$ & 1.49 (1.90)\\
  $\mathrm{lda} \circ \mathrm{pca}(d=2)$ & 3.82 (4.28)\\
  $\mathrm{lda} \circ \mathrm{pca}(d=3)$ & 3.71 (3.92)\\
  DSDA & 49.35 (32.97) \\
   \hline
\end{tabular}
\end{center}

\end{table}

\subsection{Elliptical distributions}
\label{sec:elliptical}
A random vector $X \in \mathbb{R}^{p}$ is said to have an elliptical distribution with
mean $\mu$ and covariance matrix $\Sigma$ if the
probability density function for $X$ is of the form
\begin{equation}
  \label{elliptical}
f(\bm{x}) = \mathcal{C}_{f} |\Sigma|^{-1/2} h((\bm{x} - \mu)^{\top}
\Sigma^{-1} (\bm{x} - \mu)), \quad \text{for all $\bm{x} \in
  \mathbb{R}^{p}$}.
\end{equation}
Here $\mathcal{C}_{f}$ is a normalization constant, $|\Sigma|$ is the
determinant of $\Sigma$ and $h$ is a monotone function on $[0, \infty)$. 
The class of elliptical distributions is analytically tractable and
many results that hold for multivariate normal distributions can
be extended to general elliptical distributions. In particular, \cite{EllipticalFisher} showed that Fisher's LDA rule in \cref{Fisher} is also Bayes optimal whenever the
feature vectors $X$ are sampled from a mixture of two elliptical
distributions with {\em known} covariance matrix $\Sigma$ and {\em known} class
conditional means $\mu_1$ and $\mu_2$. 
\citet{JunShao2011} and \citet{TonyCai2011} leveraged this fact to show that, under
certain conditions on the sparsity of either $\mu_1 - \mu_2$ and
$\Sigma$ or the sparsity of $\Sigma^{-1}(\mu_1 - \mu_2)$, both SLDA and AdaLDA classifiers
also achieve the Bayes error rate for elliptical
distributions with unknown $\Sigma, \mu_1$ and $\mu_2$.
We now discuss how the theoretical properties for $\mathrm{lda} \circ \mathrm{pca}$ can be extended to elliptical
distributions, provided that they have sub-Gaussian
tails as define below. 

\begin{definition}\label{Orlicz_norm} Let $\Psi:[0,\infty) \rightarrow [0,\infty)$ be a
  non-decreasing and non-zero convex function with $\Psi(0)=0$. Let
  $Z$ be a mean $0$ random variable. The
  Birnbaum-Orlicz $\Psi$-norm of $Z$ is defined as 
\begin{equation}
    \norm{Z}_{\Psi} = \inf  \bigg\{s \geq 0: \mathbb{E}\Psi\Bigl(\frac{Z}{s}\Bigr) \leq 2 \bigg\}.
\end{equation}
Similarly, if $\bm{Z}$ is a mean $0$ random vector taking values in
$\mathbb{R}^{p}$ then its $\Psi$-norm is
defined as
\begin{equation}
    \norm{\bm{Z}}_{\Psi} = \sup_{w \in \mathbb{R}^{p},\norm{w} = 1}\|w^{\top}\bm{Z}\|_{\Psi}.
\end{equation}
\end{definition}
Let $\Psi_1(x) = \exp(|x|)$ and $\Psi_2(x) = \exp(x^2)$. A mean $0$ random
vector $Z$ is said to be sub-exponential if $\|\bm{Z}\|_{\Psi_1} <
\infty$ and is said to be sub-Gaussian if $\|\bm{Z}\|_{\Psi_2} <
\infty$. If $\bm{Z} \in \mathbb{R}^{p}$ is a mean $0$ sub-Gaussian random
vector then $w^{\top} \bm{Z}$ is a sub-Gaussian random variable for
all $w \in \mathbb{R}^{p}$. Furthermore, if $Z$ is a sub-Gaussian random
variable then there exists
a universal constant $K$ such that for all $t > 0$, we have
$$\mathbb{P}(Z > t) \leq 2 \exp\bigl(-t^2/(K \|Z\|_{\Psi_2})^2 \bigr).$$
For more on sub-Gaussian
random vectors, see \citet[Section~2.5, Section~3.4]{HDProbability}.
\begin{remark}
  If $\bm{Z} \in \mathbb{R}^{p}$ is a mean $0$ sub-Gaussian random vector
  with covariance matrix $\Sigma$ then
  $\|w^{\top}\mathbf{Z}\|_{\Psi_{2}}^{2}\geq w^{\top} \Sigma w$ for
  all $w \in \mathbb{R}^{p}$. In this paper we shall assume that a converse
  inequality also holds, namely that
    there exist a constant $c_{1}>0$ such that, 
\begin{equation}\label{Orlicz_norm_equivalence}
    w^{\top} \Sigma w \geq c_{1} \|w^{\top}\mathbf{Z}\|_{\Psi_{2}}^2, \qquad \text{for all $w \in \mathbb{R}^{p}$}. \\
\end{equation}
We note that the constant $c_1$ in \cref{Orlicz_norm_equivalence} can depend on $\mathbf{Z}$ but
does not depend on the choice of $w \in
\mathbb{R}^{p}$.   
  If $\mathbf{Z}$ is multivariate normal then
  \cref{Orlicz_norm_equivalence} always hold. If $\mathbf{Z}$ is
  a zero mean sub-Gaussian random vector but not a multivariate normal 
  then \cref{Orlicz_norm_equivalence} allows us to bound the Orlicz
  norm of $w^{\top}\mathbf{Z}$ for any $w \in
  \mathbb{R}^{p}$ in terms of its variance. This then allow us to obtain better estimate for $\hat{\Sigma}-\Sigma$ in
  spectral norm, especially in the setting where the spiked
  eigenvalues could diverge with $p$. See for example Theorem~4.7.1 in \cite{HDProbability} and Theorem~9 in \cite{Lounici2017}. 
\end{remark}

We can now reformulate the classification problem in the earlier part
of this paper to the case of elliptical distributions as follows. 
Let $\{\epsilon_{11},\dots,\epsilon_{1n_{1}}\}$ and
$\{\epsilon_{21},\dots,\epsilon_{2n_{2}}\}$ be independently and
identically distributed mean $0$ random vectors with probability density functions of the form in \cref{elliptical} and suppose that the training sample is given by
$X_{ij} = \mu_{i}+\mathbf{\epsilon}_{ij}$ for $i \in \{1,2\}$ and $j \in \{1,2,\dots,n_i\}$.  

Given these training samples $\{X_{ij}\}$,
let $\mathbf{X}$ be the $(n_1 + n_2) \times p$ matrix whose rows are
the $\{X_{ij}\}$. Then Theorem~\ref{sharp two-to-infinity}, in particular
\cref{eq:bound_U_Uhat}, also holds for these $\mathbf{X}$ as long
as the $\{\epsilon_{ij}\}$ satisfy
\cref{Orlicz_norm_equivalence}. This then implies that Theorem~\ref{infty norm converge}, in particular
\cref{eq:zetahat_zeta}, also holds when the
$\mathbf{X}$ are sub-Gaussian random vectors. The resulting bound for
$\|\hat{\zeta} - \zeta\|_{\infty}$ allows us to recover
$\mathcal{S}_{\zeta}$ by thresholding $\hat{\zeta}$, and hence
$\hat{R}_{\mathrm{lda} \circ \mathrm{pca}} \rightarrow R_{F}$. In summary, 
$\mathrm{lda} \circ \mathrm{pca}$ is asymptotically Bayes optimal whenever the feature
vectors $\{X_{ij}\}$ are sampled from a mixture of elliptical
distributions with sub-Gaussian tails.

\newpage
\section{Proofs of Stated Results}
This section contains the proofs of Theorem \ref{sharp
  two-to-infinity} through Theorem~\ref{selection consistency and
  error consistency}. We will present the proofs under the more general assumption
that the feature vectors $\{X_{ij}\}$ are sub-Gaussian random vectors; 
see also the discussion in Section~\ref{sec:elliptical}. 

\subsection{Preliminary results}
We start by listing some elementary but useful facts about the $2 \to
\infty$ norm and its relationships with other matrix norms. Recall that $\|\cdot\|$ denote the spectral norm if its argument is a matrix and denote the $\ell_2$ norm if its argument is a vector. 

\begin{prop}\label{two infinity norm prop} Let $A\in \mathbb{R}^{p_{1} \times
    p_{2}}$ and $B\in \mathbb{R}^{p_{2} \times p_{3}}$ be arbitrary
  real-valued matrices. Let $x \in \mathbb{R}^{p_{2}}$ be an arbitrary vector. For a given 
  $i \in [p_1]$, let $\bm{e}_i$ 
  denote the $i$th elementary basis vector in $\mathbb{R}^{p_1}$. Then
\begin{align}
    \label{max row}&\norm{A}_{2 \rightarrow \infty}=
    \max_{i \in [p_{1}]}\|A^{\top}\mathbf{e}_{i}\|;\\
    \label{2 norm loose1}&\norm{Ax}_{\infty} \leq 
    \norm{A}_{2 \rightarrow \infty} \times \|x\|;\\
    \label{2 norm loose2}&\norm{AB}_{2 \rightarrow \infty} \leq\norm{A}_{2 \rightarrow \infty} \times \|B\|.
\end{align}
\end{prop}
\cref{max row} states that the two-to-infinity
norm of a matrix $A$ is equivalent to the {\em maximum}
$\ell_{2}$ norm of the rows of $A$.  
\cref{2 norm loose1} provides a bound for $\|Ax\|_{\infty}$ that
is tighter than the naive bound 
$\norm{Ax}_{\infty}\leq \|Ax\| \leq \norm{A}
\norm{x}$.

Throughout the section, we will make use of Bernstein's inequality. For completeness, we state the result below without proof.

\begin{prop}
    Let \( X_1, X_2, \dots, X_n \) be independent, mean-zero, sub-exponential random variables. Then for all \( t > 0 \), there exists a constant \( c > 0 \) such that:
\begin{equation}
\mathbb{P}(|\sum_{i=1}^n X_i| \geq t) \leq 2 \exp \left( -c \cdot \min\left( \frac{t^2}{\sum_{i=1}^{n}\|X_i\|^{2}_{\psi_1}}, \frac{t}{\max_{i}\|X_i\|_{\psi_1}} \right) \right).    
\end{equation}
where \( \|\cdot\|_{\psi_1} \) denotes the sub-exponential Orlicz norm.

\end{prop}

Let $\xi_{i}=\Bar{X}_{i}-\mu_{i}$. Recall that $\eta_{k}=(\lambda_{k}+\sigma^{2})^{-1/2}$
and $\sigma^{2}= \tfrac{1}{p-d}(\mathrm{tr}(\Sigma)-\mathrm{tr}(\Lambda))$
are the eigenvalues of $\mathcal{W} = \Sigma^{-1/2}$.  
The following lemma provides several concentration inequalities for 
$\hat{\Sigma} - \Sigma$,  $\hat{\eta}_{k}-\eta_{k}$ and $\sigma^{2}-\hat{\sigma}^{2}$.

\begin{lem}
  \label{lem:basic_bounds} Assume that the random variables $\{X_i\}$
  satisfy \cref{Orlicz_norm_equivalence} and the 
  covariance matrix $\Sigma$ satisfies \cref{n_order_Divergent} and 
  \cref{Bounded_Coherence}. 
  Then the following bounds hold simultaneously with probability at 
  least $1- p^{-2}$ (where $\hat{\Sigma}_0$ is defined in 
  \cref{eq:Sigma_0})
 \begin{gather}
     \label{Lounici result}
     \|\hat{\Sigma}_{0}-\Sigma\|  =
     \mathcal{O}(p\sqrt{n^{-1} \ln p}), \\
     \label{E1 E2 concentration}
     \norm{\xi_{i}}^2=
     \mathcal{O}(p\:n^{-1} \ln p), \\
      \label{sample covariance}
      \|\hat{\Sigma}-\Sigma\|
     =  \mathcal{O}(p\sqrt{n^{-1} \ln p}), \\
     \label{eta concentration}|
     \hat{\eta}_{k}-\eta_{k}| = \mathcal{O}(\sqrt{p^{-1} n^{-1} \ln p} ), \quad \text{for all $k \in [d]$} \\
     \label{sigma concentration}|
     \sigma^{2}-\hat{\sigma}^{2}| =
    \mathcal{O}(\sqrt{n^{-1} \ln p}).
   \end{gather}

\end{lem}
\cref{Lounici result} is given in \cite{Lounici2014} and
\cite{Lounici2017} while \cref{E1 E2
concentration} follows from an application of Bernstein inequality;
see Section~2 and Section~3 of \cite{HDProbability}.
\cref{sample covariance} follows from
\cref{Lounici result} and \cref{E1 E2
  concentration} together with the observation that
$\hat{\Sigma}_{0}-\hat{\Sigma}=n ^{-1}(n_1 \xi_1 \xi_1^{\top} + n_2 \xi_2 \xi_2^{\top})$. 
Finally, \cref{eta concentration} and \cref{sigma
concentration} follow from \cref{sample
covariance} and Weyl’s inequality.

\subsection{Proof of Theorem~\ref{sharp two-to-infinity}} Recall that $\mathcal{U}$ and $\hat{\mathcal{U}}$ denote the $p \times d$ 
matrices whose columns are the orthonormal 
eigenvectors corresponding to the $d$ largest eigenvalues of $\Sigma$ and the {\em pooled} sample covariance matrix 
$\hat{\Sigma}$, respectively. Now let $\mathcal{U}_{\perp}$ and $\hat{\mathcal{U}}_{\perp}$ be the $p \times (p-d)$ matrices whose orthonormal 
columns are the remaining eigenvectors of $\Sigma$ and $\hat{\Sigma}$,
respectively, i.e., $\mathcal{I}_{p} - \mathcal{U} \mathcal{U}^{\top} =
\mathcal{U}_{\perp} \mathcal{U}_{\perp}^{\top}$ and $\mathcal{I}_{p} -
\hat{\mathcal{U}} \hat{\mathcal{U}}^{\top}  = 
\hat{\mathcal{U}}_{\perp} \hat{\mathcal{U}}_{\perp}^{\top}$. 

Let $\Xi$ be the $d \times d$ orthogonal matrix that minimizes
$$\min_{W} \|W - \mathcal{U}^{\top} \hat{\mathcal{U}}\|_{F}$$
among all orthogonal matrices. Let $\mathcal{E}_n = \hat{\Sigma} -
\Sigma$. Then by Theorem~3.7 in \cite{Cape2019} we have  
\begin{equation}
  \begin{split}
  \label{eq:decomp_thm1}
  \|\mathcal{\hat{U}}-\mathcal{U}\Xi\|_{ 2 \rightarrow \infty} &\leq 2 (\lambda_d + \sigma^2)^{-1}
                                                                      \|(\mathcal{U}_{\perp}\mathcal{U}_{\perp}^{\top})\mathcal{E}_{n}(\mathcal{U}\mathcal{U}^{\top})\|_{ 2 \rightarrow \infty} 
  \\ &+ 2 (\lambda_d + \sigma^2)^{-1}
 \|(\mathcal{U}_{\perp}\mathcal{U}_{\perp}^{\top})\mathcal{E}_{n}
       (\mathcal{U}_{\perp}\mathcal{U}_{\perp}^{\top})\|_{ 2 \rightarrow \infty} \times \|\sin\Theta(\mathcal{\hat{U}},\mathcal{U})\| \\
    &+ 2 (\lambda_d + \sigma^2)^{-1}
 \|(\mathcal{U}_{\perp}\mathcal{U}_{\perp}^{\top})\Sigma(\mathcal{U}_{\perp}\mathcal{U}_{\perp}^{\top})\|_{ 2 \rightarrow \infty} \times \|\sin\Theta(\mathcal{\hat{U}},\mathcal{U})\| \\ 
    &+\|\sin\Theta(\mathcal{\hat{U}},\mathcal{U})\|^2 \times \|\mathcal{U}\|_{2 \rightarrow \infty}.
    \end{split}
\end{equation}
Now recall the matrix $\hat{\Sigma}_0$ from \cref{eq:Sigma_0}. We then have
$$\:\:\underbrace{\hat{\Sigma}-\Sigma}_{\mathcal{E}_{n}}=\underbrace{\Hat{\Sigma}_{0}-\Sigma}_{E_{n}}-\frac{n_{1}}{n}\underbrace{(\Bar{x}_{1}-\mu_{1})(\Bar{x}_{1}-\mu_{1})^{\top}}_{\mathrm{E}_{1}}-\frac{n_{2}}{n}\underbrace{(\Bar{x}_{2}-\mu_{2})(\Bar{x}_{2}-\mu_{2})^{\top}}_{\mathrm{E}_{2}}.$$ 

Using the same argument as that for the proof of Theorem~1.1 in \cite{Cape2019} we have
with probability at least $1-p^{-2}$ that
\begin{align}
  \label{eq:proof_thm1_term1}
  \|\mathcal{U}_{\perp}\mathcal{U}_{\perp} E_n \mathcal{U}\mathcal{U}^{\top}\|_{2\rightarrow\infty}
  & \leq \,\mathcal{C}d\, \Bigl(\max_{i\in[p]}\Sigma_{ii}\Bigr)^{1/2} \times \sqrt{\frac{ (\lambda_1 + \sigma^2)
    \ln{p}}{n}}\\
  \label{eq:proof_thm1_term2}
    \|\mathcal{U}_{\perp}\mathcal{U}_{\perp}^{\top} E_n \mathcal{U}_{\perp}\mathcal{U}_{\perp}^{\top}\|_{2\rightarrow\infty}&\leq\mathcal{C}\sigma \sqrt{\frac{(\lambda_1 + \sigma^2) \ln{p}}{n}}
\end{align} where we have used the assumption that $\lambda_{k} =
\Theta(p)$ for all $k \in [d]$ so that $\bm{r}(\Sigma)$ -- the
effective rank of $\Sigma$ -- is bounded. Here and in the subsequent
derivations we will, for simplicity of presentation, use $\mathcal{C}$
to denote a finite and {\em universal} constant that can change from
line to line.

Therefore to complete the proof of Theorem~\ref{sharp
  two-to-infinity}, it suffices to show that, for $j \in \{1,2\}$, the terms
$\|\mathcal{U}_{\perp}\mathcal{U}_{\perp}\mathrm{E}_{j}\mathcal{U}\mathcal{U}^{\top}\|_{2\rightarrow\infty}$
and
$\|\mathcal{U}_{\perp}\mathcal{U}_{\perp}^{\top}\mathrm{E}_{j}\mathcal{U}_{\perp}\mathcal{U}_{\perp}^{\top}\|_{2\rightarrow\infty}$
are of the same or smaller order than those in 
\cref{eq:proof_thm1_term1,eq:proof_thm1_term2}, respectively. 

We now bound $\|\mathcal{U}_{\perp}\mathcal{U}_{\perp}\mathrm{E}_{1}\mathcal{U}\mathcal{U}^{\top}\|_{2\rightarrow\infty}$. 
From \cref{Bounded_Coherence} and \cref{two
  infinity norm prop} we have
\begin{equation}
  \label{eq:UUEUU}
  \|\mathcal{U}_{\perp}\mathcal{U}_{\perp}^{\top}\|_{\infty}\leq
\mathcal{C}\sqrt{d}, \quad \text{and} \quad
\|\mathcal{U}_{\perp}\mathcal{U}_{\perp}^{\top}\mathrm{E}_{1}\mathcal{U}\mathcal{U}^{\top}\|_{
  2 \rightarrow \infty}\leq
\mathcal{C}\sqrt{d}\|\mathrm{E}_{1}\mathcal{U}\|_{2\rightarrow\infty}.
\end{equation}
Furthermore, we also have 
\begin{align*}
    \|\mathrm{E}_{1}\mathcal{U}\|_{2 \rightarrow \infty}&\leq \sqrt{d}\max_{i\in [p],j\in [d]} \bigl|\langle\mathrm{E}_{1}\mathbf{e}_{i}^{(p)},\mathbf{u}_{j}\rangle\bigr|=\sqrt{d}\max_{i\in [p],j\in [d]} \bigl|[(\Bar{X}_{1}-\mu_{1})^{\top}\mathbf{e}_{i}^{(p)}] \times [(\Bar{X}_{1}-\mu_{1})^{\top}\mathbf{u}_{j}]\bigr|. 
\end{align*}
Since $X_i$ is sub-Gaussian, by the properties of Orlicz norms we have
\begin{align*}
  \bigl\|[(\Bar{X}_{1}-\mu_{1})^{\top}\mathbf{e}_{i}^{(p)}] \times
  [(\Bar{X}_{1}-\mu_{1})^{\top}\mathbf{u}_{j}]\bigr\|_{\Psi_{1}}\leq
  \bigl\|(\Bar{X}_{1}-\mu_{1})^{\top}\mathbf{e}_{i}^{(p)}\bigr\|_{\Psi_{2}}
  \times \bigl\|(\Bar{X}_{1}-\mu_{1})^{\top}\mathbf{u}_{j}\bigr\|_{\Psi_{2}}
\end{align*}
\cref{Orlicz_norm_equivalence} implies that there exists a
constant $\mathcal{C} > 0$ such that for any $i \in [p]$ and
$j \in [d]$, 
\begin{gather}
  \label{eq:orlicz1}
    \bigl\|(\Bar{X}_{1}-\mu_{1})^{\top}\mathbf{e}_{i}^{(p)}\bigr\|_{\Psi_{2}}\leq
                                                                            \mathcal{C}\sqrt{\frac{\Sigma_{ii}}{n_{1}}}
                                                                            \leq
                                                                            \frac{\mathcal{C}}{\sqrt{n_{1}}}
                                                                            \Bigl(\max_{i
                                                                            \in[p]}\Sigma_{ii}\Bigr)^{1/2},
  \\
  \label{eq:orlicz2}
    \bigl\|(\Bar{X}_{1}-\mu_{1})^{\top}\mathbf{u}_{j}\bigr\|_{\Psi_{2}}\leq
                                                                      \mathcal{C}\sqrt{\mathrm{Var}(\bm{u}_j^{\top}
                                                                      (\Bar{X}_{1}-\mu_{1})}
                                                                      \leq
                                                                      \mathcal{C}
                                                                      \sqrt{\bm{u}_j^{\top}
                                                                      n^{-1}
                                                                      \Sigma
                                                                      \bm{u}_j} 
                                                                      \leq\mathcal{C} \sqrt{\frac{\lambda_{1}+\sigma^{2}}{n_{1}}}.
\end{gather}
Now fix an arbitrary pair $(i,j)$ with $i \in [p]$ and $j \in [d]$. 
Then by \cref{Bounded_Coherence}, \cref{eq:orlicz1,eq:orlicz2},
and properties of sub-exponential random variables, we have
$$\mathbb{E}[|\langle\mathrm{E}_{1}\mathbf{e}_{i}^{(p)}, \mathbf{u}_{j}\rangle|]
\leq\,\frac{\mathcal{C}(\lambda_{1}+\sigma^{2})}{n_{1}} \times \sqrt{\frac{d}{p}}.$$
Furthermore, by Bernstein inequality
\citep[Section 2.8]{HDProbability}, there exists
a constant $\mathcal{C} > 0$ such that with probability at least
$1-\mathcal{O}(p^{-3})$,
\begin{align}
  \label{eq:EU_1}
|\langle\mathrm{E}_{1}\mathbf{e}_{i}^{(p)},\mathbf{u}_{j}\rangle|\leq
& \;\;\,\mathbb{E}[|\langle\mathrm{E}_{1}\mathbf{e}_{i}^{(p)},\mathbf{u}_{j}\rangle|]
+ \frac{\mathcal{C} \ln p}{n_1} \Bigl(\max_{i \in[p]}\Sigma_{ii}\Bigr)^{1/2} \sqrt{\lambda_{1}+\sigma^{2}}.
\end{align}
Now recall \cref{eq:UUEUU}. Then by \cref{eq:EU_1} together with a union bound over all $i \in
[p]$ and $j \in [d]$ we have, with probability at least $1 -
\mathcal{O}(p^{-2})$,
\begin{equation}
  \label{eq:UUE11}
\|\mathcal{U}_{\perp} \mathcal{U}_{\perp} \mathrm{E}_1 \mathcal{U}
\mathcal{U}\|_{2 \rightarrow \infty} \leq \mathcal{C} \sqrt{d} \|\mathrm{E}_1
\mathcal{U}\|_{2 \rightarrow \infty} \leq \mathcal{C} d \max_{i,j}
|\langle\mathrm{E}_{1}\mathbf{e}_{i}^{(p)},\mathbf{u}_{j}\rangle| \leq
\mathcal{C}\frac{d^{3/2} \sqrt{p} \ln{p}}{n}.
\end{equation}
where we had used \cref{n_order_Divergent}, namely $n_1 \asymp n$ and $\lambda_1 \asymp p$, when simplifying the above expression. An almost identical argument also yields
$$\|\mathcal{U}_{\perp} \mathcal{U}_{\perp} \mathrm{E}_2 \mathcal{U}
\mathcal{U}\|_{2 \rightarrow \infty} \leq \mathcal{C}\frac{d^{3/2} \sqrt{p} \ln{p}}{n}$$
with probability at least $1 - \mathcal{O}(p^{-2})$. 
We therefore have
\begin{equation}
  \begin{split}
  \label{a-1}
  \frac{2 \|\mathcal{U}_{\perp}\mathcal{U}_{\perp}^{\top}\mathcal{E}_{n}\mathcal{U}\mathcal{U}^{\top}\|_{ 2 \rightarrow \infty}}{\lambda_{d}+\sigma^{2}}
  &\leq \, \frac{\mathcal{C}}{\lambda_d + \sigma^2} \Bigl(d \Bigl(\max_{i} \Sigma_{ii}\Bigr)^{1/2} \times \sqrt{\frac{(\lambda_1 + \sigma^2) \ln p}{n}} + \frac{d^{3/2} \sqrt{p} \ln p}{n}\Bigr)
  \\ & \leq \mathcal{C}d^{3/2}
    \sqrt{\frac{\ln{p}}{np}}
  \end{split}
\end{equation}
We next consider
$\|\mathcal{U}_{\perp}\mathcal{U}_{\perp}^{\top}\mathrm{E}_{1}\mathcal{U}_{\perp}\mathcal{U}_{\perp}^{\top}\|_{2\rightarrow\infty}$. For $j > d$, we have
$$\|(\bar{X}_1 - \mu_1)^{\top} \bm{u}_j\|_{\Psi_{2}} \leq \mathcal{C}
\sqrt{\bm{u}_j^{\top} n^{-1} \Sigma \bm{u}_j} \leq \mathcal{C}
n^{-1/2} \sigma.$$
Then following the same argument as that used for showing
\cref{eq:UUE11}, we have
\begin{align}\label{eq:UUE11 perp}
    &\bigl\|\mathcal{U}_{\perp}\mathcal{U}_{\perp}^{\top}\mathrm{E}_{1}\mathcal{U}_{\perp}
      \mathcal{U}_{\perp}^{\top}\bigr\|_{2 \rightarrow \infty}
      \leq\mathcal{C}\,\sqrt{p\,d}\,\sigma \Bigl(\max_{i \in[p]}\Sigma_{ii}\Bigr)^{1/2}
       \times \frac{\ln{p}}{n_{1}} \leq\mathcal{C}\,\sigma\sqrt{p\,d}\;\frac{\ln{p}}{n}
\end{align}
with probability at least $1 - \mathcal{O}(p^{-2})$, and similarly for
$\|\mathcal{U}_{\perp}\mathcal{U}_{\perp}^{\top}\mathrm{E}_{2}
\mathcal{U}_{\perp} \mathcal{U}_{\perp}^{\top}\|_{2 \rightarrow \infty}$. 

Next we have, by the Davis-Kahan theorem and Lemma
\ref{lem:basic_bounds}, that
\begin{align}\label{equ:sin_theta}
    \|\sin\Theta(\mathcal{\hat{U}},\mathcal{U})\| \leq
  \frac{\norm{\mathcal{E}_{n}}}{\lambda_d + \sigma^{2}}\leq\mathcal{C}\,
  \sqrt{\frac{\ln{p}}{n}}
\end{align} with probability at least $1 -
\mathcal{O}(p^{-2})$.
From \cref{eq:UUE11 perp,equ:sin_theta,eq:proof_thm1_term2},
together with a similar argument as that for showing \cref{a-1},
we have
\begin{align}
    \label{a-2}
  \frac{2 \|\mathcal{U}_{\perp}\mathcal{U}_{\perp}^{\top}\mathcal{E}_{n}\mathcal{U}_{\perp}\mathcal{U}_{\perp}^{\top}\|_{2\rightarrow\infty} \times \|\sin\Theta(\mathcal{\hat{U}},\mathcal{U})\|}{(\lambda_{d}+\sigma^{2})}
  \leq \mathcal{C} \sigma d \times \frac{\ln p}{n \sqrt{p}}
\end{align}
with probability at least $1 - \mathcal{O}(p^{-2})$. \cref{equ:sin_theta} together with
\cref{Bounded_Coherence} also imply
\begin{align}\label{a-4}
  \|\sin\Theta(\mathcal{\hat{U}},\mathcal{U})\|_{2}^{2} \times \|\mathcal{U}\|_{2\rightarrow \infty}
  \leq\frac{\mathcal{C} \ln{p}}{n} \times \frac{\sqrt{d}}{\sqrt{p}}= \frac{\mathcal{C} \sqrt{d} \ln{p}}{n \sqrt{p}}.
\end{align} with probability at least $1 - \mathcal{O}(p^{-2})$.
Next note that
\[ \|\mathcal{U}_{\perp}\mathcal{U}_{\perp}^{\top}\Sigma \mathcal{U}_{\perp}\mathcal{U}_{\perp}^{\top}\|_{
    2 \rightarrow \infty}=\|\sigma^2 \mathcal{U}_{\perp}\mathcal{U}_{\perp}^{\top}\|_{2 \rightarrow \infty} \leq \sigma^2
\|\mathcal{U}_{\perp}\mathcal{U}_{\perp}^{\top}\| = \sigma^2. \]
We therefore have
\begin{align}\label{a-3}
  &\frac{2 \|\mathcal{U}_{\perp}\mathcal{U}_{\perp}^{\top} \Sigma \mathcal{U}_{\perp}\mathcal{U}_{\perp}^{\top}\|_{ 2 \rightarrow \infty} \times \|\sin\Theta(\mathcal{\hat{U}},\mathcal{U})\|}{\lambda_{d}+\sigma^{2}}
    \leq \frac{\mathcal{C} \sigma^2}{\lambda_d + \sigma^2} \times
    \sqrt{\frac{\ln{p}}{n}} \leq \frac{\mathcal{C} \sigma^2 \sqrt{\ln{p}}}{\sqrt{n} p}
\end{align}
Subtituting the bounds in \cref{a-1,a-2,a-3,a-4} into \cref{eq:decomp_thm1}
we obtain
\begin{align*}
  \bigl\|\mathcal{\hat{U}}-\mathcal{U}\Xi\bigr\|_{ 2 \rightarrow \infty} \leq
  \,\mathcal{C} \Bigl(\frac{d^{3/2} \sqrt{\ln p}}{\sqrt{np}} + \frac{d \ln p}{n \sqrt{p}} + \frac{\sigma^2 \sqrt{\ln p}}{\sqrt{n} p}\Bigr) 
    \leq \mathcal{C} \sqrt{\frac{d^{3} \ln p}{np}}
\end{align*}
with probability at least $1 - p^{-2}$. This completes the proof of \cref{sharp two-to-infinity}. 
  
\subsection{Proof of Theorem~\ref{infty norm converge}}
First recall that $\hat{\zeta} = \hat{\mathcal{W}}(\bar{X}_2 -
\bar{X}_1)$ and $\zeta = \mathcal{W}(\mu_2 - \mu_1)$ where the
whitening matrix $\mathcal{W}$ and its estimate $\hat{\mathcal{W}}$
are given by \cref{eq:whitening_form} and
\cref{eq:whitening_estimate}, respectively. We now consider the decomposition
\begin{align*}
  \hat{\zeta} - \zeta = \underbrace{\mathcal{W}\bigl[(\Bar{X}_{2}-\Bar{X}_{1})-(\mu_{2}-\mu_{1})\bigr]}_{A}
  + \underbrace{(\hat{\mathcal{W}}-\mathcal{W})\bigl[(\Bar{X}_{2}-\Bar{X}_{1})-(\mu_{2}-\mu_{1}) \bigr]}_{B}
+ \underbrace{(\hat{\mathcal{W}}-\mathcal{W})(\mu_{2}-\mu_{1})}_{C}. 
\end{align*}
We will now bound each of the term in the right hand side of the above
display. We start with the term in $(A)$. Let $\delta =
(\Bar{X}_{2}-\Bar{X}_{1})-(\mu_{2}-\mu_{1})$ and let $\xi =
\mathcal{W} \delta$. We then have $\mathbb{E}[\xi] = \bm{0}$ and $\mathrm{Var}[\xi]=c\;n^{-1}\mathcal{I}_{p}$
for some finite constant $c$. Since $\delta$ satisfies
\cref{Orlicz_norm_equivalence}, $\xi$ also satisfies
\cref{Orlicz_norm_equivalence}.
Hence, by Bernstein inequality for
sub-Gaussian random vectors, there exists a constant $\mathcal{C}
> 0$ such that with probability at least $1 - \mathcal{O}(p^{-2})$, 
\begin{equation}
  \label{eq:A_term}
 \bigl\|\mathcal{W} \bigl((\bar{X}_2 - \bar{X}_1) - (\mu_2 -
\mu_1)\bigr)\bigr\|_{\infty} = \|\xi\|_{\infty} \leq \mathcal{C} \sqrt{\frac{\ln p}{n}}. 
\end{equation}

We now bound the terms in $(B)$ and $(C)$. Let $\hat{\mathcal{D}}$ and
$\mathcal{D}$ be diagonal matrices where
$$\hat{\mathcal{D}} = \bigl(\hat{\Lambda}+ \hat{\sigma}^{2} \mathcal{I}_{d}\bigr)^{-1/2}, \quad \mathcal{D} = \bigl(\Lambda+ \sigma^{2} \mathcal{I}_{d}\bigr)^{-1/2}.$$
We start by decomposing $\hat{\mathcal{W}}-\mathcal{W}$ as 
\begin{equation}
  \label{eq:decomposition_hatW-W}
\hat{\mathcal{W}}-\mathcal{W}  =
                                 \underbrace{\hat{\mathcal{U}}\hat{\mathcal{D}}
                                 \hat{\mathcal{U}}^{\top} -\mathcal{U} \mathcal{D} \mathcal{U}^{\top}}_{\rom{1}}
+ \underbrace{(\hat{\sigma}^{-1} -
    \sigma^{-1})(\mathcal{I}_{p}-\mathcal{U}\mathcal{U}^{\top})}_{\rom{2}}
  + \underbrace{\hat{\sigma}^{-1}
    (\hat{\mathcal{U}}\hat{\mathcal{U}}^{\top}-\mathcal{U}\mathcal{U}^{\top})}_{\rom{3}}.
\end{equation}
Now consider the term $(\hat{\sigma}^{-1} - \sigma^{-1})(\mathcal{I}_{p} -
\mathcal{U} \mathcal{U}^{\top}) \delta$ obtained by combining the
expressions in $(B)$ and $(II)$. The covariance matrix for $\delta$ is $n^{-1} \Sigma$ and hence $(\mathcal{I}_{p} - \mathcal{U}
\mathcal{U})^{\top} \delta$ satisfies
\cref{Orlicz_norm_equivalence}
with covariance matrix $n^{-1} \sigma^2 (\mathcal{I}_{p} - \mathcal{U}
\mathcal{U}^{\top})$.
Therefore, by Bernstein inequality, there exists a constant
$\mathcal{C} > 0$ such that with probability at least $1 -
\mathcal{O}(p^{-2})$,
$$\|(\mathcal{I}_{p} - \mathcal{U} \mathcal{U}^{\top}) \delta \|_{\infty}
\leq \mathcal{C} \sigma \sqrt{\frac{\ln p}{n}}.$$
Furthermore, from Lemma~\ref{lem:basic_bounds}, we have with
probability at least $1 - \mathcal{O}(p^{-2})$ that
\begin{equation}\label{sigma-1}
    |\hat{\sigma}^{-1} - \sigma^{-1}| = \frac{|\hat{\sigma}^2 -
  \sigma^2|}{\hat{\sigma} \sigma (\hat{\sigma} + \sigma)} \leq
\mathcal{C} \sqrt{\frac{\ln p}{n}}.
\end{equation}
Combining the above bounds, we obtain
\begin{equation}\label{B-2}
    \|(\hat{\sigma}^{-1} - \sigma^{-1})(\mathcal{I} -
\mathcal{U} \mathcal{U}^{\top}) \delta\|_{\infty} \leq \frac{C \ln p}{n}.
\end{equation}
with probability at least $1 - \mathcal{O}(p^{-2})$.

We next consider the term $(\hat{\sigma}^{-1} -
\sigma^{-1})(\mathcal{I}_{p} - \mathcal{U} \mathcal{U}^{\top}) (\mu_2 -
\mu_1)$. Recall that if $\mathcal{U}$ has bounded coherence as in
\cref{Bounded_Coherence} then
$\|\mathcal{I}_{p} - \mathcal{U} \mathcal{U}^{\top}\|_{\infty} \leq (1 +
\mathcal{C}_{\mathcal{U}}) \sqrt{d}$ where $\mathcal{C}_{\mathcal{U}}$ is a finite constant. 
We therefore have, by Lemma~\ref{lem:basic_bounds}, that
\begin{equation*}
  \begin{split}\|(\hat{\sigma}^{-1} -
\sigma^{-1})(\mathcal{I}_{p} - \mathcal{U} \mathcal{U}^{\top}) (\mu_2 -
\mu_1)\|_{\infty} &= \|(\hat{\sigma}^{-1} -
\sigma^{-1})(\mathcal{I}_{p} - \mathcal{U} \mathcal{U}^{\top})\Sigma^{1/2}
\mathcal{W}(\mu_2 - \mu_1)\|_{\infty}\\
&= \|(\hat{\sigma}^{-1} -
\sigma^{-1}) \sigma (\mathcal{I}_{p} - \mathcal{U} \mathcal{U})^{\top}
\zeta \|_{\infty} \\
&\leq |(\hat{\sigma}^{-1} -
\sigma^{-1})| \times \sigma (1 + \mathcal{C}_{\mathcal{U}}) \sqrt{d} \times
\|\zeta\|_{\infty} \\ &\leq \mathcal{C}\sqrt{\frac{\ln p}{n}} \times \|\zeta\|_{\infty},
\end{split}
\end{equation*}
with probability at least $1 - \mathcal{O}(p^{-2})$. 

We now focus our efforts on terms involving
$\mathcal{\hat{U}}\mathcal{\hat{D}}\mathcal{\hat{U}}^{\top}-\mathcal{U}\mathcal{D}\mathcal{U}^{\top}$. Let
$\Xi$ be the minimizer of $\|W -
\mathcal{U}^{\top} \hat{\mathcal{U}}\|_{F}$ among all $d \times d$ orthogonal
matrices $W$. We then have
\begin{equation*}
  \begin{split}
    \mathcal{\hat{U}}\mathcal{\hat{D}}\mathcal{\hat{U}}^{\top}-\mathcal{U}\mathcal{D}\mathcal{U}^{\top}
    &=(\mathcal{\hat{U}}-\mathcal{U}\mathcal{U}^{\top}\mathcal{\hat{U}})\mathcal{\hat{D}}\mathcal{\hat{U}}^{\top}+\mathcal{U}\mathcal{U}^{\top}\mathcal{\hat{U}}\mathcal{\hat{D}}\mathcal{\hat{U}}^{\top}-\mathcal{U}\mathcal{D}\mathcal{U}^{\top}\\
    &=\underbrace{\Bigl[(\mathcal{\hat{U}}-\mathcal{U}\Xi)-\mathcal{U}(\mathcal{U}^{\top}\mathcal{\hat{U}}-\Xi)\Bigr]\mathcal{\hat{D}}\mathcal{\hat{U}}^{\top}}_{\text{Part 1}}+\underbrace{\mathcal{U}\mathcal{U}^{\top}\mathcal{\hat{U}}\mathcal{\hat{D}}\mathcal{\hat{U}}^{\top}-\mathcal{U}\mathcal{D}\mathcal{U}^{\top}}_{\text{Part 2}}
  \end{split}
\end{equation*}
We now note a few elementary but useful algebraic facts frequently used in
the subsequent derivations. 
\begin{fact} \label{sin fact} 
\begin{align}
  \label{algebraic1}
    &\|\mathcal{U}^{\top}\mathcal{\hat{U}}-\Xi\|\leq \|\sin \Theta(\mathcal{\hat{U}},\mathcal{U})\|^{2},\\
  \label{algebraic2}
    &\|\mathcal{U}\mathcal{U}^{\top}\mathcal{\hat{U}}_{\perp}\mathcal{\hat{U}}_{\perp}^{\top}\|
      =\|\mathcal{\hat{U}}\mathcal{\hat{U}}^{\top}\mathcal{U}_{\perp}
       \mathcal{U}_{\perp}^{\top}\| =\|\mathcal{U}^{\top}
       \mathcal{\hat{U}}_{\perp}\mathcal{\hat{U}}_{\perp}^{\top}\|=\|\mathcal{\hat{U}}^{\top}\mathcal{U}_{\perp}
       \mathcal{U}_{\perp}^{\top}\| =\|\sin\Theta(\mathcal{\hat{U},\mathcal{U}})\|,\\
  \label{algebraic3}
     &\|\sin \Theta(\mathcal{U},\hat{\mathcal{U}})\| \leq \|\mathcal{\hat{U}}\mathcal{\hat{U}}^{\top}-\mathcal{U}\mathcal{U}^{\top}\| 
       \leq 2\|\sin \Theta(\mathcal{\hat{U}},\mathcal{U})\|,\\
  \label{algebraic4}
  &\|\mathcal{U}^{\top}\mathcal{\hat{U}}\mathcal{\hat{D}}-\mathcal{D}\mathcal{U}^{\top}\mathcal{\hat{U}}\|
    \leq\|\mathcal{U}^{\top}(\Sigma-\hat{\Sigma})\mathcal{\hat{U}}\| \times \|\mathbb{H}\|.
\end{align}
\end{fact}
where $\mathbb{H}=(\mathbb{H}_{ij})$ is a $d \times d$ matrix with entries
\[\mathbb{H}_{ij}=
\frac{1}{\sqrt{(\hat{\lambda}_{j}+\hat{\sigma}^{2})(\lambda_{i}+\sigma^{2})}(
  \sqrt{\lambda_{i}+\hat{\sigma}^{2}} +\sqrt{\hat{\lambda}_{j}+\sigma^{2}})}.
\]
\cref{algebraic1} is from Lemma~6.7 in \cite{Cape2019} while
\cref{algebraic2,algebraic3} are standard results
for the $\sin$-$\Theta$ distance (see for example
Lemma~1 in \cite{TonyCai_wedin}).  
Finally, \cref{algebraic4} follows from the
observation \begin{align*}
    (\mathcal{U}^{\top}\mathcal{\hat{U}}\mathcal{\hat{D}}-\mathcal{D}\mathcal{U}^{\top}\mathcal{\hat{U}})_{ij}
    &=(\mathcal{U}^{\top}\mathcal{\hat{U}})_{i,j}(\hat{\mathcal{D}}_{jj}-\mathcal{D}_{ii})\\
    &=(\mathcal{U}^{\top}\mathcal{\hat{U}})_{ij}
      \frac{(\lambda_{i}+\sigma^{2}) - (\hat{\lambda}_{j}+\hat{\sigma}^{2})}{
      (\hat{\lambda}_{j}+\hat{\sigma}^{2})^{1/2}(\lambda_{i}+\sigma^{2})^{1/2}\bigl((\hat{\lambda}_{j}+\hat{\sigma}^{2})^{1/2}+(\lambda_{i}+\sigma^{2})^{1/2}\bigr)}\\
                        &= (\Lambda \mathcal{U}^{\top} \hat{\mathcal{U}}
                           - \mathcal{U}^{\top}
                          \hat{\mathcal{U}} \hat{\Lambda})_{ij}
                          \mathbb{H}_{ij} = (\mathcal{U}^{\top}(\Sigma
                          - \hat{\Sigma}) \hat{\mathcal{U}})_{ij} \mathbb{H}_{ij}
\end{align*}
We thus have
$\mathcal{U}^{\top}\mathcal{\hat{U}}\mathcal{\hat{D}}-\mathcal{D}\mathcal{U}^{\top}\mathcal{\hat{U}}=
(\mathcal{U}^{\top}(\Sigma-\hat{\Sigma})\mathcal{\hat{U}})\circ\mathbb{H}$.
Therefore, by Schur inequality for Hadamard product (see e.g.,
Theorem~5.5.1 of \cite{Horn_Johnson}), we have
$$\|(\mathcal{U}^{\top}(\Sigma-\hat{\Sigma})\mathcal{\hat{U}})\circ\mathbb{H}\|
\leq \|(\mathcal{U}^{\top}(\Sigma-\hat{\Sigma})\mathcal{\hat{U}})\|
\times \|\mathbb{H}\|.$$

We next state a technical lemma for bounding several terms that
appears frequently in our analysis. 

\begin{lem}
  \label{sin con}
  Suppose that \cref{spiked_covariance} through \cref{Bounded_Coherence} are satisfied. 
  Then with probability at least
  $1-\mathcal{O}(p^{-2})$, the following bounds hold simultaneously
   \begin{gather}
     \label{eq:norm_H} \norm{\mathbb{H}} =
     \mathcal{O}(d^{3/2}p^{-3/2}), \\
     \label{eq:U_E_U}
     \|\mathcal{U}^{\top}(\Sigma-\hat{\Sigma})\mathcal{U}\| =
     \mathcal{O}(p \sqrt{n^{-1} \ln p}), \\
     \label{eq:U_E_hat_U}
     \|\mathcal{U}^{\top}(\Sigma-\hat{\Sigma})\mathcal{\hat{U}}\|
     = \mathcal{O}(p \sqrt{n^{-1} \ln p}), \\
     \label{eq:Utop_hatU_hatD}
     \|\mathcal{U}^{\top}\mathcal{\hat{U}}\mathcal{\hat{D}}-\mathcal{D}\mathcal{U}^{\top}\mathcal{\hat{U}}\| = \mathcal{O}(n^{-1/2} p^{-1/2} \ln p), \\
     \label{eq:diff_projection}
     \|\mathcal{\hat{U}}\mathcal{\hat{U}}^{\top}-\mathcal{U}\mathcal{U}^{\top}\| =
     \mathcal{O}(\sqrt{n^{-1} \ln p}).
   \end{gather}
\end{lem}
\cref{eq:U_E_U,eq:U_E_hat_U} follows from the
sub-multiplicativity of the spectral norm together with bounds for
$\|\hat{\Sigma}_0 - \Sigma\|$ from \cite{Lounici2014} and
\cite{Lounici2017}. \cref{eq:diff_projection} is a consequence of 
\cref{algebraic3} and the Davis-Kahan
theorem. \cref{eq:norm_H} follows from Weyl's inequality and the
bound for $\|\hat{\Sigma} - \Sigma\|$. 
Finally, \cref{algebraic4,eq:norm_H,eq:U_E_hat_U} together imply \cref{eq:Utop_hatU_hatD}.

With the above preparations in place, we now resume our proof of
Theorem~\ref{infty norm converge}. We first have
\begin{equation}
  \label{eq:bound_part1}
  \begin{split}
    \|(\hat{\mathcal{U}} - \mathcal{U} \mathcal{U}^{\top}\mathcal{\hat{U})}
      \mathcal{\hat{D}}\mathcal{\hat{U}}^{\top}\delta
      \|_{\infty} &\leq (\|\mathcal{\hat{U}}-\mathcal{U}\Xi\|_{2
        \rightarrow \infty}+\|\mathcal{U}\|_{2 \rightarrow
        \infty}\|\mathcal{U}^{\top}\mathcal{\hat{U}}-\Xi\|)
      \|\hat{\mathcal{D}} \mathcal{\hat{U}}^{\top}\delta\|  \\
    &\leq
    \,\mathcal{C}\Bigl(\sqrt{\frac{d^{3} \ln{p}}{n p}}+\frac{\sqrt{d} \ln p}{n \sqrt{p}} \Bigr)
      \|\hat{\mathcal{D}} \mathcal{\hat{U}}^{\top}\delta\| 
    \\ & \leq
    \,\mathcal{C}\Bigl(\sqrt{\frac{d^{3} \ln{p}}{n p}}+\frac{\sqrt{d} \ln p}{n \sqrt{p}} \Bigr) (\|\hat{\mathcal{D}}\| + \|\hat{\mathcal{D}} - \mathcal{D}\|)
      \|\mathcal{\hat{U}}^{\top}\delta\| 
    \\ & \leq
    \,\mathcal{C}\Bigl(\sqrt{\frac{d^{3} \ln{p}}{n p}}+\frac{\sqrt{d} \ln p}{n \sqrt{p}} \Bigr)      \Bigl(\frac{1}{\sqrt{\lambda_{1}+\sigma^{2}}}+ \sqrt{\frac{\ln{p}}{np}}\Bigr)
      \|\mathcal{\hat{U}}^{\top}\delta\| 
\\ & \leq \frac{\mathcal{C} \sqrt{\ln p}}{ p \sqrt{n}}       \|\mathcal{\hat{U}}^{\top}\delta\| 
\end{split}
\end{equation}
with probability at least $1 - \mathcal{O}(p^{-2})$. 
For the above inequality, we have used Lemma~\ref{lem:basic_bounds} to bound
$\|\hat{\mathcal{D}} - \mathcal{D}\|$ and used Theorem~\ref{sharp
  two-to-infinity} to bound $\|\hat{\mathcal{U}} -  \mathcal{U}
\Xi\|_{2 \rightarrow \infty}$. Finally we used \cref{algebraic1},
\cref{algebraic3} and \cref{eq:diff_projection}
to bound $\|\hat{\mathcal{U}}^{\top} \mathcal{U} - \Xi\|$.

Next let $T = \mathcal{I}_{p} - \mathcal{\hat{U}}\mathcal{\hat{U}}^{\top}$. Then
\begin{equation}
  \label{eq:bound_part2}
  \begin{split}
\|(\mathcal{U} \mathcal{U}^{\top}\mathcal{\hat{U}}\mathcal{\hat{D}} \mathcal{\hat{U}} - \mathcal{U} \mathcal{D} \mathcal{U}^{\top}) \delta\|_{\infty}
& \leq\|\mathcal{U}(\mathcal{U}^{\top}\mathcal{\hat{U}}\mathcal{\hat{D}}-
\mathcal{D}\mathcal{U}^{\top}\mathcal{\hat{U}})\mathcal{\hat{U}}^{\top}\delta\|_{\infty}+
\|\mathcal{U}\mathcal{D}\mathcal{U}^{\top}T \delta\|_{\infty}\\
& \leq \|\mathcal{U}\|_{2 \rightarrow \infty}
(\|\mathcal{U}^{\top}\mathcal{\hat{U}}\mathcal{\hat{D}}-\mathcal{D}\mathcal{U}^{\top}\mathcal{\hat{U}}\| + \|\mathcal{D}\| \times
  \|\mathcal{U}^{\top}T \|) \|\delta\|  \\
    &\leq \Bigl(\frac{\mathcal{C} \sqrt{d} \ln p}{p \sqrt{n}} + \frac{\mathcal{C} \sqrt{d \ln p}}{p \sqrt{n}}\Bigr) \times \sqrt{\frac{p \ln p}{n}} 
\leq \frac{\mathcal{C} \ln p}{n \sqrt{p}}
\end{split}
\end{equation}
with probability at least $1 - O(p^{-2})$. 
In the above derivations, we bound $\|\mathcal{U}^{\top} \hat{\mathcal{U}} \hat{\mathcal{D}} -
\mathcal{D} \mathcal{U}^{\top} \hat{\mathcal{U}}\|$ using \cref{eq:Utop_hatU_hatD}, and 
bound
$\|\mathcal{U}^{\top} T\| = \|\mathcal{U}^{\top}(\mathcal{I}_{p} - \hat{\mathcal{U}}
\hat{\mathcal{U}}^{\top})\|$ using \cref{algebraic2,algebraic3,eq:diff_projection}.
The bound for $\|\mathcal{D}\|$ and $\|\mathcal{U}\|_{2 \rightarrow \infty}$ follows from \cref{n_order_Divergent} and \cref{Bounded_Coherence}, respectively.

Combining \cref{eq:bound_part1,eq:bound_part2} we obtain
\begin{equation}\label{B-1}
    \|(\mathcal{\hat{U}}\mathcal{\hat{D}}\mathcal{\hat{U}}^{\top}-\mathcal{U}\mathcal{D}\mathcal{U}^{\top})\delta\|_{\infty} = \mathcal{O}\Bigl(\frac{\ln
p}{n \sqrt{p}}\Bigr)
\end{equation}
with probability at least $1 - \mathcal{O}(p^{-2})$. 

Using similar arguments as that for \cref{eq:bound_part1,eq:bound_part2} we also have
\begin{equation}\label{B-3}
  \begin{split}
  \frac{1}{\hat{\sigma}}  \|(\mathcal{\hat{U}}\mathcal{\hat{U}}^{\top}-\mathcal{U}\mathcal{U}^{\top})\delta\|_{\infty}
   &   
   \leq
   \frac{\|\bigl((\mathcal{\hat{U}}-\mathcal{U}\Xi)-\mathcal{U}(\mathcal{U}^{\top}\mathcal{\hat{U}}-\Xi)
   \bigr)\mathcal{\hat{U}}^{\top}\delta\|_{\infty}+
   \|\mathcal{U}\mathcal{U}^{\top} T
   \delta\|_{\infty}}{\hat{\sigma}} \\
   &\leq\mathcal{C}\Bigl(\sqrt{\frac{d^3 \ln{p}}{np}}+ \frac{\sqrt{d} \ln p}{n \sqrt{p}}\Bigr) \times \sqrt{\frac{p \ln p}{n}} +
     \mathcal{C}\sqrt{\frac{d \ln p}{np}} \times \sqrt{\frac{p \ln p}{n}}\\
    &=\mathcal{O}\Bigl(\frac{\ln{p}}{n}\Bigr)
  \end{split}
\end{equation}
with probability at least $1 - \mathcal{O}(p^{-2})$, where once again $T = \mathcal{I}_{p} - \hat{\mathcal{U}} \hat{\mathcal{U}}^{\top}$. Note that in the
above derivations we have used Lemma~\ref{lem:basic_bounds} to show
that $\hat{\sigma}^{-1}$ is bounded away from $0$ by some constant not
depending on $p$ and $n$. 

Finally we can also replace $\delta$ with $\Sigma^{1/2}\zeta$ in the derivations of \cref{eq:bound_part1,eq:bound_part2,B-3} to obtain
\begin{gather}
\label{C-1}
    \|(\mathcal{\hat{U}}\mathcal{\hat{D}}\mathcal{\hat{U}}^{\top}-\mathcal{U}\mathcal{D}\mathcal{U}^{\top})\Sigma^{1/2}\zeta\|_{\infty} = \mathcal{O}\Bigl(\sqrt{\frac{\ln
p}{n p}}\norm{\zeta}_{2} \Bigr) \\
\label{C-3}
    \|\hat{\sigma}^{-1}\bigl(\mathcal{\hat{U}}\mathcal{\hat{U}}^{\top}-\mathcal{U}\mathcal{U}^{\top})\Sigma^{1/2}\zeta\|_{\infty}=\mathcal{O}\Bigl(\sqrt{\frac{\ln{p}}{n}}\norm{\zeta}_{2}\Bigr)
\end{gather}
simultaneously, with probability at least $1 - \mathcal{O}(p^{-2})$. A summary of the bounds for the terms 
$(A)$, $(B)$-$(\rom{1})$ through $(B)$-$(\rom{3})$, and $(C)$-$(\rom{1})$ through
$(C)$-$(\rom{3})$, are provided in \cref{tab:asymptotic term}. Combining the terms in this table we obtain the bound for $\|\hat{\zeta} - \zeta\|_{\infty}$ given in \cref{eq:zetahat_zeta}
(note that both $\|\zeta\|$ and $\|\zeta\|_{\infty}$ are bounded, see \cref{min+card}). This concludes the proof of \cref{infty norm converge}. 
\begin{table}
 \caption{Asymptotic Order of Each Term:}
    \centering
    \begin{tabular}{|c|c|c|c|}
\hline
Expression & $\mathbf{v}=\delta$ & $\mathbf{v}=\Sigma^{1/2}\zeta$& Corresponding Terms \\
 \hline
$\|\mathcal{W}\mathbf{v}\|_{\infty}$&$\sqrt{\frac{\ln{p}}{n}}$ & n.a.&(A)\\[2ex]
\hline
$\bigl\|\bigl[\mathcal{\hat{U}}\mathcal{\hat{D}}\mathcal{\hat{U}}^{\top}-\mathcal{U}\mathcal{D}\mathcal{U}^{\top}\bigr]\mathbf{v}\bigr\|_{\infty}
$& $\frac{\ln{p}}{n \sqrt{p}}$ & $\sqrt{\frac{\ln{p}}{np}}\|\zeta\|$& (B)-$(\rom{1})$ and (C)-$(\rom{1})$ \\[2ex]
\hline
$\bigl\|(\hat{\sigma}^{-1}-\sigma^{-1})(\mathcal{I}_{p}-\mathcal{U}\mathcal{U}^{\top})\mathbf{v}\bigr\|_{\infty}$& $\frac{\ln{p}}{n}$ & $\sqrt{\frac{\ln{p}}{n}}\norm{\zeta}_{\infty}$& (B)-$(\rom{2})$ and (C)-$(\rom{2})$ \\[2ex]
\hline
$\|\hat{\sigma}^{-1}(\hat{\mathcal{U}}\hat{\mathcal{U}}^{\top}-\mathcal{U}\mathcal{U}^{\top})\mathbf{v}\|_{\infty}$& $\frac{\ln{p}}{n}$ & $\sqrt{\frac{\ln{p}}{n}}\; \norm{\zeta} $& (B)-$(\rom{3})$ and (C)-$(\rom{3})$ \\[2ex]
\hline
\end{tabular}
    \label{tab:asymptotic term}
\end{table}

Finally, for ease of reference, we state two collaries summarizing the main derivations in the proof of Theorem \ref{infty norm converge}. These corollaries will be used in the proof of \cref{selection consistency and error consistency} below. 

\begin{corollary}\label{W-W hat} Suppose that \cref{spiked_covariance} through \cref{Bounded_Coherence}
  are satisfied. Let $\mathbf{v}$ be either a {\em fixed} vector in
  $\mathbb{R}^{p}$ or a $p$-variate sub-Gaussian random vector with
  $\mathbb{E}[\mathbf{v}] = \bm{0}$. We then have
\begin{align*}
    \norm{(\hat{\mathcal{W}}-\mathcal{W})\mathbf{v}}_{\infty}=\begin{cases}
    \mathcal{O}\bigl(n^{-1/2} (\ln p) \max_{i} \varsigma_{i} \bigr) &\text{if $\mathbf{v}$ is a sub-Gaussian vector}\\
    \mathcal{O}\bigl(n^{-1/2} (\ln p)^{1/2} \|\Sigma^{-1/2}\mathbf{v}\| \bigr) &\text{if $\mathbf{v}$ is a constant vector}
    \end{cases}
\end{align*}
with probability at least $1 - \mathcal{O}(p^{-2})$, where $\varsigma_{i}^2$ is the variance of the $i^{th}$ element of $\mathbf{v}$.
\end{corollary}
\begin{corollary}\label{coro2} Suppose that \cref{spiked_covariance} through \cref{Bounded_Coherence} are satisfied. Let $\mathbf{v}$ be a $p$-variate sub-Gaussian random
  vector with $\mathrm{Var}[\mathbf{v}]=c\;n^{-1}\Sigma$ for some
  finite $c > 0$. We then have
\begin{align*}
    \bigl\|\hat{\mathcal{W}}\mathbf{v}-\mathcal{W}\mathbb{E}[\mathbf{v}]\|_{\infty}=\mathcal{O}\Bigl(\sqrt{n^{-1}\ln p} \Bigr)
\end{align*}
with probability at least $1 - \mathcal{O}(p^{-2})$.
\end{corollary}

\subsection{Proof of Theorem~\ref{selection consistency and error consistency}}
\label{sec:proof_selection1}
For simplicity of notation we will write $\mathcal{S}$ instead of
$\mathcal{S}_{\zeta}$ since Theorem~\ref{selection consistency and
  error consistency} only depends on the whitened vector $\zeta$. Now
recall the definition of $\Tilde{\mathcal{S}}$ as
$$\Tilde{\mathcal{S}} = \{j:|\hat{\zeta}_{j}|> t_{n}\}$$
where $t_n = \bigl(\tfrac{\ln p}{n}\bigr)^{\alpha}$ for
some constant $0 < \alpha < \tfrac{1}{2}$. We now show $\Tilde{\mathcal{S}} = \mathcal{S}$
asymptotically almost surely. 

From \cref{infty norm converge} there exists a choice of $\mathcal{C}$ such
that if $\beta_n = \mathcal{C} \sqrt{n^{-1} \ln p}$ then
\begin{equation*}
     \mathbb{P}\Bigl(\bigcup_{j=1}^p\{|\zeta_{j}-\hat{\zeta}_{j}|>\beta_{n}\}\Bigr) =
     \mathcal{O}(p^{-2}).
\end{equation*}
Now suppose $\mathcal{S}^{c}\cap\Tilde{\mathcal{S}}\neq\emptyset$ where $(\cdot)^{c}$ denote set complement. 
Then there exists a $j$ such that $\zeta_{j}=0$ and
$|\hat{\zeta}_{j}|>t_{n}$, and for this $j$ we have $|\zeta_j -
\hat{\zeta}_j| > t_n > \beta_n$, provided that $n$ is sufficiently
large. We thus have
\begin{equation}
  \label{eq:sym_diff_1}
\mathbb{P}(\mathcal{S}^{c}\cap\Tilde{\mathcal{S}}\neq\emptyset)
\leq
\mathbb{P}\Bigl(\bigcup_{j=1}^p\{|\zeta_{j}-\hat{\zeta}_{j}|>\beta_{n}\Bigr)
= \mathcal{O}(p^{-2}).
\end{equation}
Similarly, if $\mathcal{S}\cap\Tilde{\mathcal{S}}^{c}\neq\emptyset$
then there exist a $j$ such that $|\zeta_{j}|>\mathcal{C}_{0}$ and $|\hat{\zeta}_{j}|\leq
t_{n}$. Recall that $\mathcal{C}_{0} > 0$ is the constant appearing in
\cref{min+card}; in particular, $\mathcal{C}_0$ does not depend on $n$ and
$p$. By the reverse triangle inequality, 
$|\zeta_{j}-\hat{\zeta_{j}}|>\mathcal{C}_{0}-t_{n} > \beta_n$ for sufficiently large $n$
and hence
\begin{equation}
  \label{eq:sym_diff_2}
  \mathbb{P}(\mathcal{S}\cap\Tilde{\mathcal{S}}^{c}\neq\emptyset) \leq
\mathbb{P}\Bigl(\bigcup_{j=1}^p\{|\zeta_{j}-\hat{\zeta}_{j}|>\beta_{n}\Bigr)
= \mathcal{O}(p^{-2}).
\end{equation}
Combining \cref{eq:sym_diff_1} and \cref{eq:sym_diff_2}
yields $\mathbb{P}(\mathcal{S} \not = \tilde{\mathcal{S}}) = \mathcal{O}(p^{-2})$ and hence, as
$p \rightarrow \infty$, by the Borel-Cantelli lemma we have $\mathcal{S} = \tilde{\mathcal{S}}$ asymptotically almost surely. 

We now show that the error rate for $\mathrm{lda} \circ \mathrm{pca}$ converges to the
Bayes error rate $R_F$ asymptotically almost surely. From the
description of $\mathrm{lda} \circ \mathrm{pca}$ in \cref{SCLDA rule}, it is sufficient to show
that
\begin{gather}
  \label{eq:conv_1}
  \hat{\zeta}_{\tilde{\mathcal{S}}}^{\top}\bigl[\hat{\mathcal{W}} \bigl(\mathbf{Z} - \tfrac{\bar{X}_1 + \bar{X}_2}{2}\bigr) \bigr]_{\tilde{\mathcal{S}}} -
  \zeta_{\mathcal{S}}^{\top} \bigl[\mathcal{W} \bigl(\mathbf{Z} - \tfrac{\mu_1 +
  \mu_2}{2}\bigr)\bigr]_{\mathcal{S}} \overset{\mathrm{p}}{\longrightarrow} 0, \\
  \label{eq:conv_2}
  \ln \frac{n_1}{n_2} \overset{\mathrm{p}}{\longrightarrow} \ln \frac{\pi_1}{1 - \pi_1}.
\end{gather}
Note that the convergence in \cref{eq:conv_1} is
with respect to a random testing sample $\mathbf{Z} \sim
\pi_1 \mathcal{N}(\mu_1, \Sigma) + (1 - \pi_1) \mathcal{N}(\mu_2,
\Sigma)$ together with the training data, while the convergence in
Eq.\eqref{eq:conv_2} is with respect to the training data only. 
As Eq.~\eqref{eq:conv_2} follows directly from the strong law of large
numbers, we thus focus our efforts on showing \cref{eq:conv_1}.

First, suppose
$\tilde{\mathcal{S}} = \mathcal{S}$ and let $h(\mathbf{Z}) = \hat{\zeta}_{\tilde{\mathcal{S}}}^{\top}\bigl[\hat{\mathcal{W}} \bigl(\mathbf{Z} - \tfrac{\bar{X}_1 + \bar{X}_2}{2}\bigr) \bigr]_{\tilde{\mathcal{S}}} -
  \zeta_{\mathcal{S}}^{\top} \bigl[\mathcal{W} \bigl(\mathbf{Z} - \tfrac{\mu_1 +
  \mu_2}{2}\bigr)\bigr]_{\mathcal{S}}$. 
Then 
\begin{equation}
  \begin{split}
      \bigl| h(\mathbf{Z}) \bigr|
  & \leq s_{0}
      \bigl\|\bigl[\Hat{\mathcal{W}}\bigl(\mathbf{Z}-\tfrac{\Bar{X}_{1}+\Bar{X}_{2}}{2}\bigr)-\mathcal{W}\bigl(\mathbf{Z}-\tfrac{\mu_{1}+\mu_{2}}{2}\bigr)\bigr]_{\mathcal{S}}\bigr\|_{\infty}\|\zeta\|_{\infty} \\
     & + s_{0}\;
       \bigl\|\bigl[\mathcal{W}\bigl(\mathbf{Z}-\tfrac{\mu_{1}+\mu_{2}}{2}\bigr)\bigr]_{\mathcal{S}}\bigr\|_{\infty}\|
       \hat{\zeta} - \zeta\|_{\infty}\\
      & +s_{0}
\bigl\|\bigl[\Hat{\mathcal{W}}\bigl(\mathbf{Z}-\tfrac{\Bar{X}_{1}+\Bar{X}_{2}}{2}\bigr)-\mathcal{W}\bigl(\mathbf{Z}-\tfrac{\mu_{1}+\mu_{2}}{2}\bigr)\bigr]_{\mathcal{S}}\bigr\|_{\infty}
\|\hat{\zeta} - \zeta\|_{\infty}.
\end{split}
\end{equation}
The bounds for $\|\zeta\|_{\infty}$ and $\|\hat{\zeta} - \zeta\|_{\infty}$ are given in \cref{min+card} and \cref{infty norm converge}, respectively. It thus suffices to bound
\begin{align}
     (D) :=\bigl\|\bigl[\Hat{\mathcal{W}}\bigl(\mathbf{Z}-\tfrac{\Bar{X}_{1}+\Bar{X}_{2}}{2}\bigr)-\mathcal{W}\bigl(\mathbf{Z}-\tfrac{\mu_{1}+\mu_{2}}{2}\bigr)\bigr]_{\mathcal{S}}\bigr\|_{\infty}\nonumber\\
     (E) :=\bigl\|\bigl[\mathcal{W}\bigl(\mathbf{Z}-\tfrac{\mu_{1}+\mu_{2}}{2}\bigr)\bigr]_{\mathcal{S}}\bigr\|_{\infty}\nonumber
\end{align}
Write
$\mathbf{Z}=\mu_{z}+\Sigma^{1/2}\epsilon_{z}$ where 
$\mu_{z}=\mu_{1}$ if $\bm{Z}$ is sampled from class $1$ and $\mu_{z}=\mu_{2}$ otherwise.
The term
$\|[(\Hat{\mathcal{W}}-\mathcal{W})\Sigma^{1/2}\epsilon_{z}]_{\mathcal{S}}\bigr\|_{\infty}$
can be analyzed using the decomposition for
$\Hat{\mathcal{W}}-\mathcal{W}$ given in
\cref{eq:decomposition_hatW-W} together with similar arguments to
that for deriving
\cref{B-2,eq:bound_part1,eq:bound_part2,B-3}. In particular we have, with probability at least $1-\mathcal{O}(n^{-2})$, that
\begin{align}\label{s_row_max}
    \bigl\|\bigl[(\Hat{\mathcal{W}}-\mathcal{W})\Sigma^{1/2}\epsilon_{z}\bigr]_{\mathcal{S}}\bigr\|_{\infty}=\mathcal{O}(\sqrt{\frac{\ln{n}\ln{p}}{n}})
\end{align}

Next, using Corollary \ref{W-W hat}, Corollary \ref{coro2} and \cref{s_row_max}, we obtain, with probability at least $1-\mathcal{O}(n^{-2})$
\begin{align*}
(D) &\leq \bigl\|\bigl[(\Hat{\mathcal{W}}-\mathcal{W})\Sigma^{1/2}\epsilon_{z}\bigr]_{\mathcal{S}}\bigr\|_{\infty}+\bigl\|(\Hat{\mathcal{W}}-\mathcal{W})\mu_{z}\bigr\|_{\infty}+\frac{1}{2}\bigl\|\Hat{\mathcal{W}}\Bar{X}_{1}-\mathcal{W}\mu_{1}\bigr\|_{\infty}+\frac{1}{2}\bigl\|\Hat{\mathcal{W}}\Bar{X}_{2}-\mathcal{W}\mu_{2}\bigr\|_{\infty}\\
    & \leq \mathcal{C}\Bigl(\sqrt{\frac{\ln{n}\ln{p}}{n}}+\sqrt{\frac{\ln{p}}{n}} +\sqrt{\frac{\ln{p}}{n}}\Bigr) =\mathcal{O}(\sqrt{\frac{\ln{n}\ln{p}}{n}}).
\end{align*}


Thirdly, we have
\begin{equation}\label{bdd z function}
  \begin{split}
    (E)  &\leq\bigl\|\mathcal{W}\bigl(\mu_{z}-\tfrac{\mu_{1}+\mu_{2}}{2}\bigr)\bigr\|_{\infty}+\bigl\|\bigl[\mathcal{W}(\mathbf{Z}-\mu_{z})\bigr]_{\mathcal{S}}\bigr\|_{\infty} 
           = \frac{1}{2}\norm{\zeta}_{\infty}+\norm{[\epsilon_{z}]_{\mathcal{S}}}_{\infty} =: \vartheta(\mathbf{Z})
    \end{split}
\end{equation}
where $[\epsilon_{z}]_{\mathcal{S}}$ is a mean $0$ sub-Gaussian vector
in $\mathbb{R}^{s_{0}}$ and $\mathrm{Var}[[\epsilon_z]_{\mathcal{S}}] =
\mathcal{I}_{s_0}$. Therefore, by \cref{min+card} and properties of sub-Gaussian random vectors, the term
$\vartheta(\mathbf{Z})$ is bounded in probability. 

Combining the above bounds, we conclude that with probability at least
$1 - \mathcal{O}(n^{-2})$, $\tilde{\mathcal{S}} = \mathcal{S}$ and 
\begin{align*}
\bigl| \hat{\zeta}_{\tilde{\mathcal{S}}}^{\top}\bigl[\hat{\mathcal{W}} \bigl(\mathbf{Z} - \tfrac{\bar{X}_1 + \bar{X}_2}{2}\bigr) \bigr]_{\tilde{\mathcal{S}}} -
  \zeta_{\mathcal{S}}^{\top} \bigl[\mathcal{W} \bigl(\mathbf{Z} - \tfrac{\mu_1 +
  \mu_2}{2}\bigr)\bigr]_{\mathcal{S}} \bigr|
      \leq
        \mathcal{C}s_{0}\Bigl(\sqrt{\frac{\ln{n}\ln{p}}{n}}\norm{\zeta}_{\infty}+\vartheta(\mathbf{Z})\sqrt{\frac{\ln
        p}{n}}\Bigr).
\end{align*}
Hence, for $\ln{n}\ln{p} = o(n)$, we have 
$$\bigl| \hat{\zeta}_{\tilde{\mathcal{S}}}^{\top}\bigl[\hat{\mathcal{W}} \bigl(\mathbf{Z} -
\tfrac{\bar{X}_1 + \bar{X}_2}{2}\bigr) \bigr]_{\tilde{\mathcal{S}}} -
  \zeta_{\mathcal{S}}^{\top} \bigl[\mathcal{W} \bigl(\mathbf{Z} - \tfrac{\mu_1 +
    \mu_2}{2}\bigr)\bigr]_{\mathcal{S}} \bigr| \longrightarrow 0$$
in probability. This completes the proof of \cref{selection consistency and error consistency}. 

\subsection{Theoretical results for Section~\ref{sec:multi-class}}
We now present theoretical results for multi-class $\mathrm{lda} \circ \mathrm{pca}$.
We first assume that the whitened directions
$\{\zeta^{(i)}\}$, the whitened indices $\{\mathcal{S}_i\}$, and the
covariance $\Sigma$ satisfy the following two conditions, which 
are natural generalizations of \cref{min+card} and
\cref{n_order_Divergent} to the multi-class setting. 
\begin{ass}\label{multi_min+card} Let $|\mathcal{S}_{i}|=s_{i} > 0$ for each $i=2,\cdots,K$. Recall
  that $\mathcal{S}_{i}$ is the set of indices $j$ for which $\zeta_{ij}
  \not = 0$. Let $\mathcal{C}_{0}>0$, $M > 0$ and
  $\mathcal{C}_{\zeta}>0$ be constants not depending on $p$ such that $\max_{i}s_{i} \leq M$
  and
  \begin{align*}
\min_{i=2,\cdots,K}\min_{j\in\mathcal{S}_{i}} |\zeta_{ij}| \geq \mathcal{C}_{0}, \qquad
                                \max_{i\in[K]}\|\Sigma^{-1/2}
                                \mu_{i}\|\leq \mathcal{C}_{\zeta}.
\end{align*}
\end{ass} 

\begin{ass} \label{multi_n_order_Divergent} Let $\sigma > 0$ be
  fixed and that, for sufficiently large $p$,  $n_1,\cdots, n_{K}$ and $p$ satisfy
\begin{align*} 
    \frac{n_{i}}{n_{j}} = \Theta(1) \,\, \text{for $\:i\neq j,\:i,j\in[K]$} \quad \text{and} \quad \quad \ln{p}=
    o(n).
\end{align*}
Furthermore, for sufficiently large $p$, the spiked eigenvalues
$\lambda_1, \dots, \lambda_d$ satisfy
\begin{equation*}
    \lambda_{k}=\Theta(p),\qquad \text{for all $k\in[d]$}.
\end{equation*}
\end{ass}
Given the above conditions, the next result extends
Theorem~\ref{infty norm converge} (and has an identical proof) to bound
$\|\hat{\zeta}^{(i)}-\zeta^{(i)}\|_{\infty}$ for $i\geq2$.
\begin{Thm} 
Under \cref{spiked_covariance}, \cref{Bounded_Coherence}, \cref{multi_min+card} and \cref{multi_n_order_Divergent}, there exists a constant $C>0$ such that with probability at least $1-\mathcal{O}(p^{-2})$,  
\begin{align} 
\max_{i=2,\cdots,K}\|\hat{\zeta}^{(i)}-\zeta^{(i)}\|_{\infty}\leq\mathcal{C}\sqrt{\frac{\ln{p}}{n}}. 
\end{align} 
\end{Thm} 
We now consider a hard thresholding estimate $\Tilde{\zeta}^{(i)}$ for
recovering $\zeta^{(i)}$. For a
given $i \geq 2$, let 
\begin{align}
  \label{eq:threshold_Kclass}
    \Tilde{\zeta}^{(i)}_{j} =\hat{\zeta}_{j}^{(i)}\mathbbm{1}(|\hat{\zeta}_j^{(i)}|>t_{n}),\qquad j \in [p]
\end{align}
where $t_{n}=\big(\ln{p}/n\big)^{\alpha}$ for some constant $0<\alpha<\frac{1}{2}$. Here we assume, for
simplicity, that $\alpha$ takes the same value for all
classes. 
Given $\Tilde{\zeta}^{(i)}$, define the associated active
set $\Tilde{S}_{i}=\{j:\Tilde{\zeta}_{j}^{(i)}\neq0\}$. The next
result extends Theorem~\ref{selection consistency and error
consistency} (and has an identical proof) to show that $\mathrm{lda} \circ \mathrm{pca}$ is also asymptotically
Bayes-optimal in the multi-class setting. However, we note that (to the best of our knowledge), there is no closed-form explicit
expression for the Bayes error $R_F$ when classifying data from a mixture of $K \geq 3$ multiviarate normals. 

\begin{Thm} Suppose 
  that $\mathbf{Z} \sim \sum_{i=1}^{K}\pi_i
\mathcal{N}_{p}(\mu_{i},\Sigma)$ where $\pi_{i}\geq0$ and $\sum_{i=1}^{K}\pi_i=1$. Suppose \cref{spiked_covariance}, 
\cref{Bounded_Coherence}, \cref{multi_min+card} and \cref{multi_n_order_Divergent} are satisfied. We then have 
\begin{equation}  
  \label{eq:multi_exact_recovery} 
    \max_{i=2,\cdots,K}\mathbb{P}(\Tilde{\mathcal{S}}_{i}\neq\mathcal{S}_{i}) = \mathcal{O}(p^{-2}). 
\end{equation} 
Furthermore we also have 
    $\hat{R}_{\mathrm{lda} \circ \mathrm{pca}}-R_{F} \rightarrow 0$ 
almost surely as $n, p \rightarrow \infty$. 
\end{Thm} 

\subsection{Theoretical results for Section~\ref{PCQDA}}
We now present theoretical results for $\mathrm{qda} \circ \mathrm{pca}$.
which depend on the following variant of  
\cref{min+card} through \cref{Bounded_Coherence} for 
heterogeneous covariance matrices. 
\begin{ass}\label{QDA min+card} Let $|\mathcal{A}_{i}|=\mathit{a}_{i} > 0$. Recall that $\mathcal{A}_{i}$ is the set of indices $j$ for which $
  \zeta_{ij} \not = 0,\;i=1,2$ where $\zeta_{i} = \Sigma_i^{-1/2} \mu_i$. Let $\mathcal{C}_{0}>0$, $M > 0$ and
  $\mathcal{C}_{\zeta}>0$ be constants not depending on $p$ such that $\max\{\mathit{a}_{1},\mathit{a}_{2}\} \leq M$
  and
  \begin{align*}
\min_{i \in \{1,2\}}\min_{j\in\mathcal{A}_{i}} |\zeta_{ij}| \geq 
\mathcal{C}_{0}, \qquad
\max_{i \in \{1,2\}}\max_{k \in \{1,2\}}\|\Sigma_{i}^{-1/2}\mu_{k}\| \leq \mathcal{C}_{\zeta}.
\end{align*}
\end{ass} 

\begin{ass} \label{QDA n order Divergent} Let $\sigma_{i} > 0$ be
  fixed and that, for sufficiently large $p$,  $n_1, n_2$ and $p$ satisfy
\begin{align*} 
    \frac{n_{1}}{n_{2}} = \Theta(1), \quad \ln{p}=o(n).
\end{align*}
Furthermore, for sufficiently large $p$, the spiked eigenvalues
$\lambda_{i1}, \dots, \lambda_{id_{i}}$ satisfy
\begin{equation*}
    \lambda_{ik}=\mathcal{O}(p),\qquad \text{for all $k\in[d_{i}],\:i=1,2$}.
\end{equation*}
\end{ass}

\begin{ass}[Bounded Coherence]  \label{QDA Bounded Coherence}
  There is a constant $\mathcal{C}_{\mathcal{U}}\geq1$ 
  independent of $n$ and $p$ such that 
\begin{equation*}
    \norm{\mathcal{U}_{i}}_{2\rightarrow\infty}\leq
    \frac{\mathcal{C}_{\mathcal{U}} \sqrt{d_i}}{\sqrt{p}}, \qquad \text{for $i=1,2$.} \\
\end{equation*}
\end{ass}
\cref{QDA min+card} guarantees that the noncentrality parameters for these $\chi^{2}_1$ are strictly positive and finite, 
so that the Bayes error rate $R_F$ is stricly bounded away from $0$ 
and $\min\{\pi_1, 1 - \pi_1\}$ (which corresponds to random guessing). Note that if $\Sigma_1 \not = \Sigma_2$ then there is no simple closed-form 
expression for $R_F$ as it depends on the 
tail behavior of a linear combination of independent, {\em noncentral} $\chi^{2}_1$ random variables;  
see \citet[Section~6.10]{multi_error_Anderson} for more details.  
\cref{QDA n order Divergent} allows the 
spiked eigenvalues for each $\Sigma_i$ 
to grow linearly with the dimension $p$, in contrast 
to the bounded eigenvalues assumption frequently encountered in the literature 
\citep{QDA_Shao,QDA_TonyCai}. 


We then have the following extensions of 
Theorem~\ref{infty norm converge} and \cref{selection consistency and error consistency}.
\begin{Thm} 
  \label{thm:infty_norm_QDA}
Under \cref{QDA spiked_covariance} through  \ref{QDA Bounded Coherence}, 
there exists a constant $C>0$ such that  
\begin{align} 
  \label{eq:hatzeta_err_hetero}
\max_{i=1,2}\|\hat{\zeta}_{i}-\zeta_{i}\|_{\infty}\leq\mathcal{C}\sqrt{\frac{\ln{p}}{n}}. 
\end{align} 
with probability at least $1-O(p^{-2})$. 

  Construct $\Tilde{\zeta}_i$ as in \cref{hard thresholding} 
  for $i \in \{1,2\}$ and let $\tilde{\mathcal{A}}_i$ be the indices 
  for the non-zero coordinates of $\Tilde{\zeta}_i$. 
  Let $\mathbf{Z} \sim \pi_1
\mathcal{N}_{p}(\mu_{1},\Sigma_{1})+
(1-\pi_1)\mathcal{N}_{p}(\mu_{2},\Sigma_{2})$. Then
\begin{equation}  
  \label{eq:QDA_exact_recovery} 
    \max_{i=1,2}\mathbb{P}(\Tilde{\mathcal{A}}_{i}\neq\mathcal{A}_{i}) = \mathcal{O}(p^{-2}). 
\end{equation} 
Furthermore, we also have
$\hat{R}_{\mathrm{qda} \circ \mathrm{pca}}-R_{F}\rightarrow 0$
in probability, as $n, p \rightarrow \infty$. 
\end{Thm} 

\begin{proof}
For conciseness, we omit the derivations of \cref{eq:hatzeta_err_hetero} and
\cref{eq:QDA_exact_recovery} as they follow the same argument as that in the proof of \cref{infty norm converge} and \cref{eq:exact_recovery}. 

In order to show that $Q(\mathbf{Z}\:|\,\bar{X}_{i}, \hat{\mathcal{W}}_{i},\hat{\mathcal{A}}_{0})$
is a consistent estimate for
$Q(\mathbf{Z}\:|\,\mu_{i}, \mathcal{W}_{i},\mathcal{A}_{0})$, it suffices to show that the following quantities 
\begin{align}
    q_1(\mathbf{Z}) &:= \bigl|\bigl[\mathcal{W}_{1}\bigl(\mathbf{Z}-\mu_{1}\bigr)\bigr]^{\top}_{\mathcal{A}_{0}}\bigl[\mathcal{W}_{1}\bigl(\mathbf{Z}-\mu_{1}\bigr)\bigr]_{\mathcal{A}_{0}}-\bigl[\hat{\mathcal{W}}_{1}\bigl(\mathbf{Z}-\bar{X}_{1}\bigr)\bigr]^{\top}_{\mathcal{A}_{0}}\bigl[\hat{\mathcal{W}}_{1}\bigl(\mathbf{Z}-\bar{X}_{1}\bigr)\bigr]_{\mathcal{A}_{0}}\bigr|\label{W1 error}\\
   q_2(\mathbf{Z}) &:= \bigl|\bigl[\mathcal{W}_{2}\bigl(\mathbf{Z}-\mu_{2}\bigr)\bigr]^{\top}_{\mathcal{A}_{0}}\bigl[\mathcal{W}_{2}\bigl(\mathbf{Z}-\mu_{2}\bigr)\bigr]_{\mathcal{A}_{0}}-\bigl[\hat{\mathcal{W}}_{2}\bigl(\mathbf{Z}-\bar{X}_{2}\bigr)\bigr]^{\top}_{\mathcal{A}_{0}}\bigl[\hat{\mathcal{W}}_{2}\bigl(\mathbf{Z}-\bar{X}_{2}\bigr)\bigr]_{\mathcal{A}_{0}}\bigr|\label{W2 error}
\end{align}
both converge to $0$ as $n \rightarrow \infty$. 

Assume without loss of generality that $\mathbf{Z}$ is a test sample from class $1$, i.e,.
$\mathbf{Z}\sim\mathcal{N}_{p}(\mu_{1},\Sigma_{1})$. Then
\begin{equation}
  \begin{split}
    q_1(\mathbf{Z})
    &
    \leq a_0 \bigl\|[\mathcal{W}_{1}(\mathbf{Z}-\mu_{1})-\hat{\mathcal{W}}_{1}(\mathbf{Z}-\bar{X}_{1})]_{\mathcal{A}_{0}}\bigr\|^{2}_{\infty}\\
    &+ 2 a_0 \bigl\|\mathcal{W}_{1}(\mathbf{Z}-\mu_{1})]_{\mathcal{A}_{0}}\bigr\|_{\infty} \times
      \bigl\| [\mathcal{W}_{1}(\mathbf{Z}-\mu_{1})-\hat{\mathcal{W}}_{1}(\mathbf{Z}-\bar{X}_{1})]_{\mathcal{A}_{0}}\bigr\|_{\infty}
    \end{split}
\end{equation}
where $a_0 = |\mathcal{A}_0| \leq a_1 + a_2$, with $a_i = |\mathcal{A}_i|$. 

Let $h_1(\mathbf{Z}) := \bigl\|[\mathcal{W}_{1}(\mathbf{Z}-\mu_{1})-\hat{\mathcal{W}}_{1}(\mathbf{Z}-\bar{X}_{1})]_{\mathcal{A}_{0}}\bigr\|_{\infty}$. 
Then by \cref{W-W hat,coro2}, and following the same arguments as that for \cref{s_row_max}, we have
\begin{equation}
  \label{eq:proof_qda1}
  \begin{split}
    h_1(\mathbf{Z}) & \leq\|(\mathcal{W}_{1}-\hat{\mathcal{W}}_{1})\mu_{1}\|_{\infty}+\|
                      [(\mathcal{W}_{1}-\hat{\mathcal{W}}_{1})\Sigma_{1}^{1/2}\epsilon_{z}]_{\mathcal{A}_{0}}\|_{\infty}
                      +\|\hat{\mathcal{W}}_{1}\bar{X}_{1}-\mathcal{W}_{1}\mu_{1}\|_{\infty} \\
    &\leq \mathcal{C}\Bigl(\sqrt{\frac{\ln{p}}{n}}+\sqrt{\frac{\ln{n}\ln{p}}{n}} +\sqrt{\frac{\ln{p}}{n}}\Bigr)
  \end{split}
\end{equation} with probability at least $1 - \mathcal{O}(n^{-2})$,
where $\epsilon_z \sim \mathcal{N}(0, \mathcal{I}_{p})$.  An almost identical 
bound also holds for $h_2(\mathbf{Z}) :=
\bigl\|[\mathcal{W}_{2}(\mathbf{Z}-\mu_{2})-\hat{\mathcal{W}}_{2}(\mathbf{Z}-\bar{X}_{2})]_{\mathcal{A}_{0}}\bigr\|_{\infty}$, namely
\begin{equation}
  \begin{split}
    h_2(\mathbf{Z})  
     & \leq \|(\mathcal{W}_{2}-\hat{\mathcal{W}}_{2})\mu_{1}\|_{\infty}+\|(\mathcal{W}_{2}-\hat{\mathcal{W}}_{2})\Sigma_{1}^{1/2}\epsilon_{z}]_{\mathcal{A}_{0}}\|_{\infty}+\|\hat{\mathcal{W}}_{2}\bar{X}_{2}-\mathcal{W}_{2}\mu_{2}\|_{\infty}\\
       \leq& \mathcal{C}\Bigl(\sqrt{\frac{\ln{p}}{n}}\|\Sigma^{-1/2}_{2}\mu_{1}\|_{2}+\sqrt{\frac{\ln{n}\ln{p}}{n}} +\sqrt{\frac{\ln{p}}{n}}\Bigr)
 \end{split}
\end{equation}
with probability at least $1 - \mathcal{O}(n^{-2})$. 

Next, $\|\bigl[\mathcal{W}_{1}(\mathbf{Z}-\mu_{1})\bigr]_{\mathcal{A}_{0}}\bigr\|_{\infty}$ and $\|\bigl[\mathcal{W}_{1}(\mathbf{Z}-\mu_{1})\bigr]_{\mathcal{A}_{0}}\bigr\|_{\infty}$ can be bounded by the quantity $\vartheta(\mathbf{Z})$
as defined in \cref{bdd z function} which, by \cref{QDA min+card}, is bounded in probability. 



Finally, by Proposition~2 and Lemma~2 in \cite{QDA_lasso}, we have
$|\hat{\kappa} - \kappa| \rightarrow 0$ in probability. Combining the
above statements yield the proof of \cref{thm:infty_norm_QDA}.
\end{proof}
\bibstyle{plain}
\bibdata{main}
\end{document}